\documentclass{article}

\usepackage{times}
\usepackage{graphicx} 
\usepackage{subfigure} 

\usepackage{natbib}

\usepackage{algorithm}
\usepackage{algorithmic}

\usepackage{calc}
\usepackage{amsmath, amssymb, amsthm} 
\usepackage{parskip}
\usepackage{color,hyperref}
\usepackage{epsfig}
\usepackage{subfigure}
\usepackage{verbatim}
\usepackage{rotating}
\usepackage{kky}
\usepackage{mockusDefns}
\usepackage{multicol}
\usepackage{ifthen}

\newcommand{\toworkon}[1]{}
\newboolean{istwocolumn}
\newboolean{isabridged}
\setboolean{isabridged}{false} 

\usepackage{hyperref}

\usepackage[accepted]{icml2015}

\icmltitlerunning{Additive Gaussian Process Optimisation and Bandits}

\begin{document} 
\pdfoutput=1

\twocolumn[
\icmltitle{High Dimensional Bayesian Optimisation and Bandits via Additive
Models}


\icmlauthor{Kirthevasan Kandasamy}{kandasamy@cs.cmu.edu}
\icmlauthor{Jeff Schneider}{schneide@cs.cmu.edu}
\icmlauthor{Barnab\'as P\'oczos}{bapoczos@cs.cmu.edu}
\icmladdress{Carnegie Mellon University, Pittsburgh, PA, USA}

\icmlkeywords{Bayesian Optimization, Gaussian Process Bandits, Additive Gaussian
Processes }

\vskip 0.3in
]

\setboolean{istwocolumn}{true}

\begin{abstract}
\vspace{0.05in}
Bayesian Optimisation (BO) is a technique used in optimising a $D$-dimensional 
function  which is typically expensive to evaluate.
While there have been many successes for BO in low dimensions, scaling it to
high dimensions has been notoriously difficult.
Existing literature on the topic are under very restrictive settings.
In this paper, we identify two key challenges in this 
endeavour. We tackle these challenges by
assuming an additive structure for the function.
This setting is substantially more expressive and contains a richer class of
functions than previous work. 
We prove that, for additive functions the regret 
has only linear 
dependence on $D$
even though the function depends on all $D$ dimensions.
We also demonstrate several other statistical and computational benefits in our 
framework.
Via synthetic examples, a scientific simulation and a
face detection problem we demonstrate that our
method outperforms naive BO on additive functions and on 
several examples where the function is not additive.
\vspace{-0.215in}
\end{abstract}


\newcommand{\insertFigureAddGPGraph}{
\begin{figure}
\centering
\includegraphics[width=3.2in]{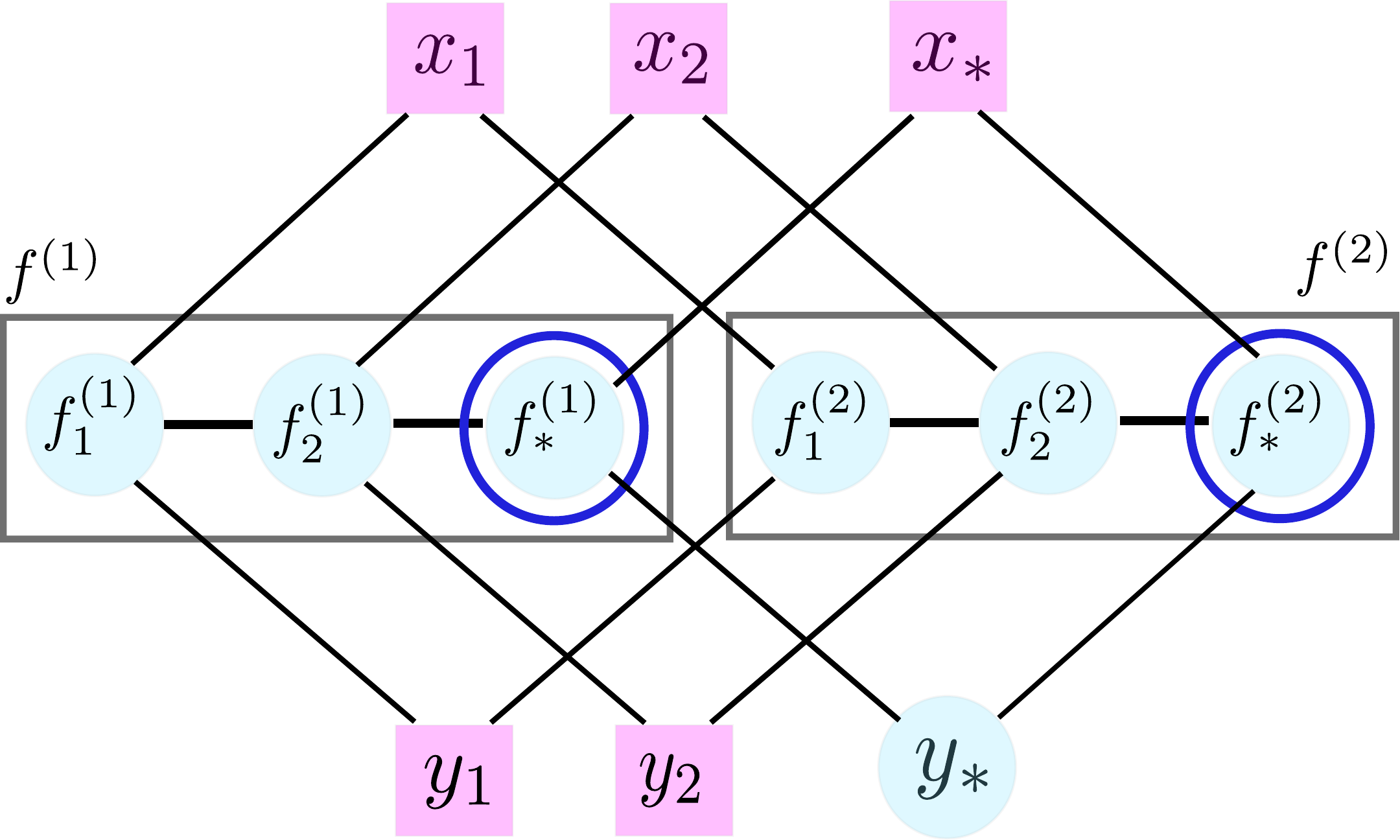}
\vspace{\imcaptionspace}
\caption[]{\small Illustration of the additive GP model for $2$ observations
where $M = 2$ in~\eqref{eqn:addmodel}. The squared variables
are observed while the circled variables are not. For brevity we have
denoted $\funcj_i = \funcj(\xii{j}_i)$ for $i = 1, 2, *$. We wish to infer the
posterior distributions of the individual GPs $\funcj(\xii{j}_*)$ 
(outlined in blue). }
\label{fig:addGPgraph}
\vspace{\imtextspace}
\end{figure}
}

\newcommand{\insertFigureLRGResults}{
\begin{figure}
  \includegraphics[width=\imsinglecol]{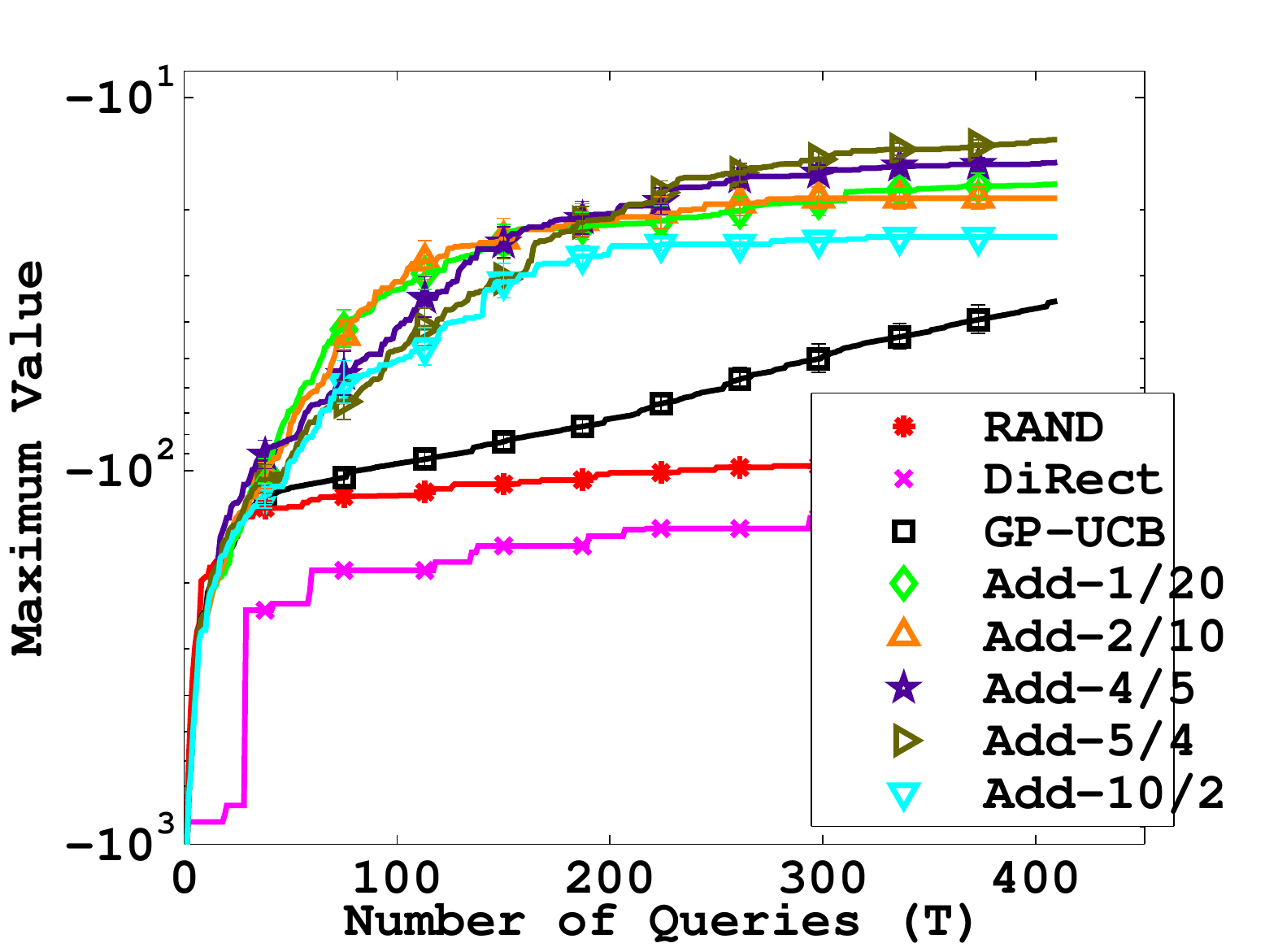} 
\vspace{\imcaptionspace}
\label{fig:lrg}
\caption{Results on the SDSS LRGs dataset. The $y$-axis is the log likelihood.}
\vspace{\imtextspace}
\end{figure}
}

\newcommand{\insertFigureVJResults}{
\begin{figure}
  \includegraphics[width=\imsinglecol]{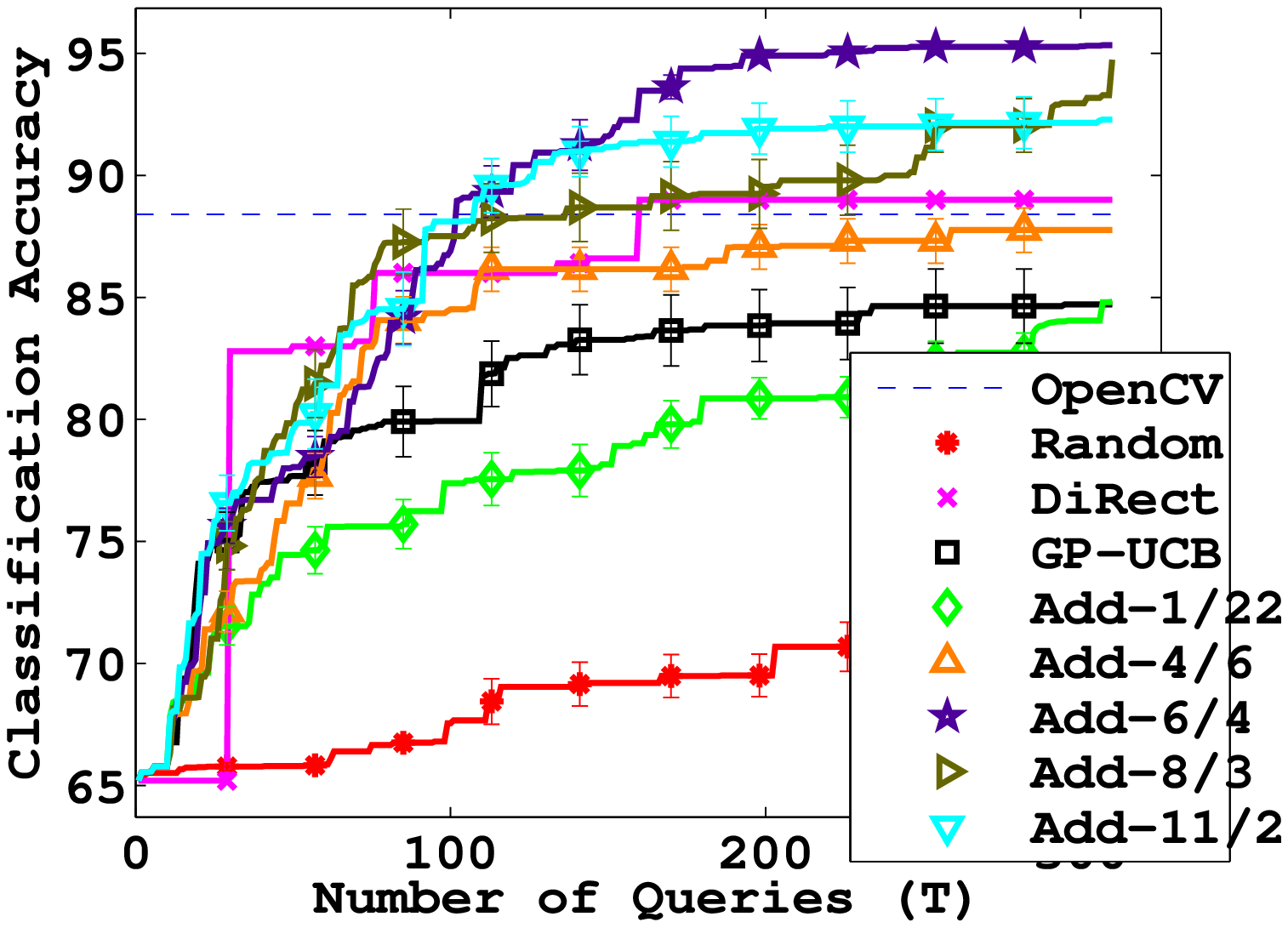} 
\vspace{\imcaptionspace}
\label{fig:vj}
\caption{Results on the Viola \& Jones Classification Problem. The $y$-axis is
the classification accuracy. The blue dashed
line is the accuracy of the setting given in OpenCV.}
\vspace{\imtextspace}
\end{figure}
}

\newcommand{\insertFigureToyResults}{
\begin{figure*}
\centering
\subfigure[]{
  \includegraphics[width=\imarrwthree]{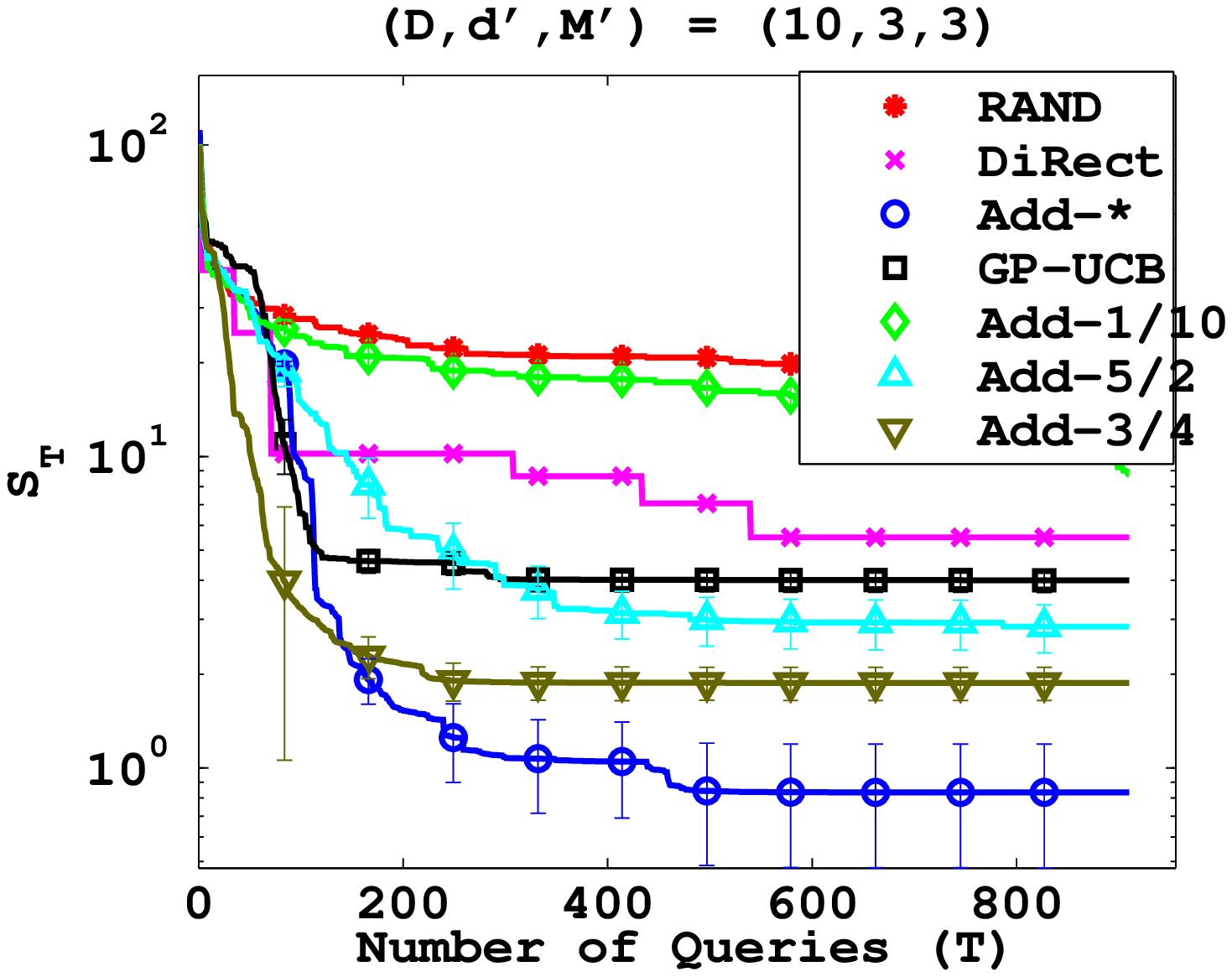} \hspace{\imhspthree}
  \vspace{\imlabelspace}
  \label{fig:SRD10d3}
}
\subfigure[]{
  \includegraphics[width=\imarrwthree]{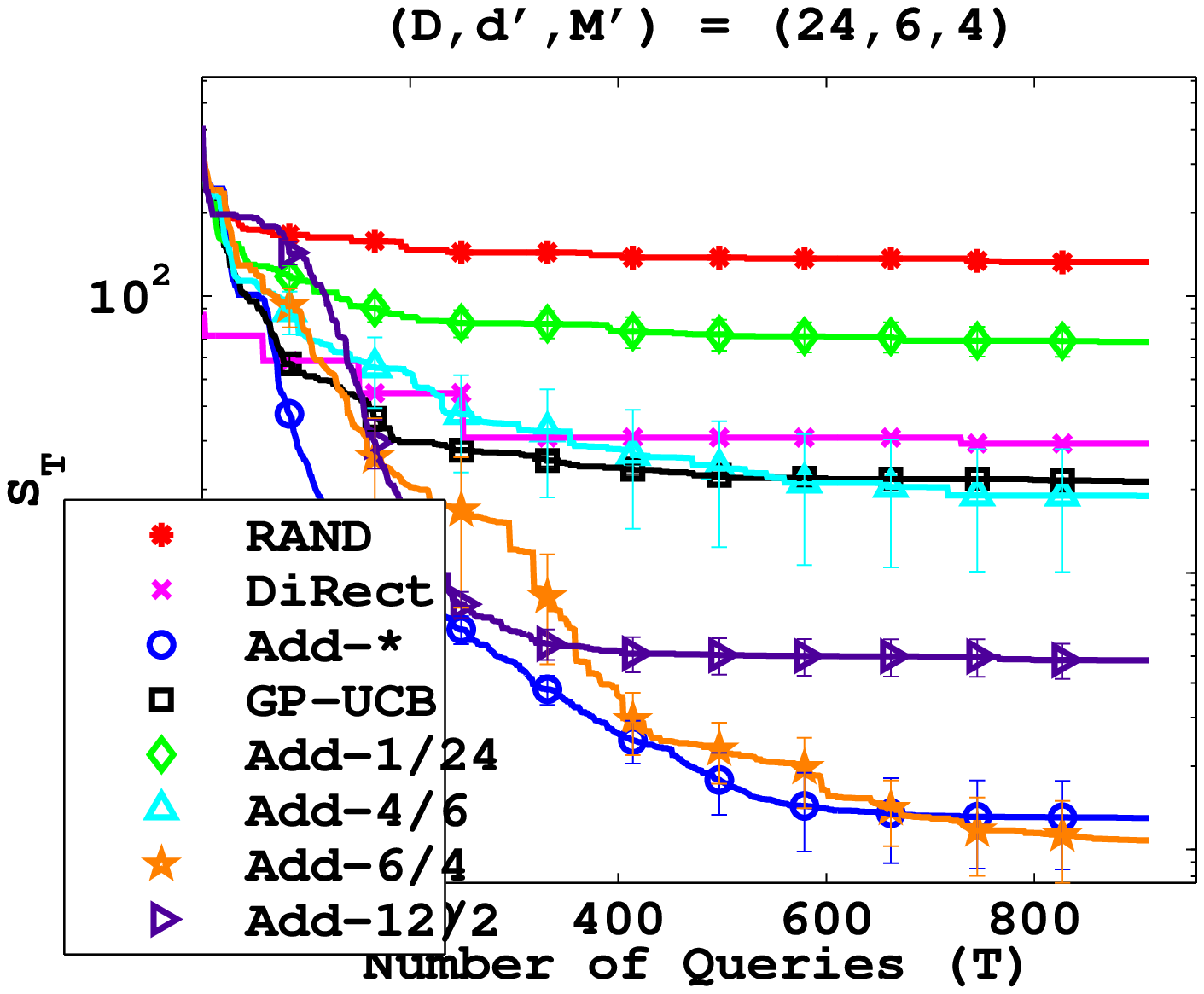} \hspace{\imhspthree}
  \vspace{\imlabelspace}
  \label{fig:SRD24d11}
}
\subfigure[]{
  \includegraphics[width=\imarrwthree]{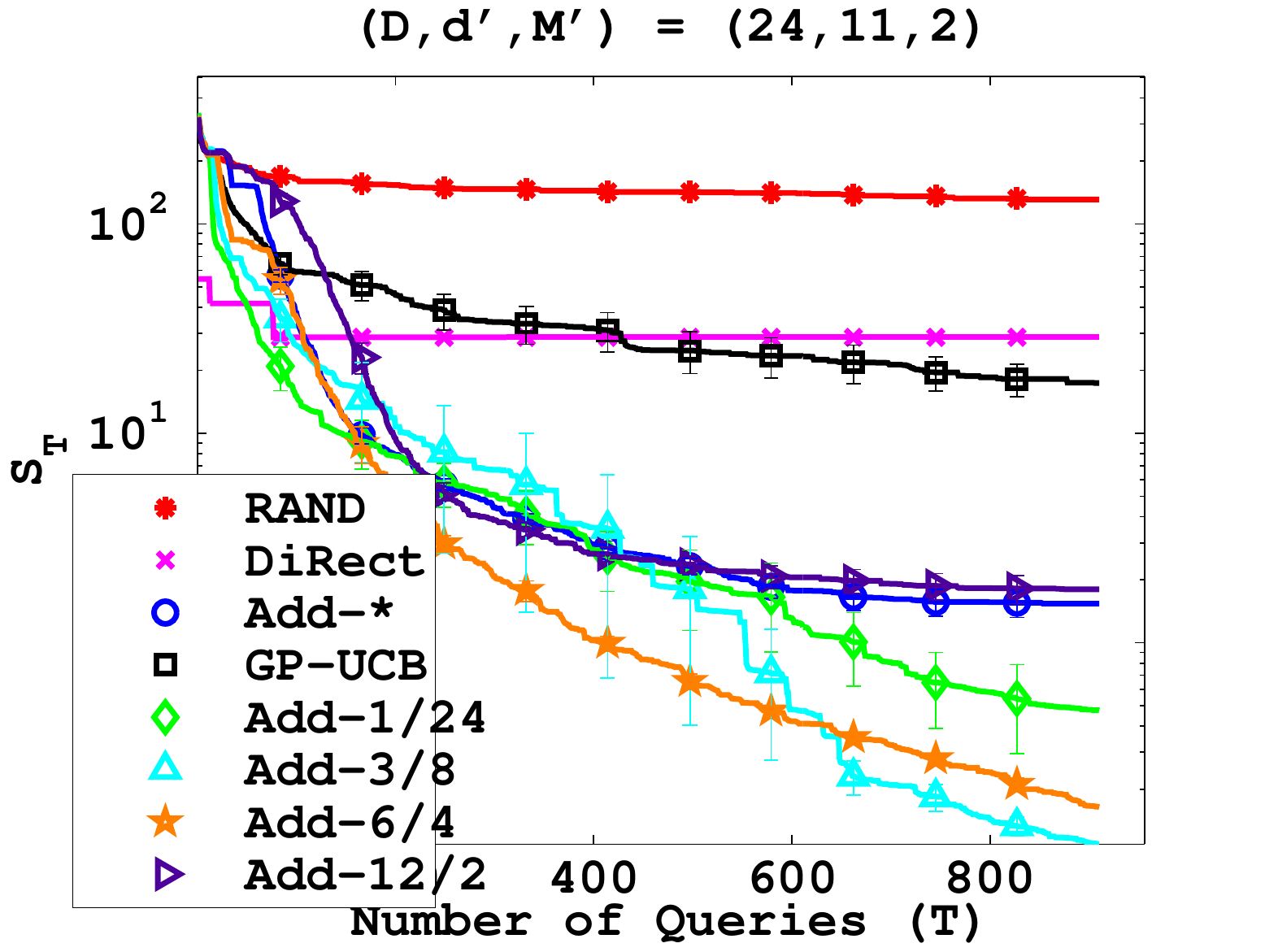}
  \vspace{\imlabelspace}
  \label{fig:SRD40d18}
} \\

\vspace{\imrowspace}
\subfigure[]{
  \includegraphics[width=\imarrwthree]{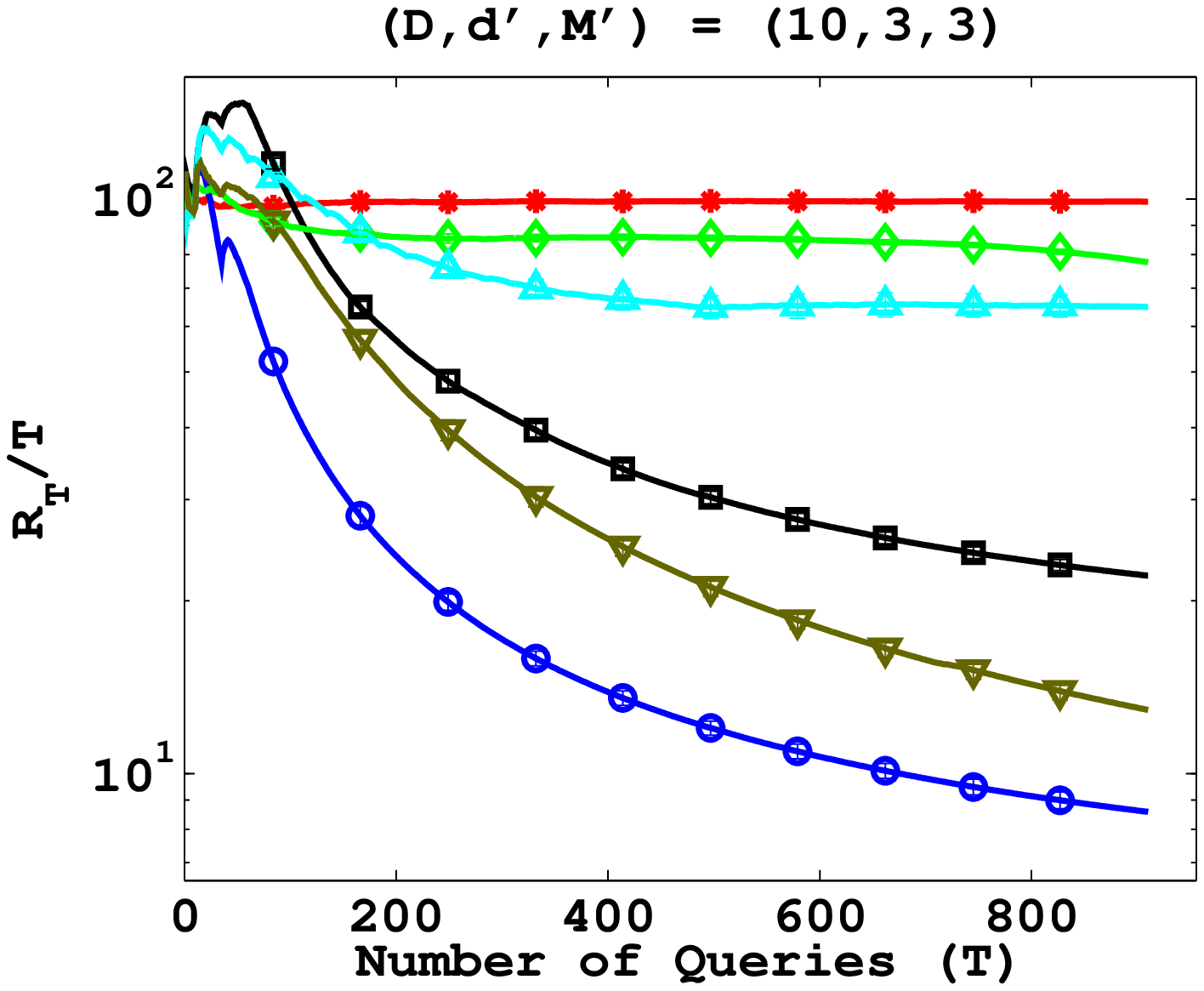} \hspace{\imhspthree}
  \label{fig:CRD10d3}
}
\subfigure[]{
  \includegraphics[width=\imarrwthree]{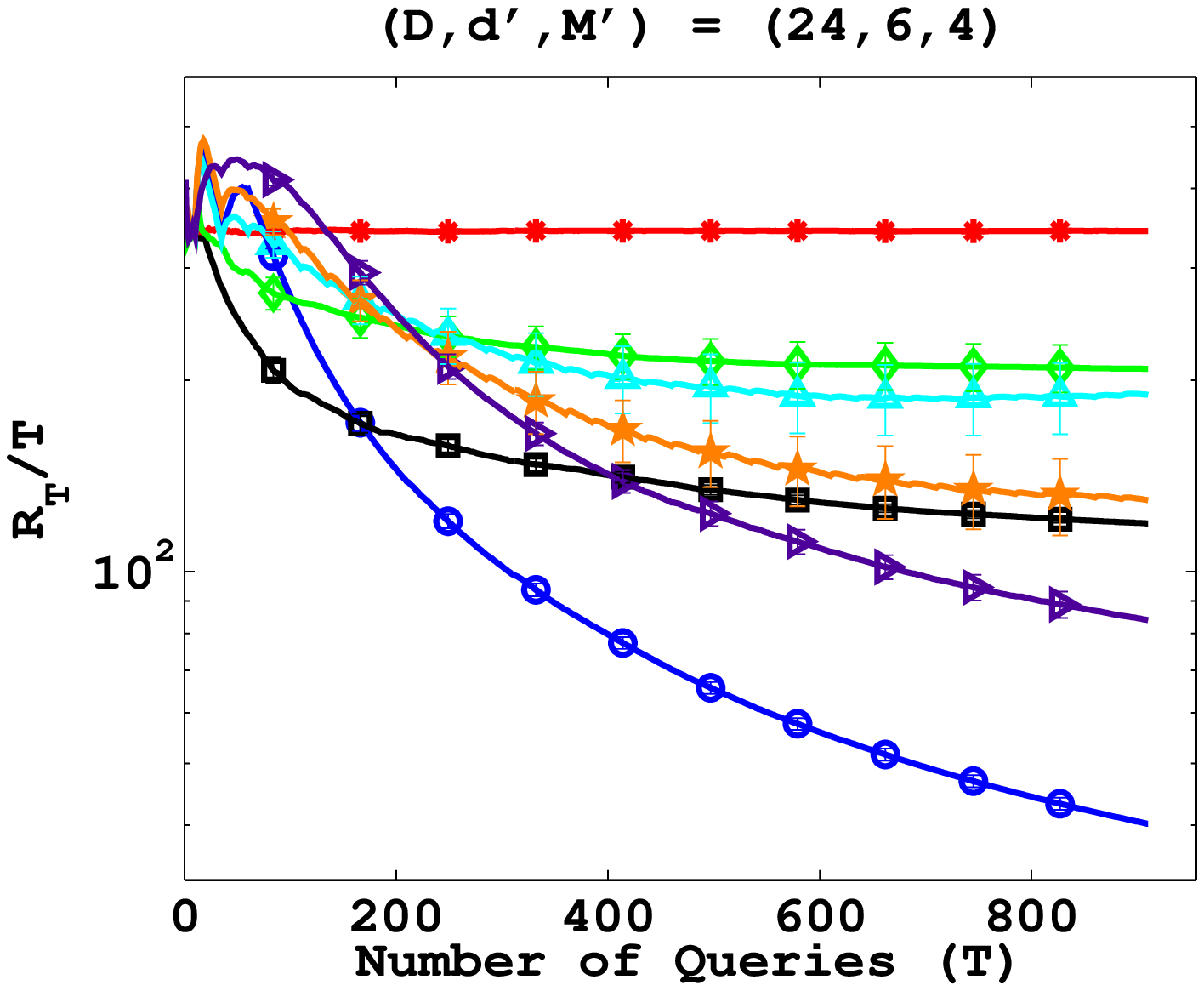} \hspace{\imhspthree}
  \label{fig:CRD24d11}
}
\subfigure[]{
  \includegraphics[width=\imarrwthree]{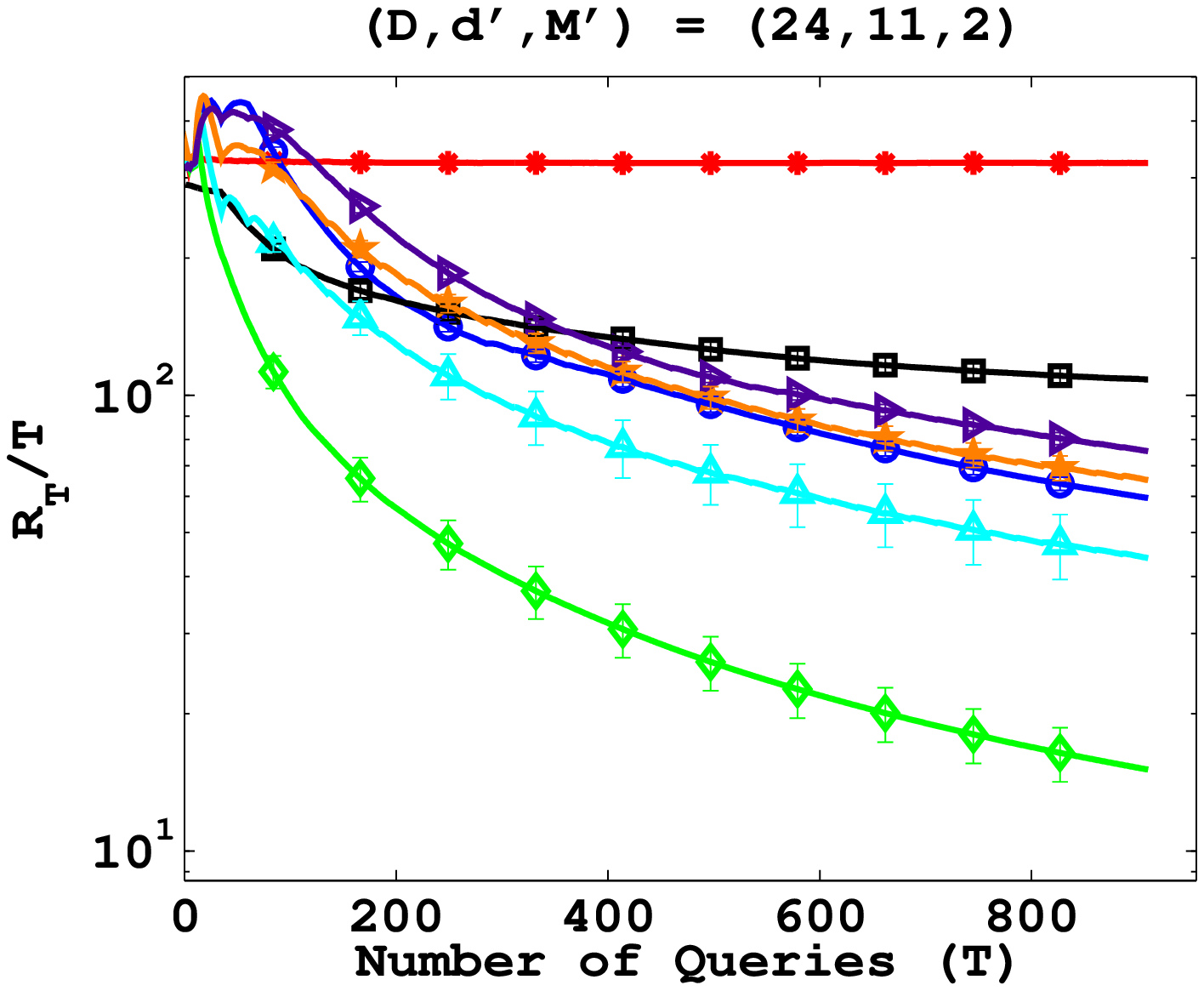}
  \label{fig:CRD40d18}
} \\
\vspace{\imrowspace}

\subfigure[]{
  \includegraphics[width=\imarrwthree]{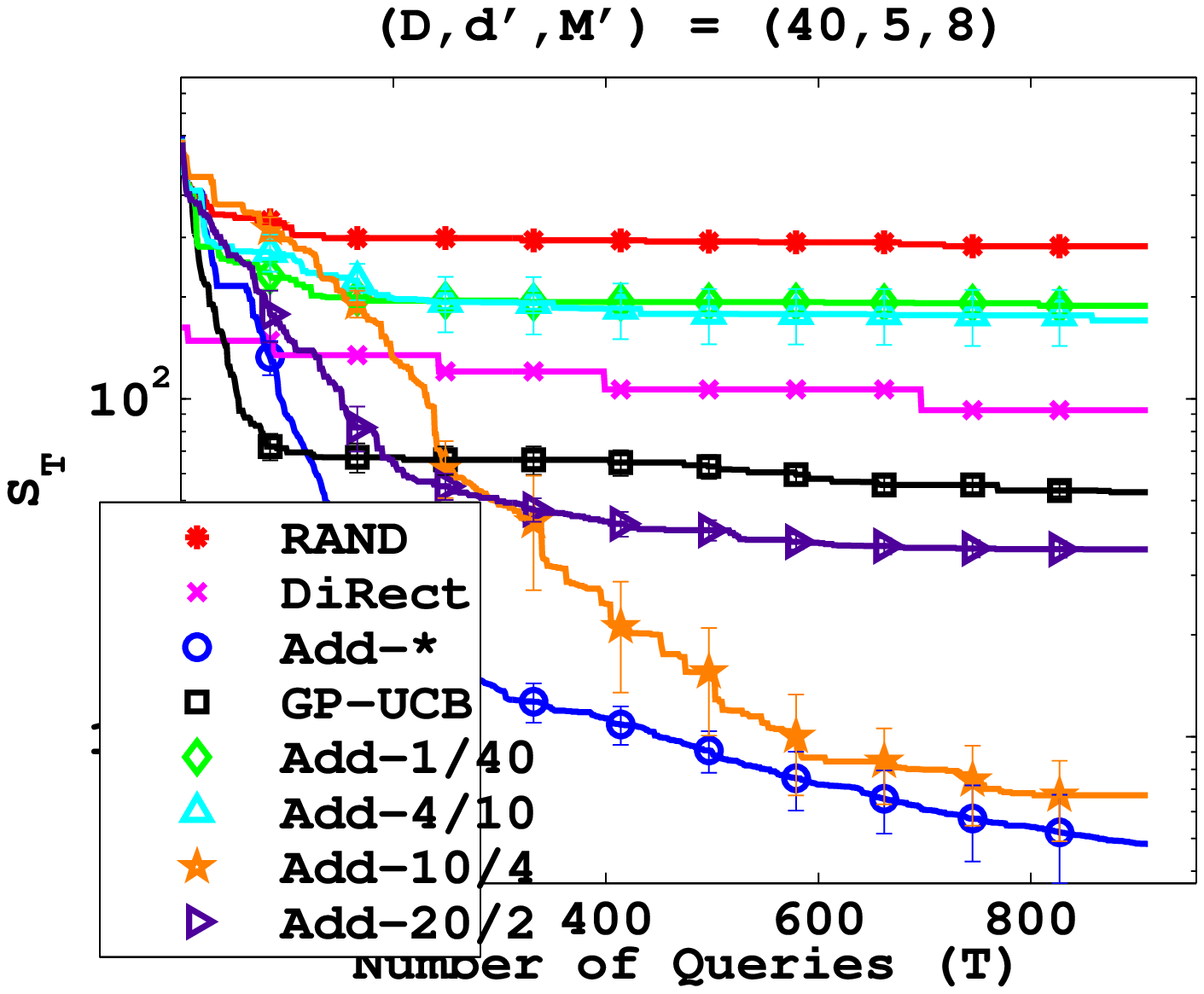} \hspace{\imhspthree}
  \vspace{\imlabelspace}
  \label{fig:SRD40d5}
}
\subfigure[]{
  \includegraphics[width=\imarrwthree]{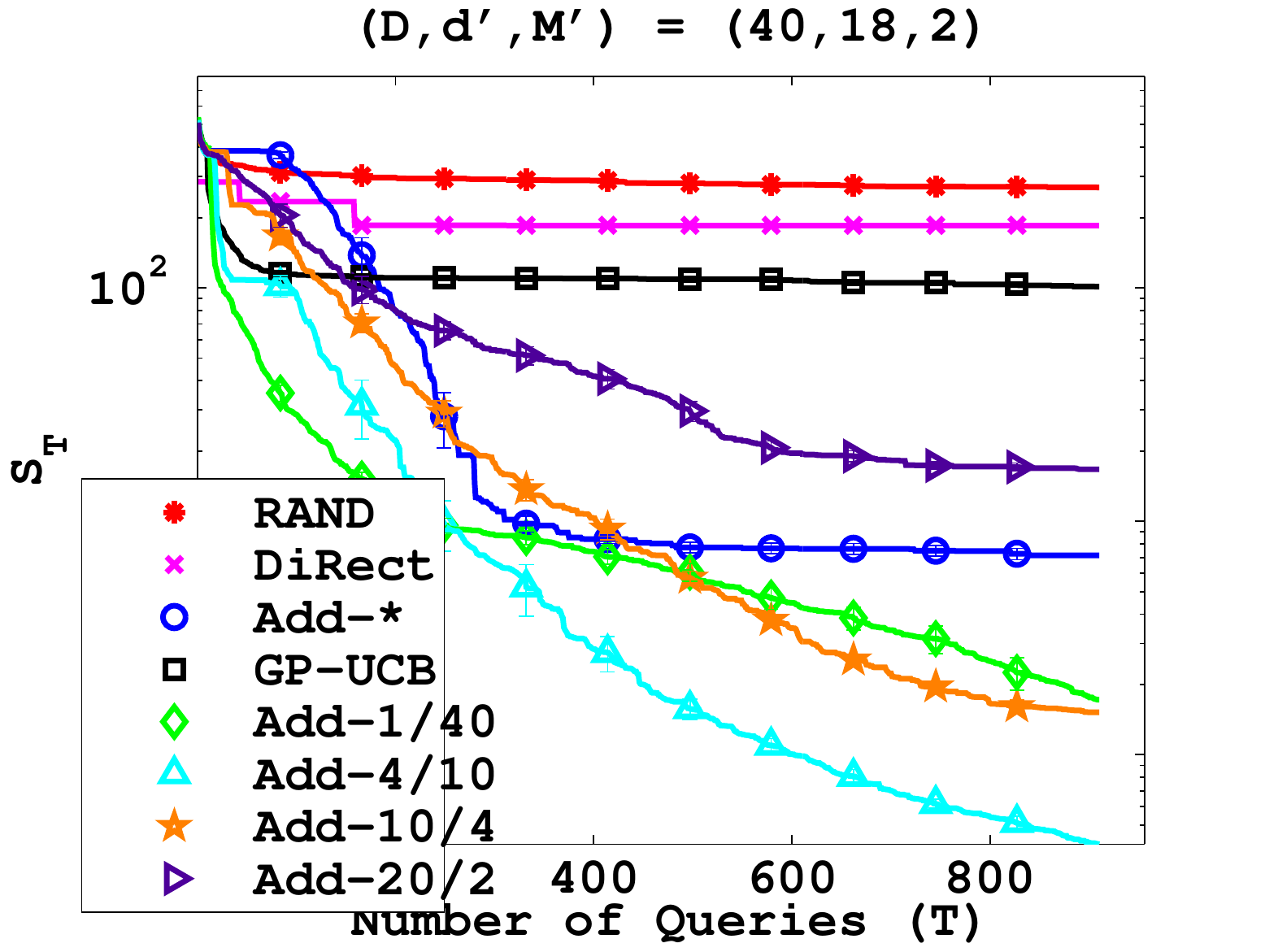} \hspace{\imhspthree}
  \vspace{\imlabelspace}
  \label{fig:SRD96d5}
}
\subfigure[]{
  \includegraphics[width=\imarrwthree]{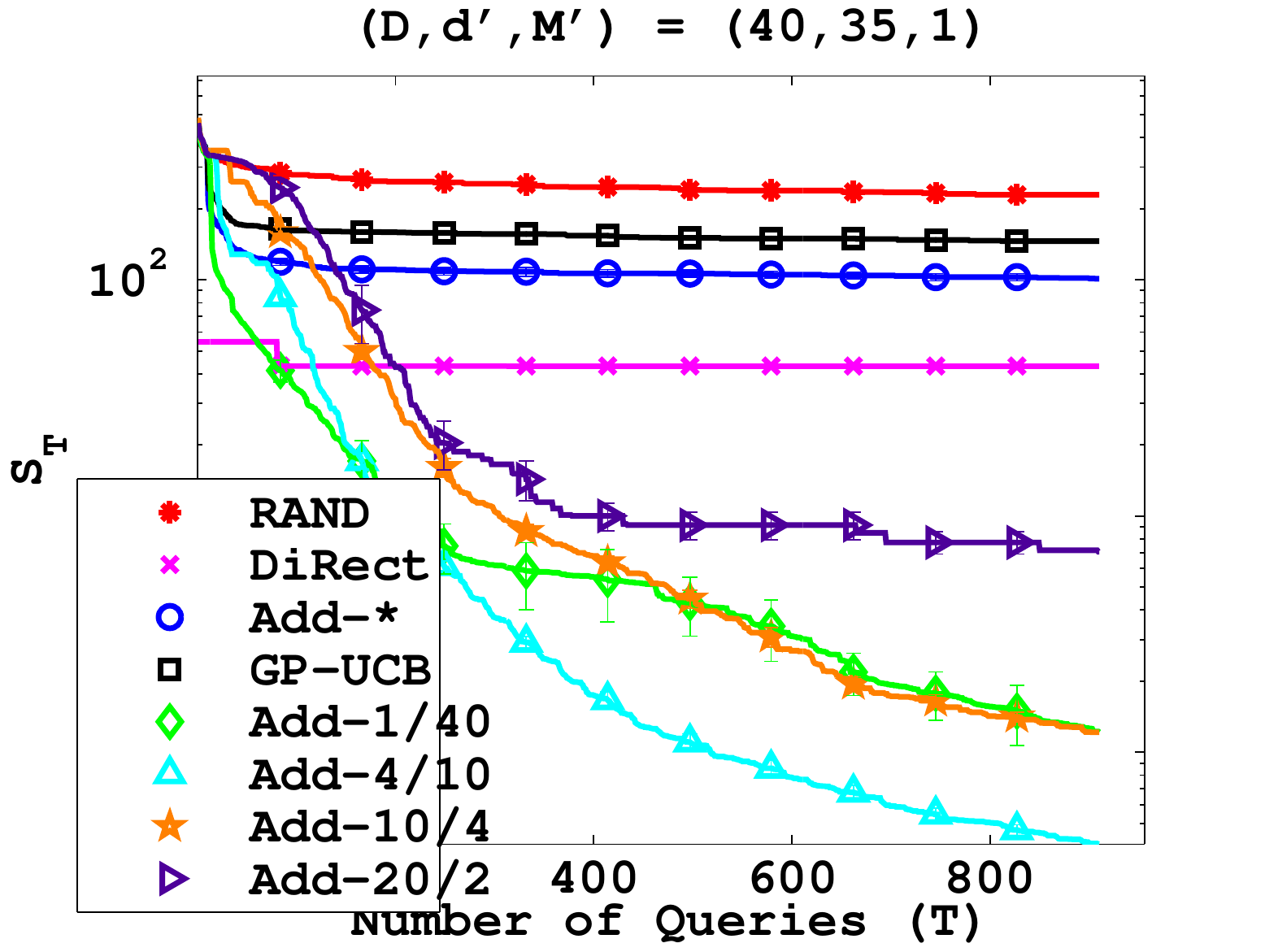}
  \vspace{\imlabelspace}
  \label{fig:SRD41d18}
} \\

\vspace{\imrowspace}
\subfigure[]{
  \includegraphics[width=\imarrwthree]{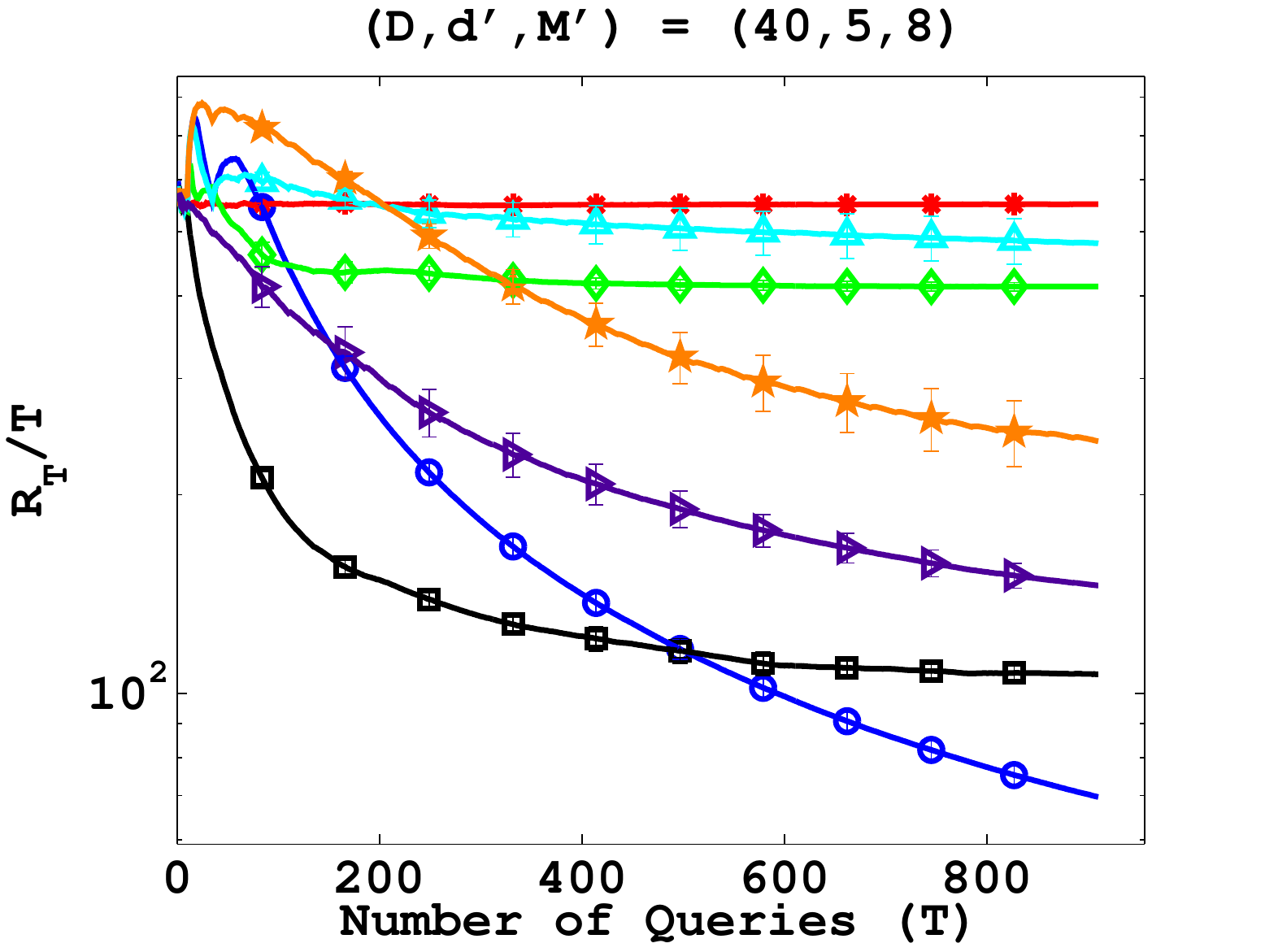} \hspace{\imhspthree}
  \label{fig:CRD40d5}
}
\subfigure[]{
  \includegraphics[width=\imarrwthree]{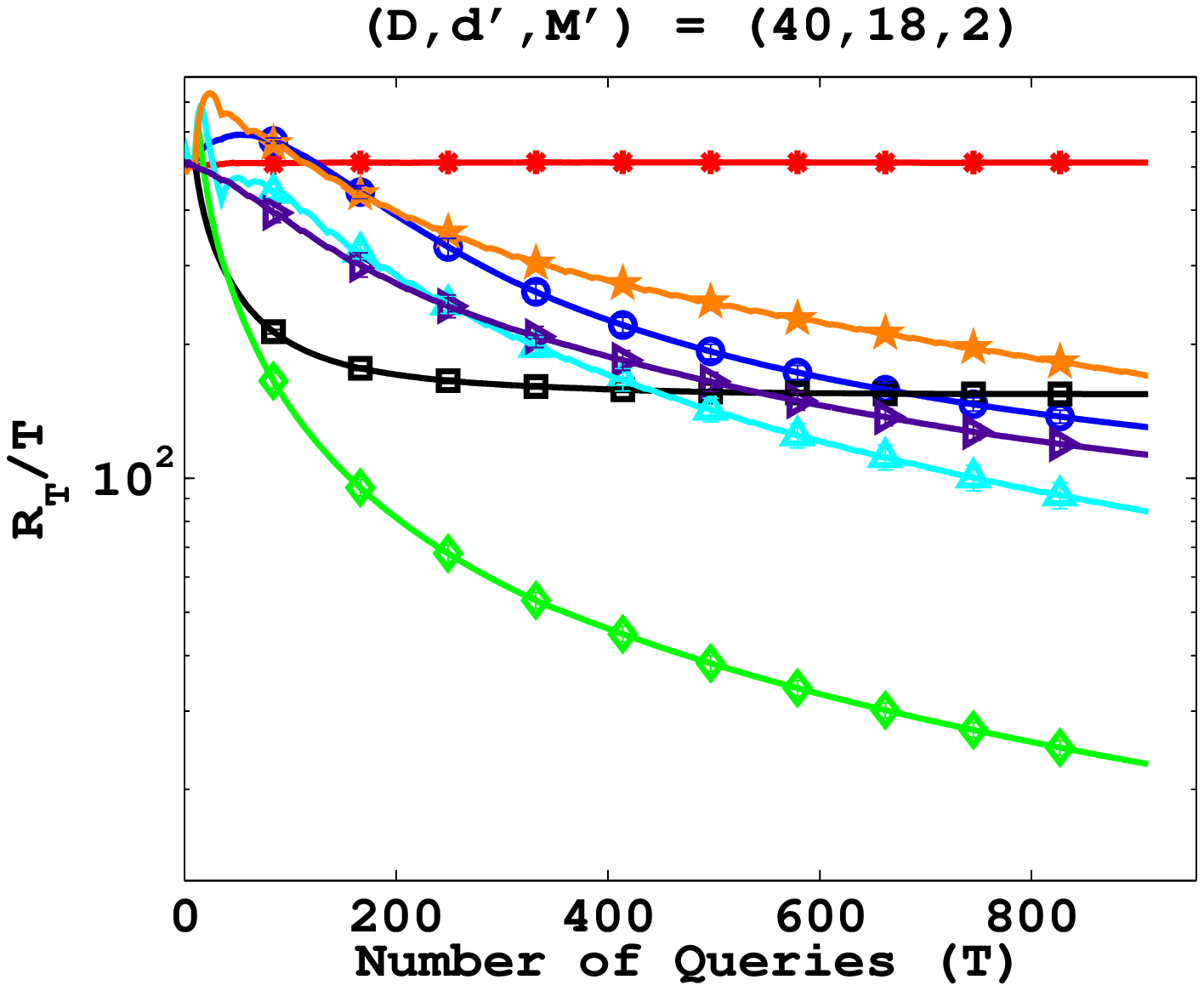} \hspace{\imhspthree}
  \label{fig:CRD96d5}
}
\subfigure[]{
  \includegraphics[width=\imarrwthree]{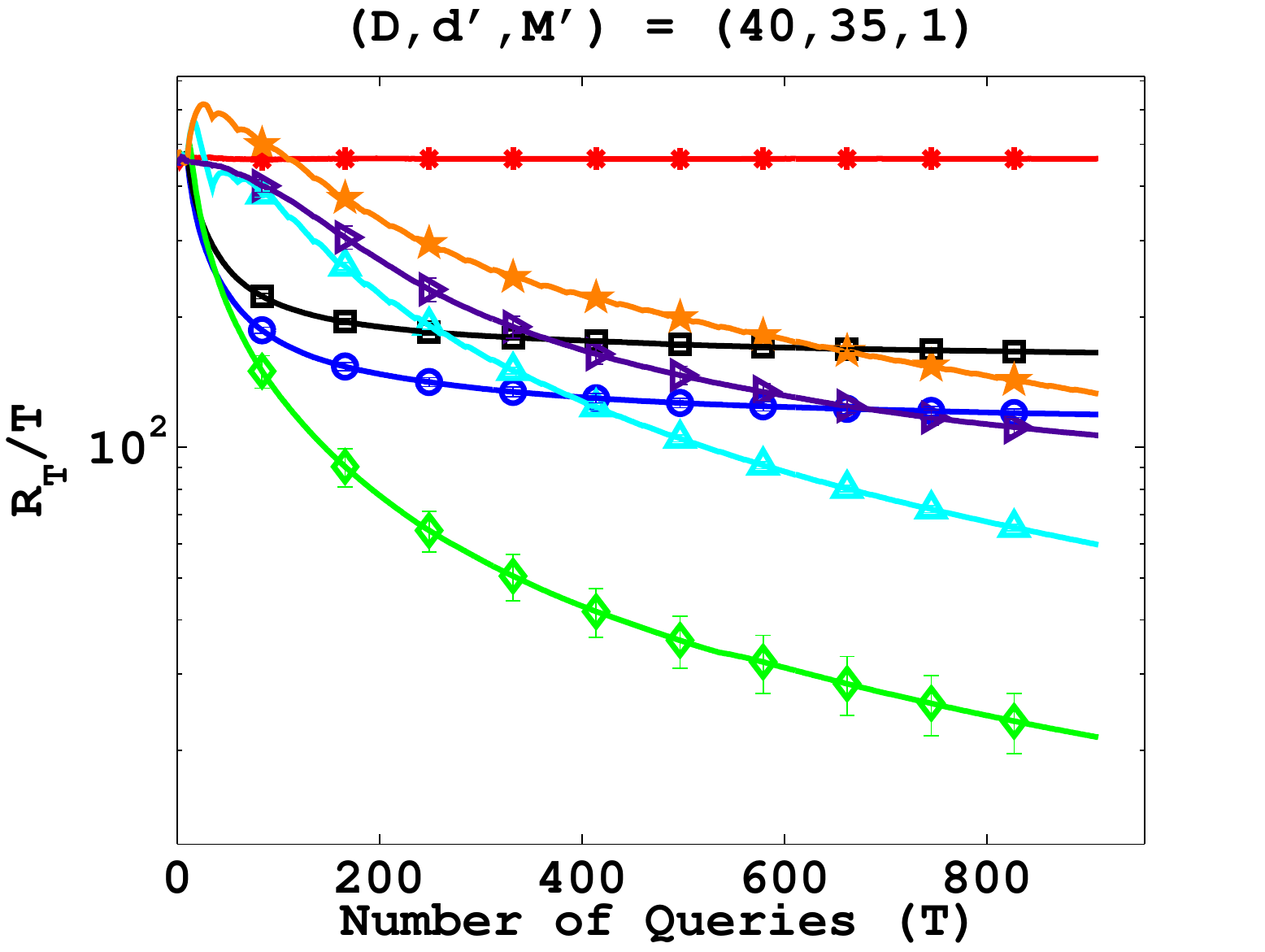}
  \label{fig:CRD41d5}
} \\
\caption[]{ \small
Results on the synthetic datasets. In all images the $x$-axis is the number
of queries and the $y$-axis is the regret in $\log$ scale. 
We have indexed each experiment by their $(D,\dtilde,\Mtilde)$ values.
The first row is $S_T$ for the experiments with $(D,\dtilde,\Mtilde)$ set to
$(10,3,3), (24,6,4), (24,11,2)$ and the second row is $R_T/T$ for the same
experiments. The third row is $S_T$ for $(40,5,8), (40,18,2), (40,35,1)$
and the fourth row is the corresponding $R_T/T$.
In some figures, the error bars are not visible since they are small and hidden
by the bullets. 
All figures were produced  by averaging over $20$ runs.
}
\label{fig:toy}
\vspace{\imtextspace}
\end{figure*}
}

\newcommand{\insertFigureToyResultsThree}{
\begin{figure*}
\centering
\subfigure[]{
  \includegraphics[width=\imarrwthree]{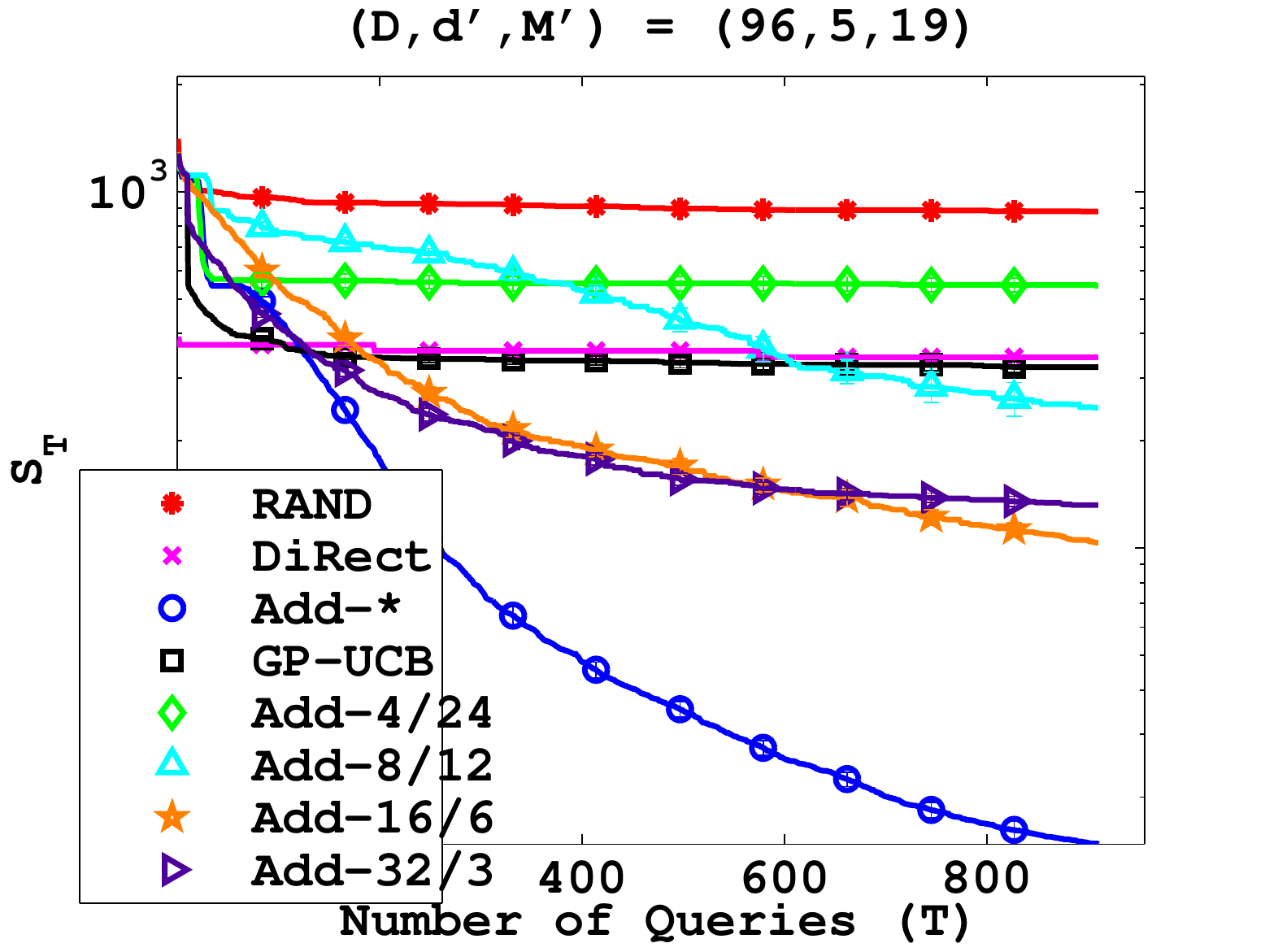} \hspace{\imhspthree}
  \vspace{\imlabelspace}
  \label{fig:SRD96d5}
}
\subfigure[]{
  \includegraphics[width=\imarrwthree]{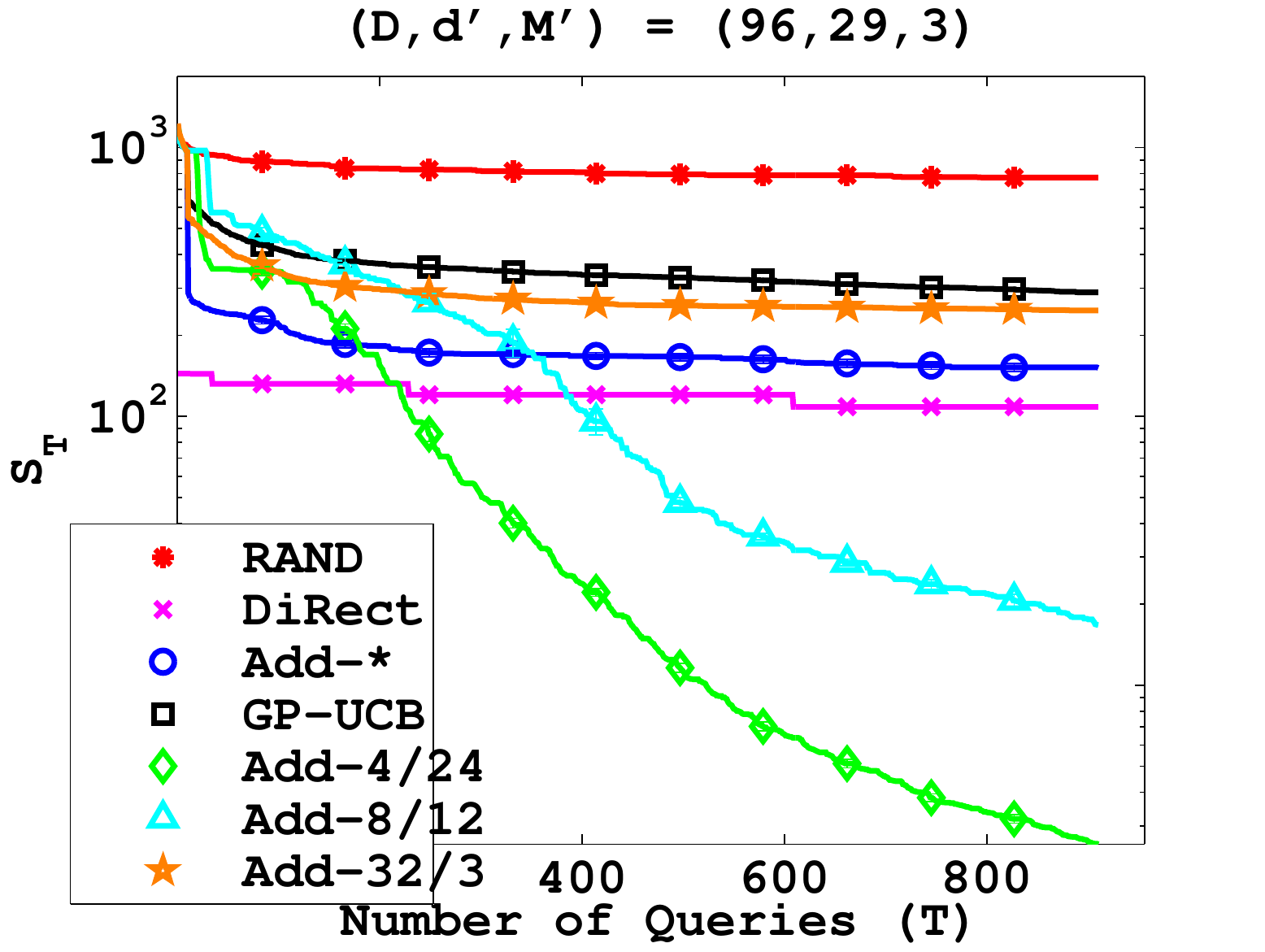} \hspace{\imhspthree}
  \vspace{\imlabelspace}
  \label{fig:SRD96d29}
}
\subfigure[]{
  \includegraphics[width=\imarrwthree]{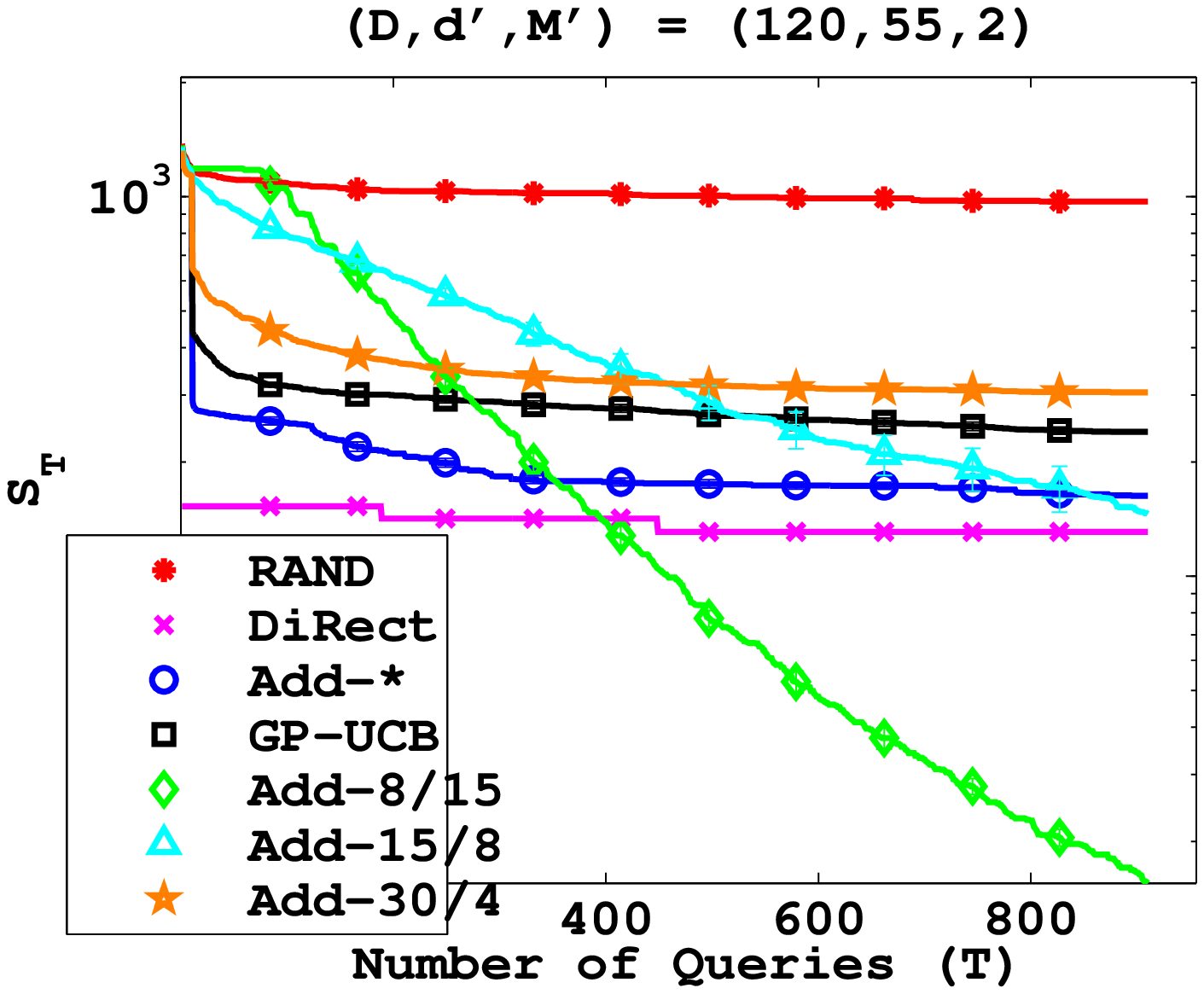}
  \vspace{\imlabelspace}
  \label{fig:SRD120d55}
} \\

\vspace{\imrowspace}
\subfigure[]{
  \includegraphics[width=\imarrwthree]{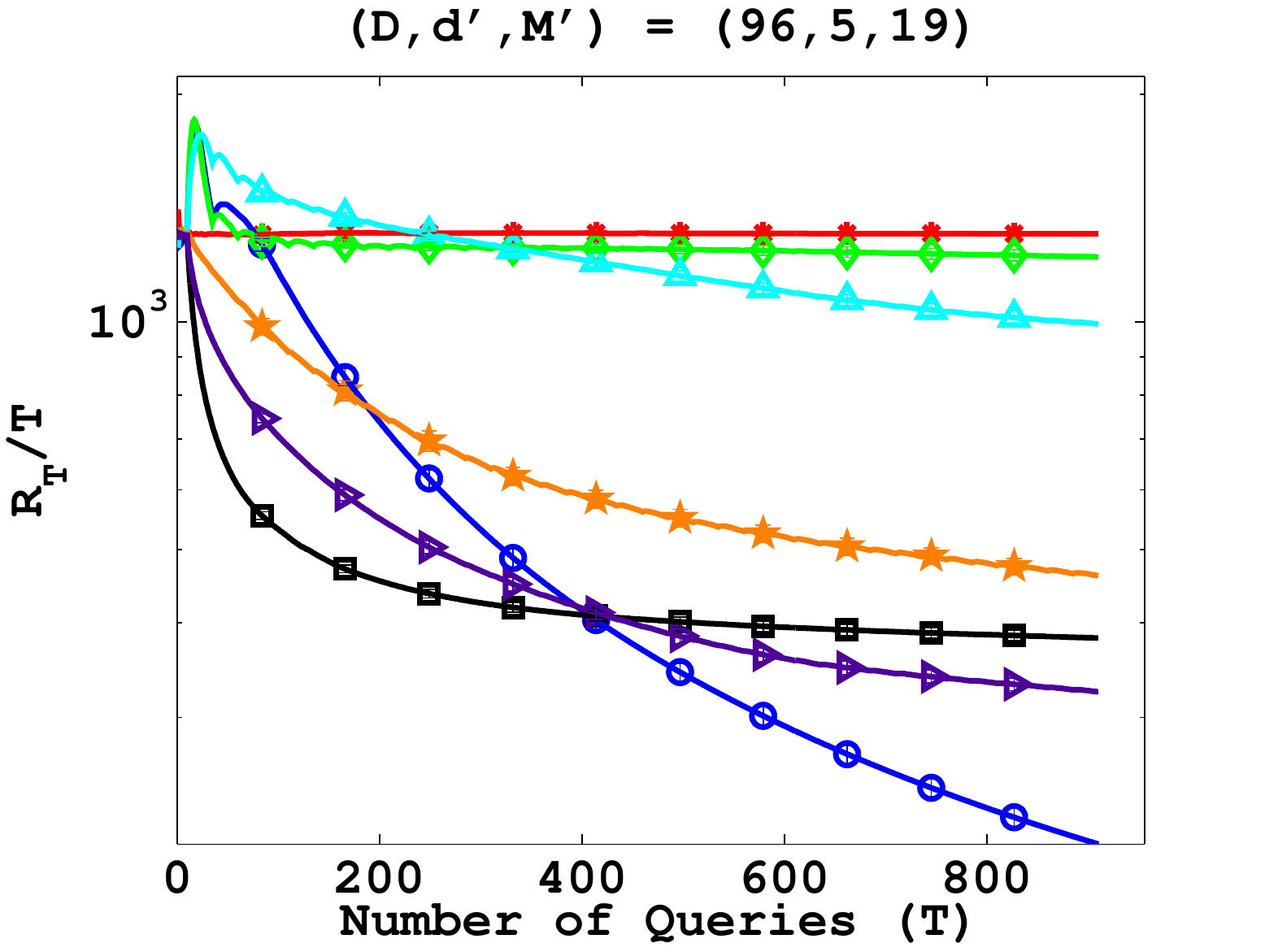} \hspace{\imhspthree}
  \label{fig:CRD96d5}
}
\subfigure[]{
  \includegraphics[width=\imarrwthree]{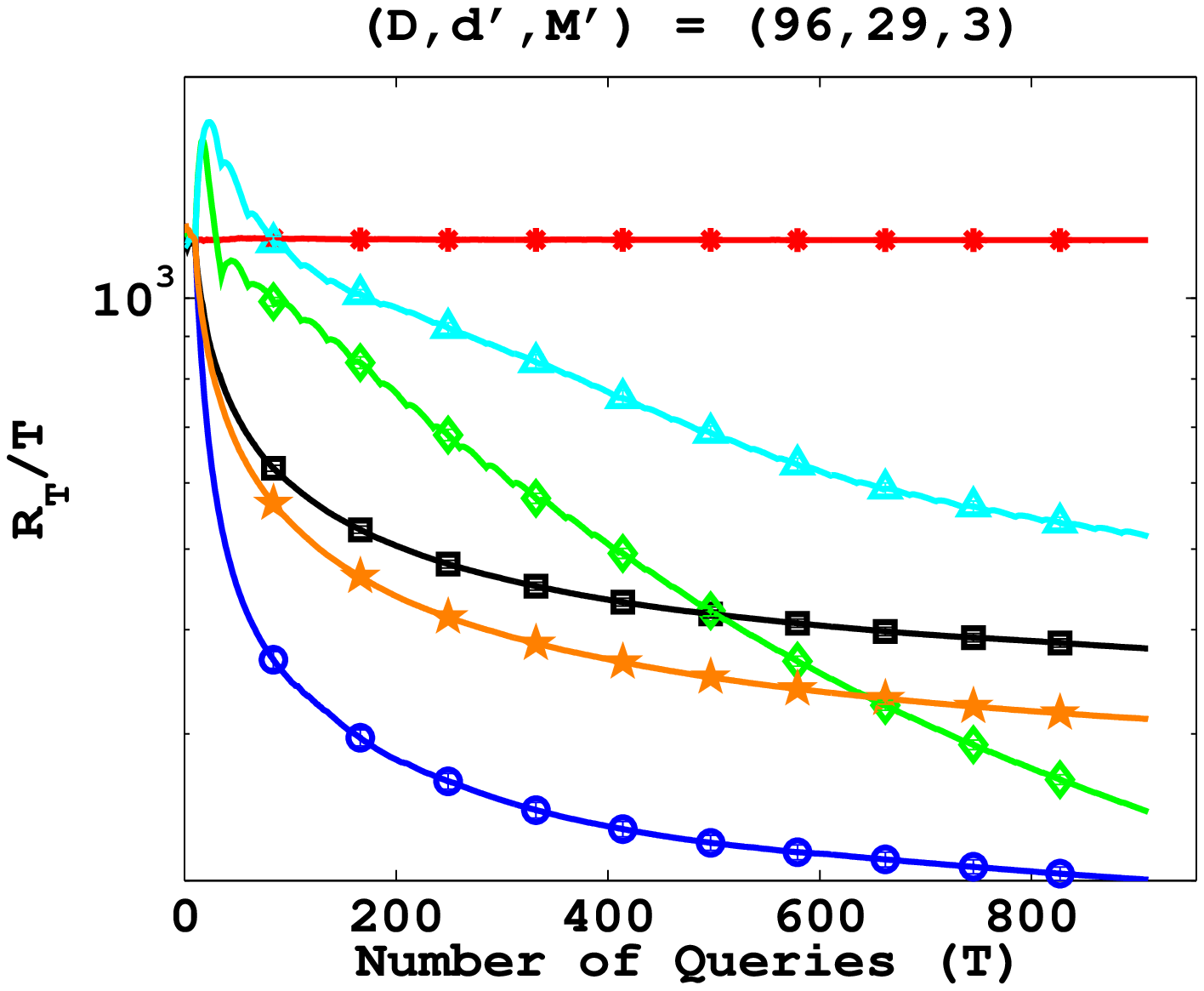} \hspace{\imhspthree}
  \label{fig:CRD96d29}
}
\subfigure[]{
  \includegraphics[width=\imarrwthree]{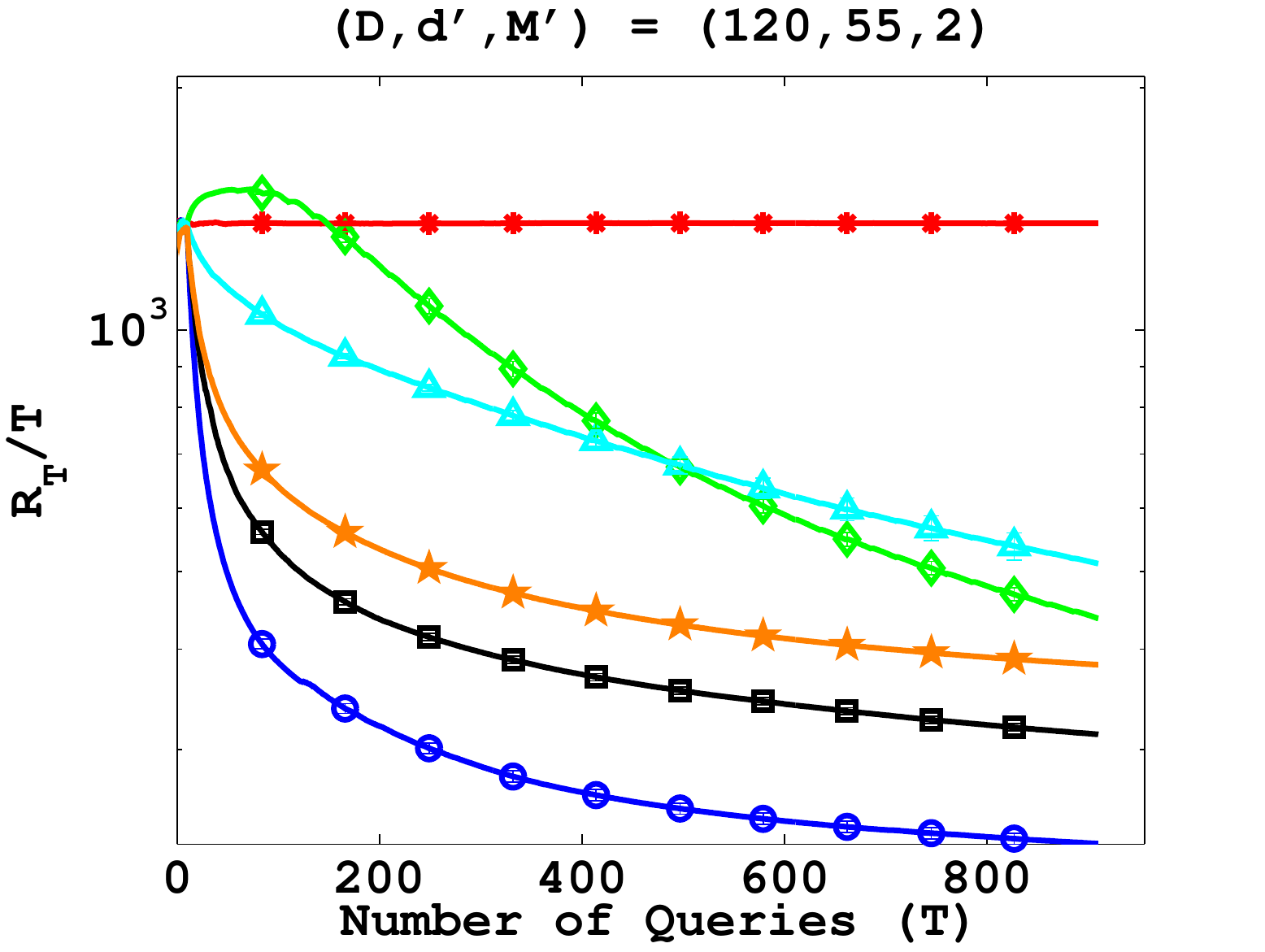}
  \label{fig:CRD120d55}
} \\
\caption[]{ 
More results on synthetic experiments.
The simple regret $S_T$ (first row) and cumulative regret $R_T/T$ (second row) 
for functions with $(D, \dtilde, \Mtilde)$ set to
$(96, 5, 19), (96,29,3), (120,55,2)$ respectively. 
Read the caption under Figure~\ref{fig:toy} for more details.
}
\label{fig:toythree}
\vspace{\imtextspace}
\end{figure*}
}


\newcommand{\insertFigureReal}{
\ifthenelse{\boolean{istwocolumn}}{
\begin{figure}
\centering
\vspace{\imrowspace}
\subfigure[]{
  \includegraphics[width=\imsinglecol]{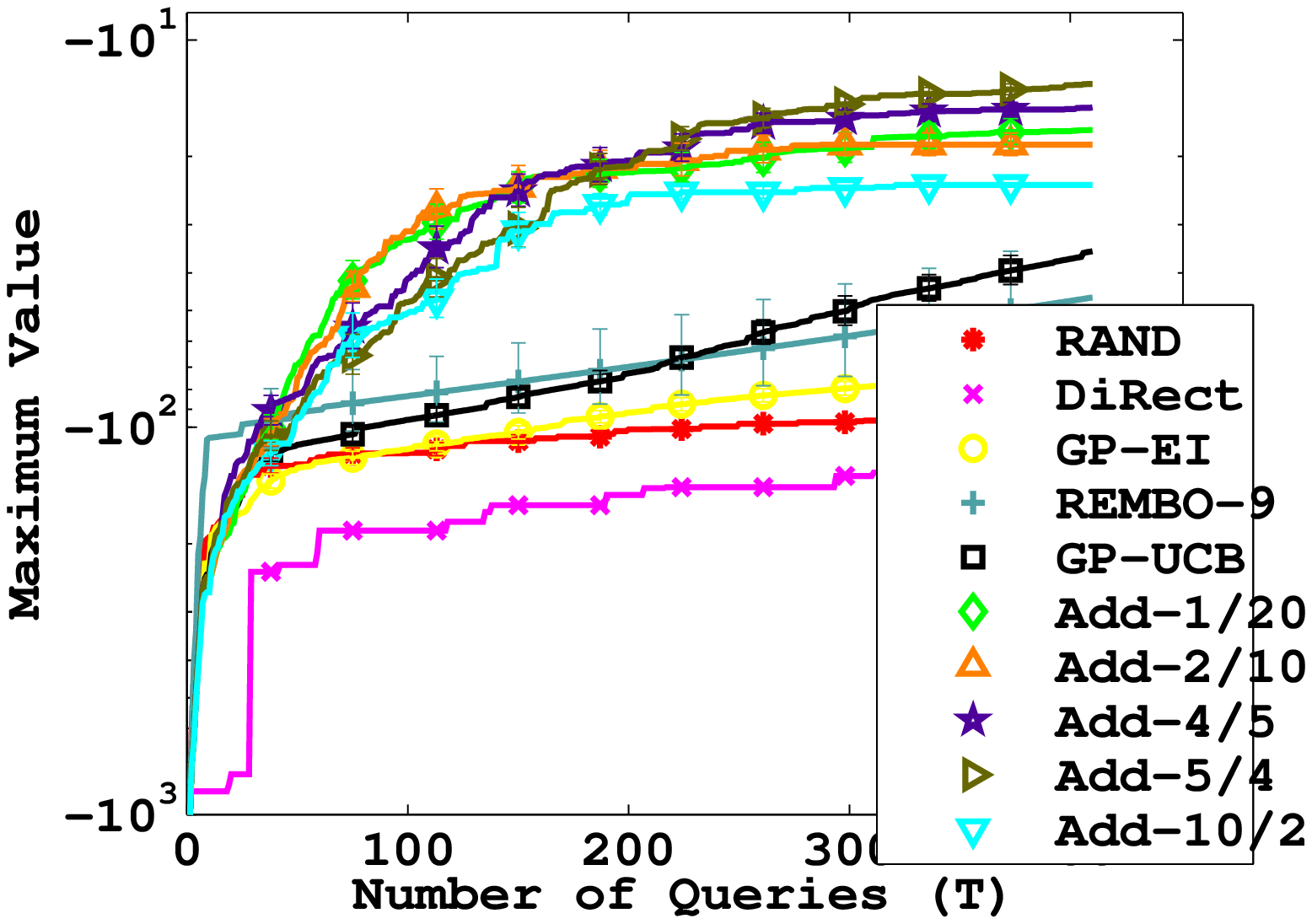} 
  \label{fig:sumlrg}
}
\\[-0.1in]
\subfigure[]{
  \includegraphics[width=\imsinglecol]{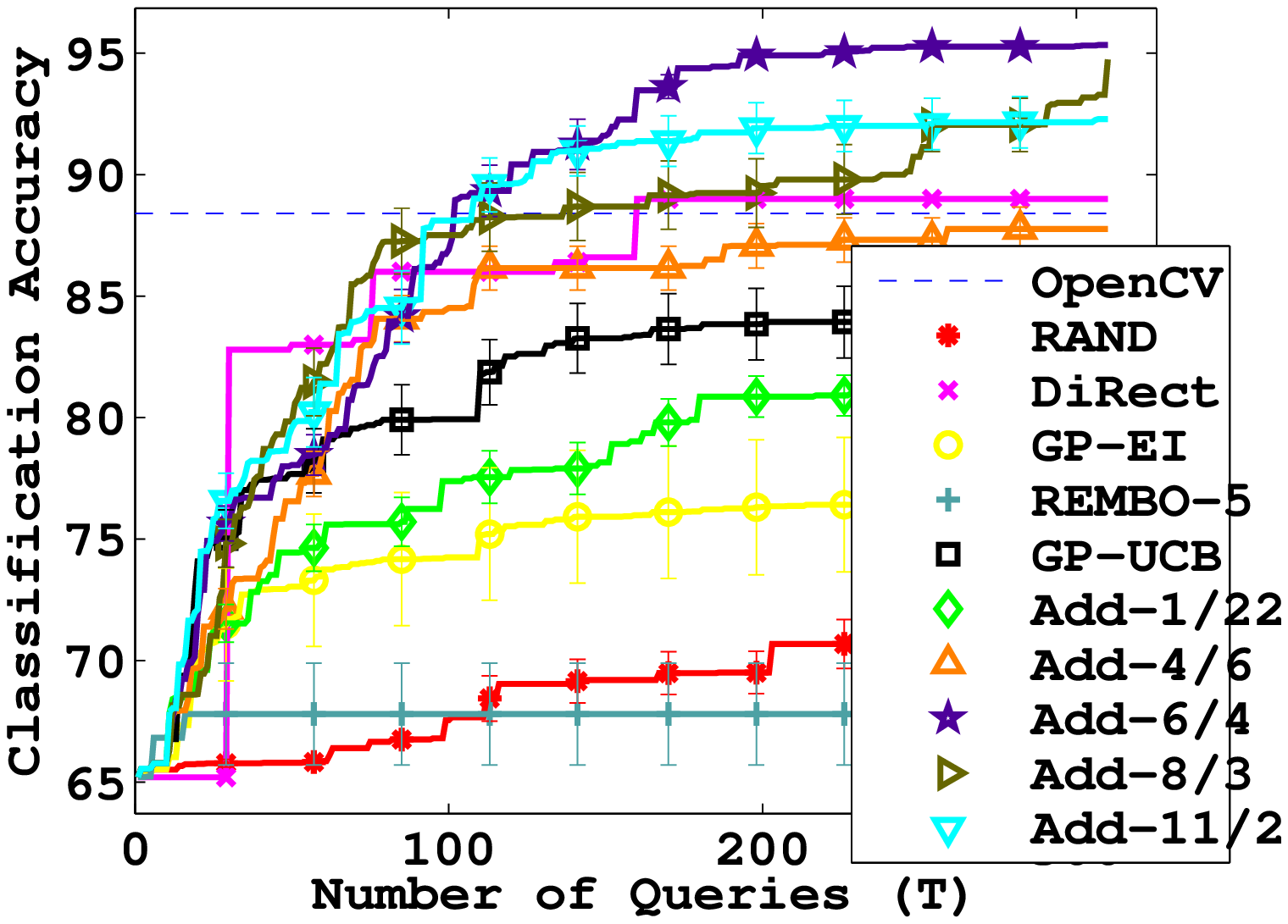} 
  \label{fig:sumvj}
}
\caption[]{ \small
Results on the Astrophysical experiment~\subref{fig:sumlrg} and the Viola and
Jones dataset~\subref{fig:sumvj}. The $x$-axis is the number of queries and the
$y$-axis is the maximum value. 
}
\label{fig:realResultsOneCol}
\vspace{\imtextspace}
\end{figure}
}
{ 
\begin{figure*}
\centering
\subfigure[]{
  \includegraphics[width=\imarrwthree]{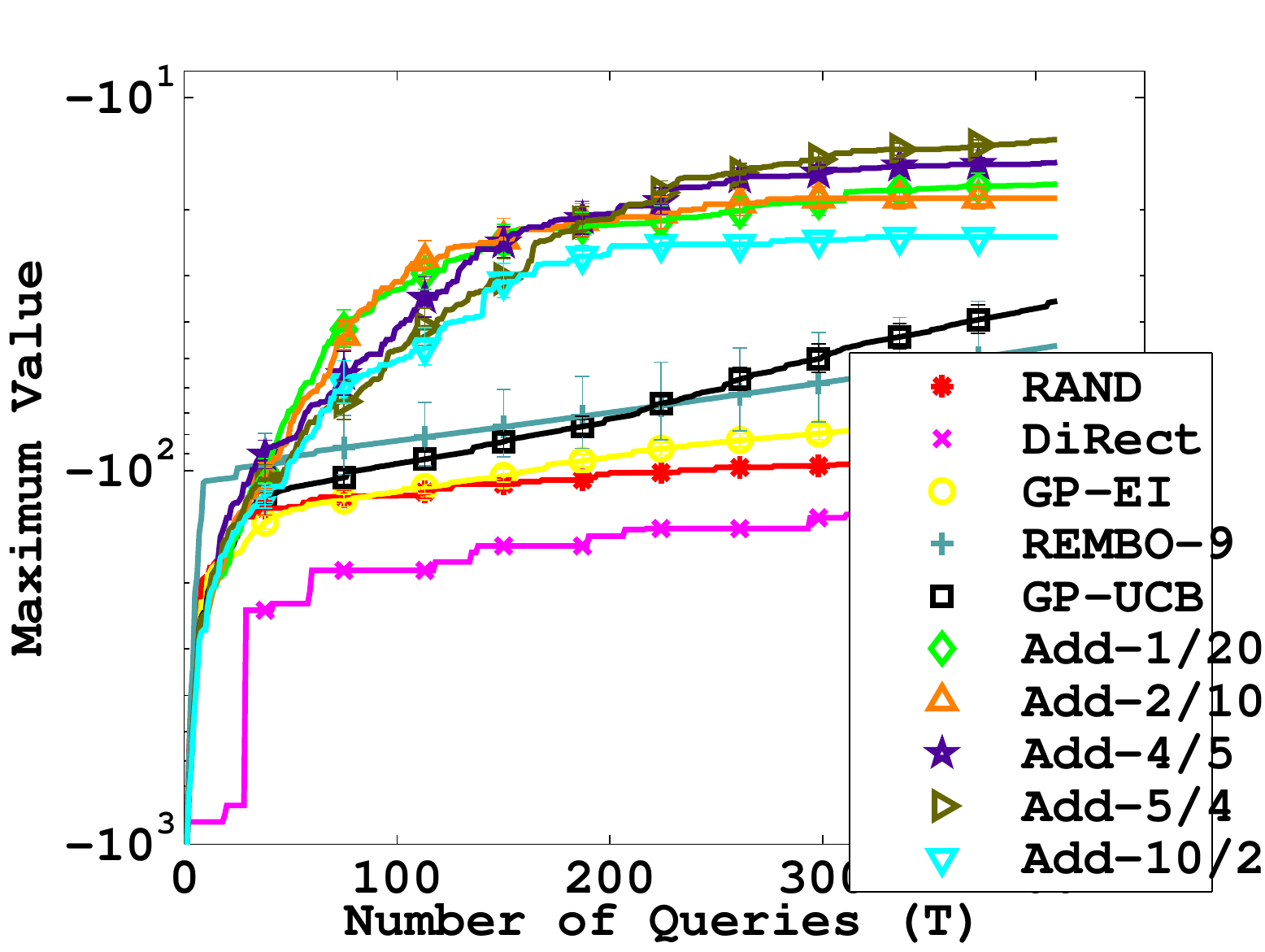} \hspace{\imhsptwo}
  \label{fig:lrg}
}
\subfigure[]{
  \includegraphics[width=\imarrwthree]{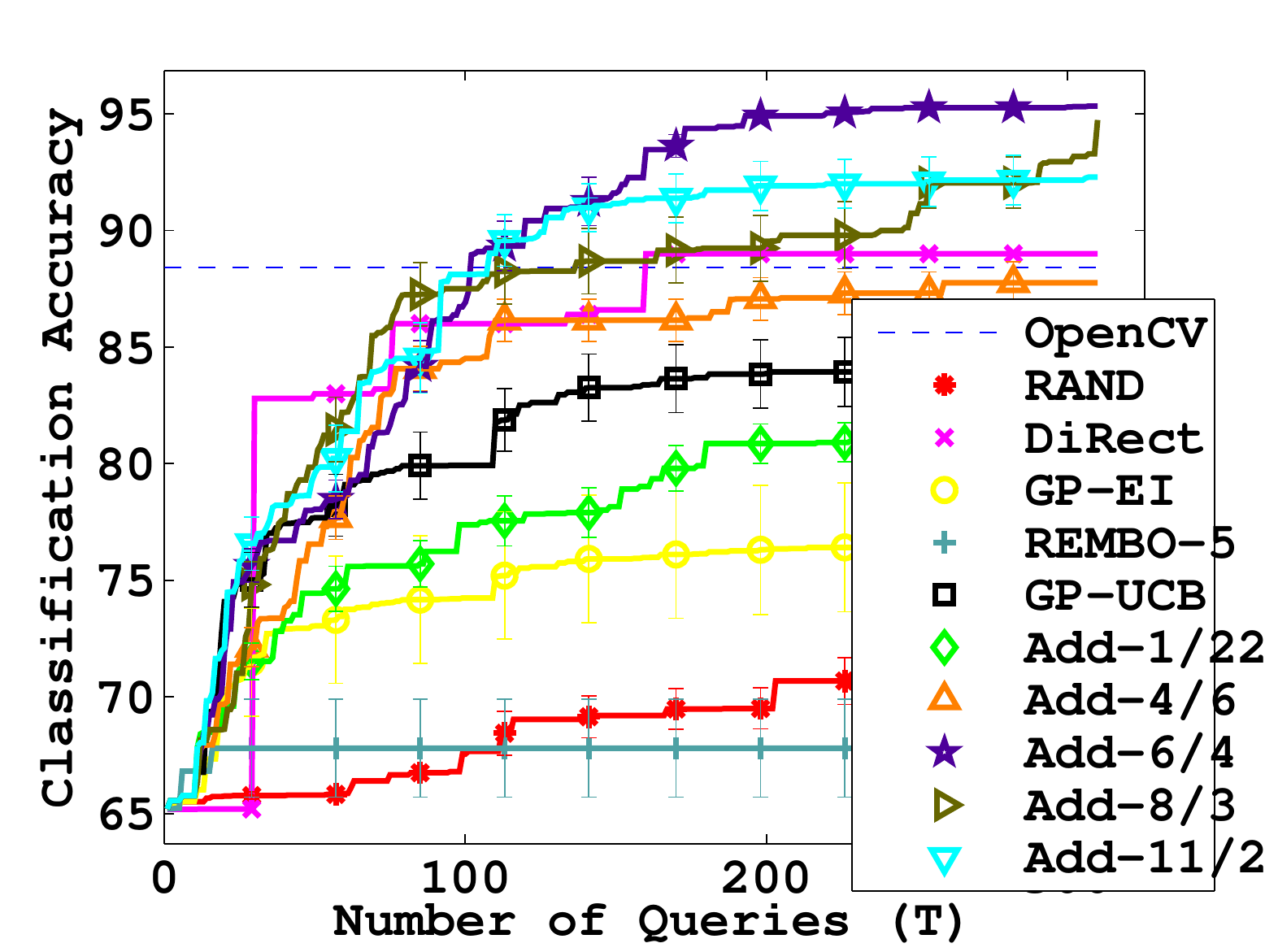} \hspace{\imhsptwo}
  \label{fig:vj}
}
\caption[]{ \small
Results on the Astrophysical experiment~\subref{fig:lrg} and the Viola and
Jones dataset~\subref{fig:vj}. The $x$-axis is the number of queries and the
$y$-axis is the maximum value.
\subref{fig:lrg} was produced by averaging over 20 runs and~\subref{fig:vj} over
11 runs.
 }
\label{fig:realResultsTwoCol}
\vspace{\imtextspace}
\end{figure*}
}
}

\newcommand{\insertFigureSummaryToy}{
\begin{figure*}
\centering
\subfigure{
  \includegraphics[width=\imarrwthreesmall]{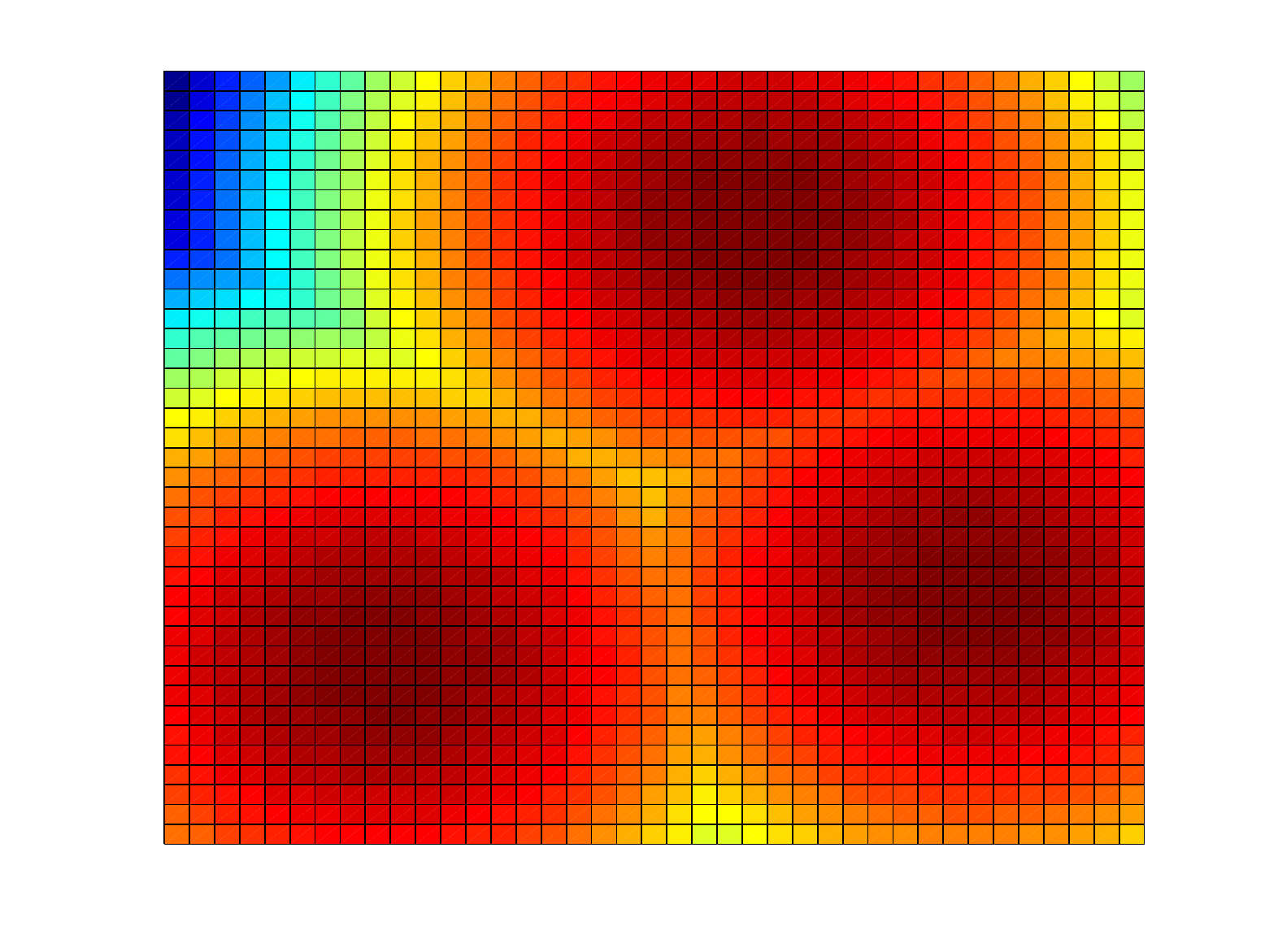} \hspace{\imhspthree}
  \vspace{\imlabelspace}
  \label{fig:func2D}
}
\subfigure{
  \includegraphics[width=\imarrwthreesmall]{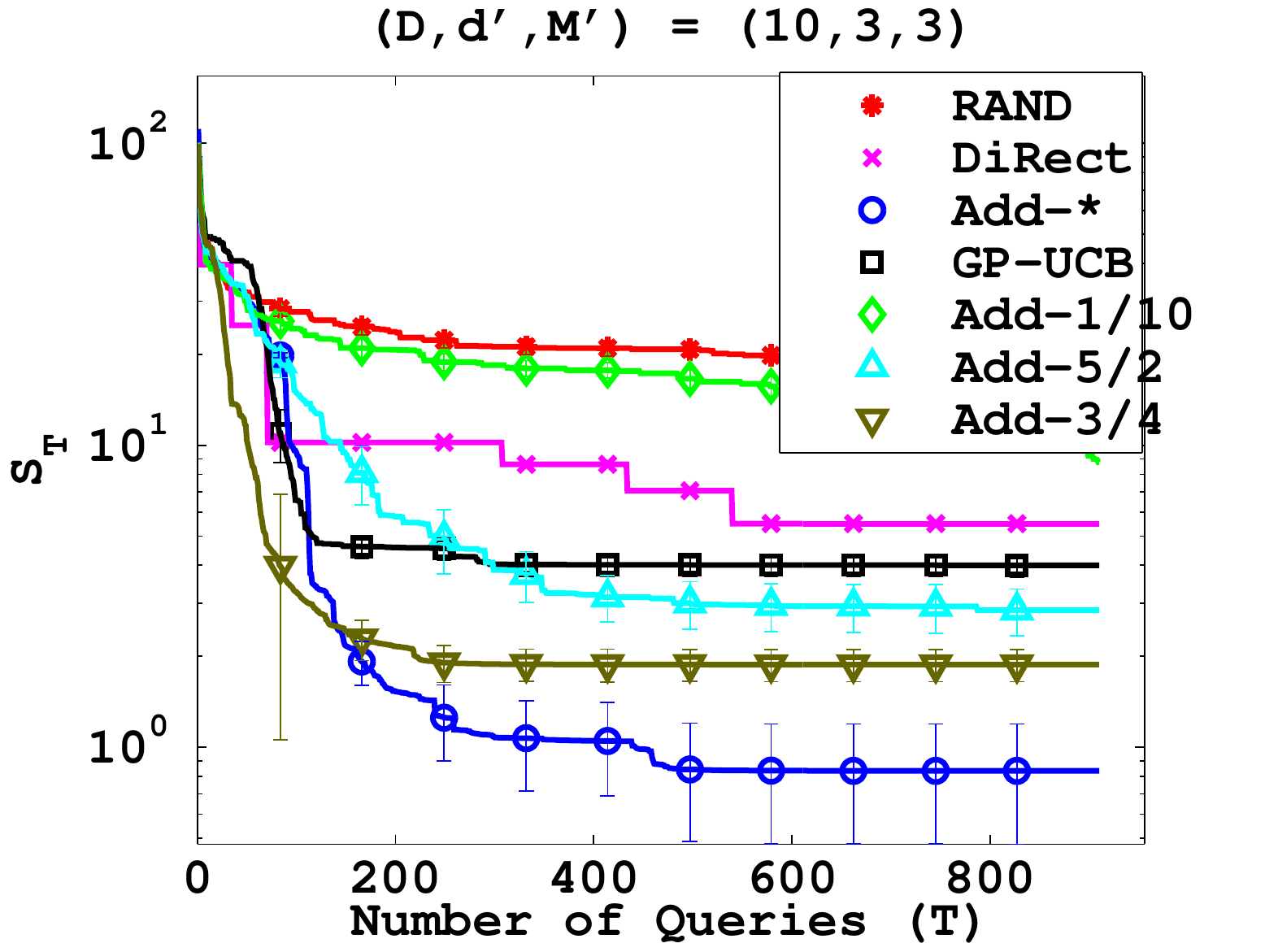} \hspace{\imhspthree}
  \vspace{\imlabelspace}
  \label{fig:sumSRD10d3}
}
\subfigure{
  \includegraphics[width=\imarrwthreesmall]{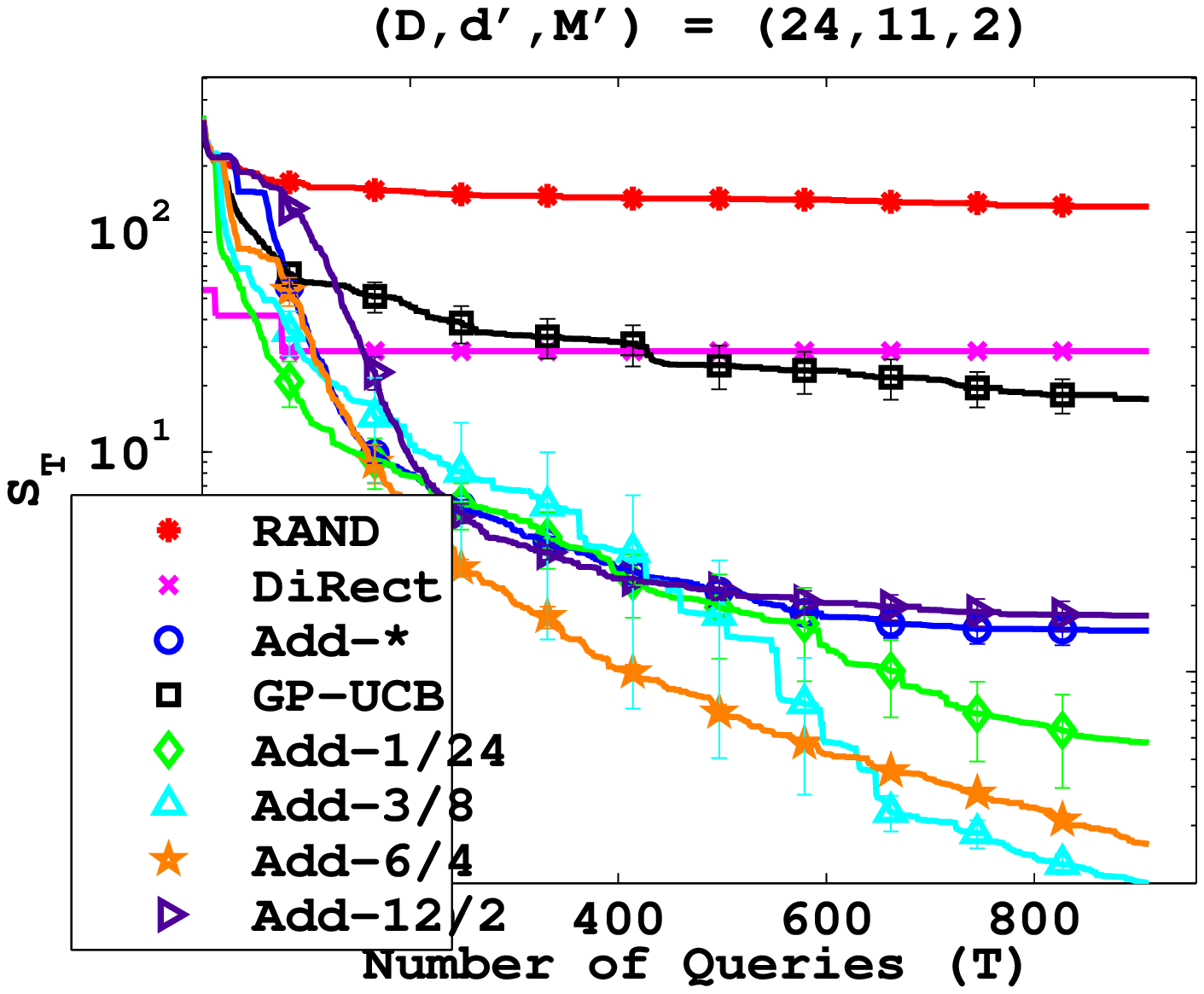} \hspace{\imhspthree}
  \vspace{\imlabelspace}
  \label{fig:sumSRD24d11}
} \\

\subfigure{
  \includegraphics[width=\imarrwthreesmall]{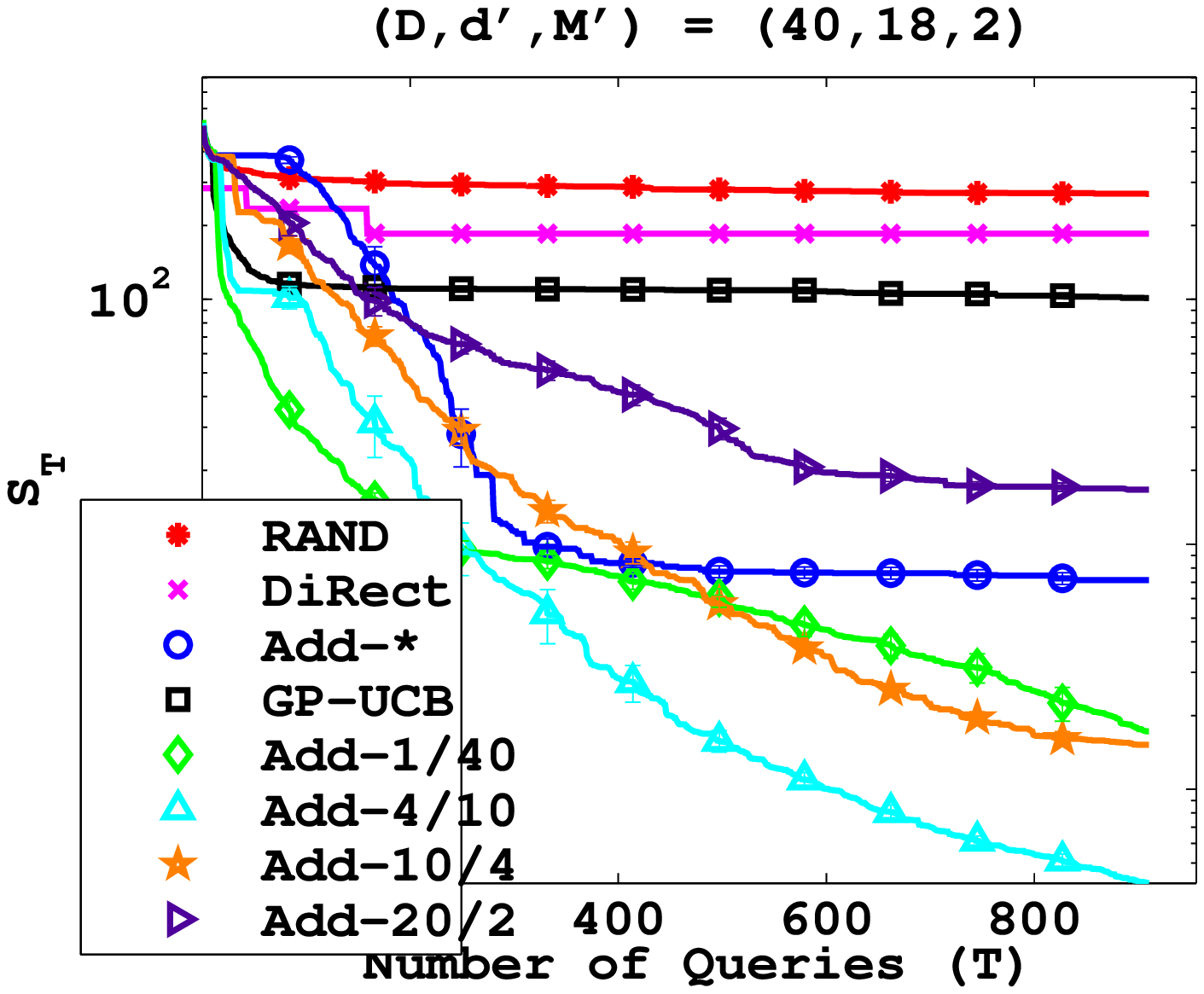}
  \vspace{\imlabelspace}
  \label{fig:sumSRD40d18}
} 
\subfigure{
  \includegraphics[width=\imarrwthreesmall]{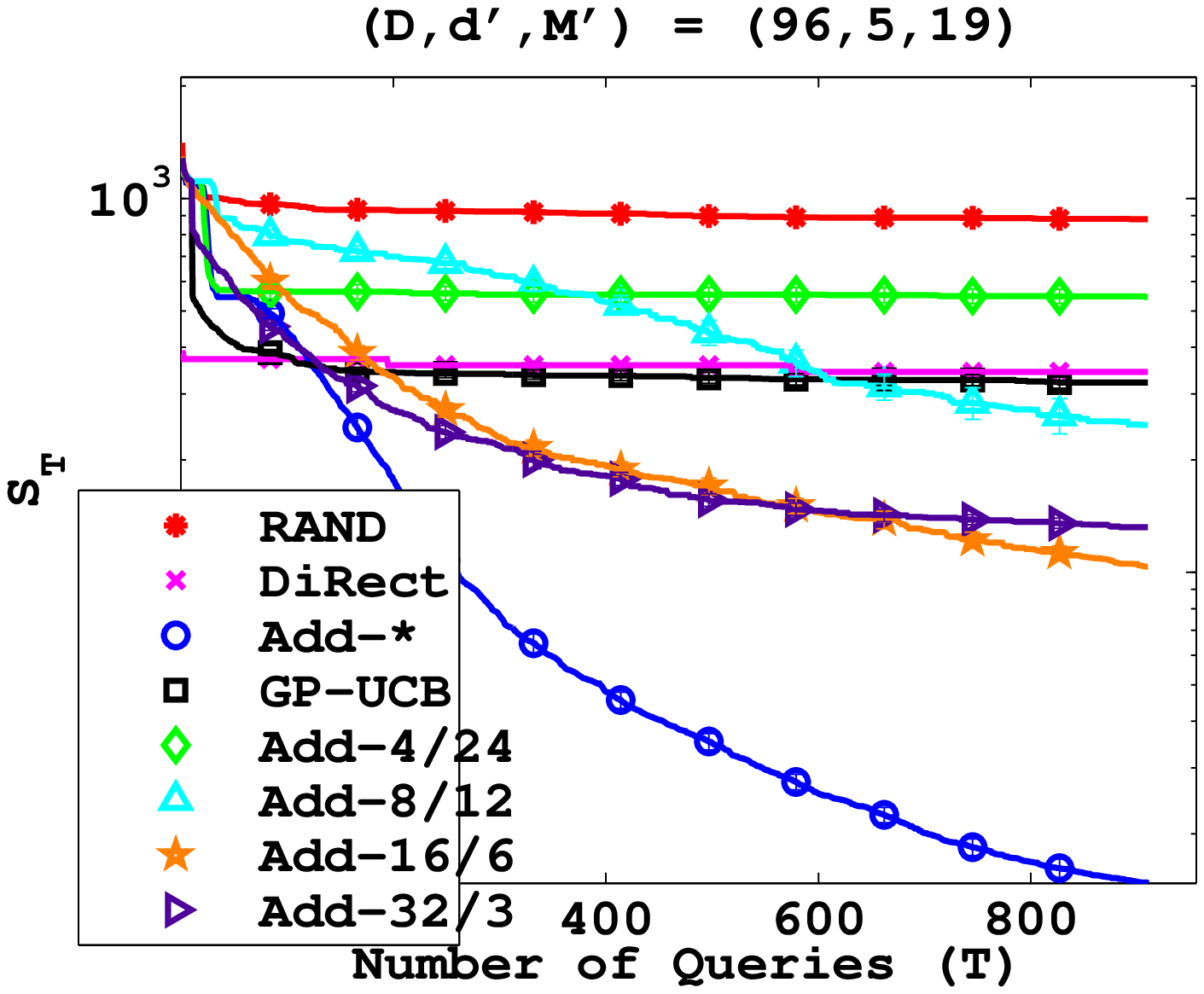} \hspace{\imhspthree}
  \vspace{\imlabelspace}
  \label{fig:sumSRD96d5}
}
\subfigure{
  \includegraphics[width=\imarrwthreesmall]{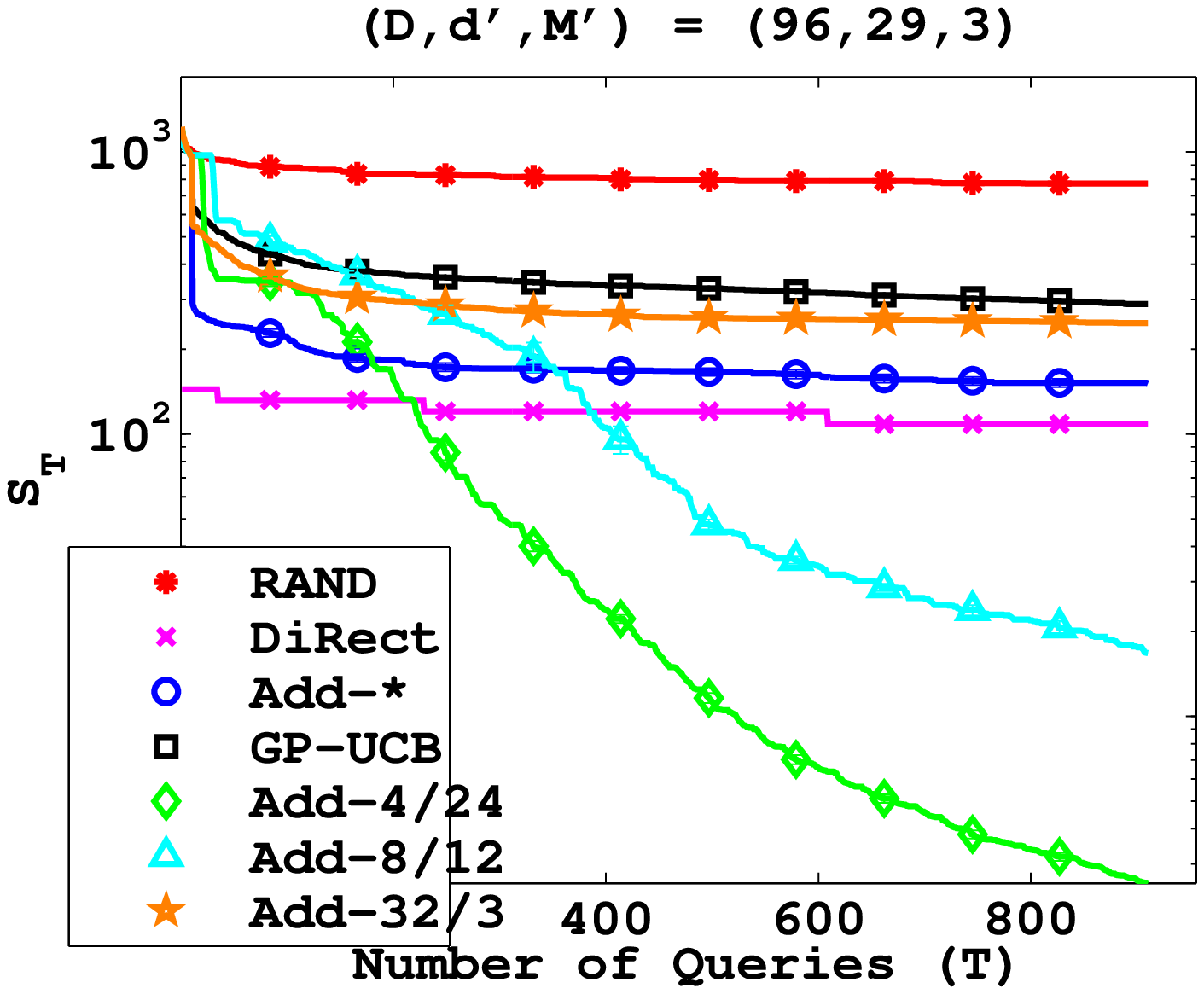}
  \vspace{\imlabelspace}
  \label{fig:sumD96d29}
} \\

\caption[]{ \small
Results on the synthetic experiments. 
The $x$-axis is the number
of queries and the $y$-axis is the regret in $\log$ scale. 
We have indexed the experiments by their $(D,\dtilde,\Mtilde)$ values.
In some figures, the error bars are not visible since they are small and hidden
by the bullets. 
All figures were produced  by averaging over $20$ runs.
}
\label{fig:summaryToyResults}
\vspace{\imtextspace}
\end{figure*}
}

\newcommand{\insertFigureFuncTwoD}{
\begin{figure}
\centering
  \includegraphics[width=\imarrwthreesmall]{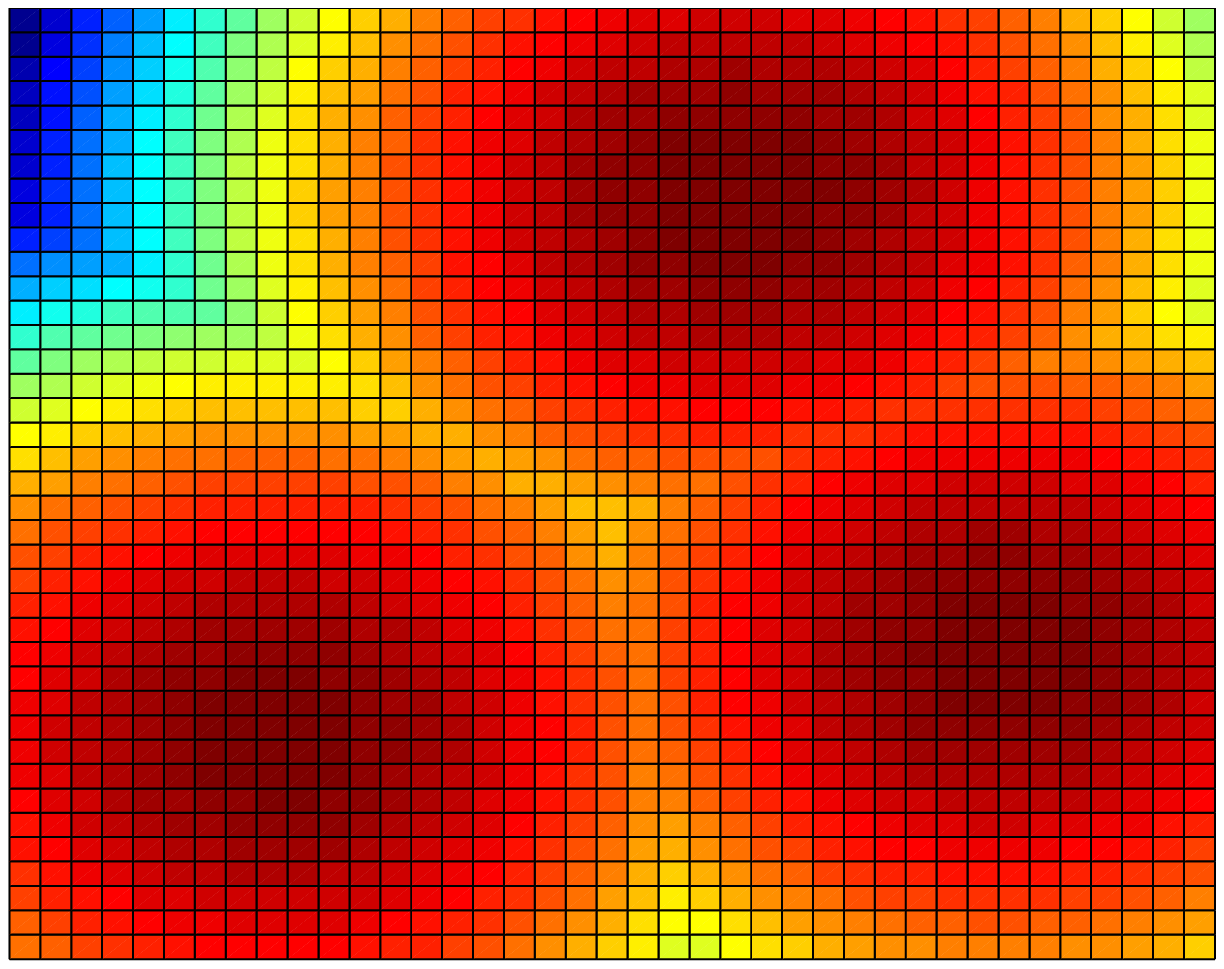} \hspace{\imhspthree}
\caption[]{ \small
Illustration of the trimodal function $\func_\dtilde$ in $\dtilde = 2$.
}
\vspace{\imtextspace}
\label{fig:fdillus}
\end{figure}
}

\newcommand{\insertTableSummary}{
\begin{table*}
\centering
\begin{tabular}{|l|c|c|}
\hline
Kernel & Squared Exponential & \matern \\
\hline
\rule{0pt}{3ex}
\gpbalgos on $D$\superscript{th} order kernel  & $\sqrt{ D^{D+2} T (\log T)^{D+2}}$ &
  $ 2^D \sqrt{D} T^{\frac{\nu + D(D+1)}{2\nu + D(D+1)}} \log T$ \\
\hline 
\rule{0pt}{3ex}
\addgpbalgos on additive kernel & $\sqrt{ d^{d} D^2 T (\log T)^{d+2}}$ & 
  $ 2^d D T^{\frac{\nu + d(d+1)}{2\nu + d(d+1)}} \log T$ \\
\hline
\end{tabular}
\caption{Comparison of Cumulative Regret for \gpbalgos and \addgpbalgos for the
Squared Exponential and \matern kernels.
}
\label{tb:theorySummary}
\end{table*}
}

\newcommand{\insertAlgorithmADDGPBALGO}{
\begin{algorithm}[h]
\textbf{Input: }
Kernels $\kernelii{1}, \dots, \kernelii{M}$, Decomposition
$(\Xcalii{j})_{j=1}^M$ 
\begin{itemize}
\item $\Dcal_0 \leftarrow \emptyset$, 
\item \textbf{for} $j = 1,\dots, M$, $(\gpmeanii{j}_0, \kernelii{j}_0)
\leftarrow (\zero, \kernelii{j})$.
\item \textbf{for} $t=1, 2, \dots$
\begin{enumerate}

  \item \textbf{for} $j=1,\dots,M$, \\ \hspace{0.11in}
    $\xpttii{j} \leftarrow \argmax_{z\in \Xcalii{j}} \gpmeanii{j}_{t-1}(z) + 
    \sqrt{\beta_t} \gpstdii{j}_{t-1}(z)$ 
  \item  $\xptt\leftarrow \bigcup_{j=1}^M \xpttii{j}$.

  \item $\yptt \leftarrow \textrm{Query $\func$ at $\xptt$}$.
  \item $\Dcal_{t} = \Dcal_{t-1} \cup \{ (\xptt, \yptt)\}$.
  \item Perform Bayesian posterior updates conditioned on $\Dcal_t$ 
        to obtain $\gpmeanii{j}_{t},
    \gpstdii{j}_{t}$ for $j=1, \dots, M$.
\end{enumerate}
\end{itemize}
\caption{\addgpbalgo \label{alg:addgpbalgo}}
\end{algorithm}
}

\newcommand{\insertAlgorithmPRACTICALADDGPBALGO}{
\begin{algorithm}[h]
\textbf{Input: }
$\Ninit$, $\Ncyc$, $d$, $M$
\begin{itemize}
\item $\Dcal_0 \leftarrow $ $N_{init}$ points chosen uniformly at random.
\item \textbf{for} $t=1, 2, \dots$
\begin{enumerate}
  \item \textbf{if} ($t\mod \Ncyc = 0$), Learn the kernel hyper parameters and
    the decomposition $\{\Xcal_j\}$ by maximising the GP marginal 
    likelihood. 
  \item Perform steps 1-3 in Algorithm~\ref{alg:addgpbalgo} with 
        $\beta_t = 0.2d\log 2t$.

%
  \item $\Dcal_{t} = \Dcal_{t-1} \cup \{ (\xptt, \yptt)\}$.
  \item Perform Bayesian posterior updates conditioned on $\Dcal_t$ 
        to obtain $\gpmeanii{j}_{t},
    \gpstdii{j}_{t}$ for $j=1, \dots, M$.
\end{enumerate}
\end{itemize}
\caption{\addgpbalgopractical \label{alg:addgpbalgopractical}}
\end{algorithm}
}

\newcommand{\insertAlgorithmGPBALGO}{
\begin{algorithm}
\textbf{Input: }
Kernel $\kernel$, Input Space $\Xcal$. \\
For $t = 1,2 \dots$
\begin{itemize}
\item $\Dcal_0 \leftarrow \emptyset$, 
\item $(\gpmean_0, \kernel_0) \leftarrow (\zero, \kernel)$
\item \textbf{for} $t=1, 2, \dots$
\begin{enumerate}
  \item $\xptt \leftarrow \argmax_{z\in \Xcal} \gpmean_{t-1}(z) + 
          \sqrt{\beta_t}\gpstd_{t-1}(z)$
  \item $\yptt \leftarrow \textrm{Query $\func$ at $\xptt$}$.
  \item $\Dcal_{t} = \Dcal_{t-1} \cup \{ (\xptt, \yptt)\}$.
  \item Perform Bayesian posterior updates to obtain $\gpmean_{t},
    \gpstd_{t}$ for $j=1, \dots, M$.
\end{enumerate}
\end{itemize}
\caption{\gpbalgo \label{alg:gpbalgo}}
\end{algorithm}
}

\newcommand{\insertAlgorithmSEQADDGPBALGOONE}{
\begin{algorithm}
\textbf{Input: }
Kernels $\kernelii{1}, \dots, \kernelii{M}$, Decomposition
$(\Xcalii{j})_{j=1}^M$, Query Budget $T$,
\begin{itemize}
\item $\RR^D \ni \theta = \bigcup_{j=1}^M \theta^{(j)} = \textrm{rand}([0,1]^d)$
\item \textbf{for} $j=1, \dots, M$
  \begin{enumerate}
  \item $\Dcal_0^{(j)} \leftarrow \emptyset$,
  \item $(\gpmeanii{j}_0, \kernelii{j}_0) \leftarrow (\zero, \kernelii{j})$.
  \item \textbf{for} $t=1, 2, \dots T/M$
    \begin{enumerate}
      \item $\xpttii{j} \leftarrow \argmax_{z\in \Xcalii{j}} \gpmeanii{j}(z) + 
        \sqrt{\beta_t} \gpstdii{j}(z)$
      \item  $\xptt\leftarrow \xpttii{j}\bigcup_{k\neq j} \theta^{(k)}$.

      \item $\yptt \leftarrow \textrm{Query $\func$ at $\xptt$}$.
      \item $\Dcal_{t}^{(j)} = \Dcal_{t-1}^{(j)} \cup \{ (\xpttii{j}, \yptt)\}$.
      \item Perform Bayesian posterior updates to obtain $\gpmeanii{j}_{t},
        \gpstdii{j}_{t}$.
    \end{enumerate}
  \item $\theta^{(j)} \leftarrow \xptii{j}_{T/M}$
  \end{enumerate}
  \item Return $\theta$
\end{itemize}
\caption{\seqgpbalgoOne \label{alg:seqgpbalgo1}}
\end{algorithm}
}

\newcommand{\insertAlgorithmSEQADDGPBALGOTWO}{
\begin{algorithm}
\textbf{Input: }
Kernels $\kernelii{1}, \dots, \kernelii{M}$, Decomposition
$(\Xcalii{j})_{j=1}^M$ 
\begin{itemize}
\item $\Dcal_0 \leftarrow \emptyset$, 
\item \textbf{for} $j = 1,\dots, M$, $(\gpmeanii{j}_0, \kernelii{j}_0)
\leftarrow (\zero, \kernelii{j})$.
\item \textbf{for} $t=1, 2, \dots$
\begin{enumerate}
  \item $k = j \mod M$
  \item $\xpttii{k} \leftarrow \argmax_{z\in \Xcalii{k}} \gpmeanii{k}(z) + 
    \sqrt{\beta_t} \gpstdii{k}(z)$
  \item \textbf{for} $j\neq k$, $\xpttii{j} \leftarrow \xptii{j}_{t-1}$  
  \item  $\xptt\leftarrow \bigcup_{j=1}^M \xpttii{j}$.
  \item $\yptt \leftarrow \textrm{Query $\func$ at $\xptt$}$.
  \item $\Dcal_{t} = \Dcal_{t-1} \cup \{ (\xptt, \yptt)\}$.
  \item Perform Bayesian posterior updates to obtain $\gpmeanii{j}_{t},
    \gpstdii{j}_{t}$ for $j=1, \dots, M$.
\end{enumerate}
\end{itemize}
\caption{\seqgpbalgoTwo \label{alg:seqgpbalgo2}}
\end{algorithm}
}

\section{Introduction}
\label{sec:intro}


In many applications we are tasked with zeroth order optimisation of an
expensive to evaluate function $\func$ in $D$ dimensions. 
Some examples are hyper parameter tuning
in expensive machine learning algorithms, experiment design, 
optimising control strategies
in complex systems, and scientific simulation based studies.
In such applications, $\func$ is a blackbox which we can interact with
only by  querying for the value at a specific point. 
Related to optimisation is the bandits problem arising in applications such as
online advertising and reinforcement learning.
Here the objective is to maximise
the cumulative sum of all queries.
In either case, we need to find the optimum of $\func$ using
as few queries as possible by managing exploration and exploitation.

Bayesian Optimisation \cite{mockus91bo} refers to a suite of methods that
tackle this problem by modeling $\func$ as a Gaussian Process (GP).
In such methods, the challenge is two fold. At time step $t$, first estimate the unknown
$\func$ from the query value-pairs. Then use it to intelligently query at
$\xptt$ where the function is likely to be high.
For this, we first use the posterior GP to construct an acquisition 
function $\utilt$ which captures the value of the experiment at a point.
Then we maximise $\utilt$ to determine $\xptt$.

Gaussian process bandits and Bayesian optimisation (\gpbbo)
have been successfully applied in many applications such as tuning
hyperparameters in learning algorithms 
\cite{snoek12practicalBO, bergstra11hyperparameter, mahendran12adaptivemcmc}, 
robotics
\cite{lizotte07automaticGait, martinez07robotplanning} and object tracking 
\cite{denil12boImage}.
However, all such successes have been in low (typically $<10$) dimensions 
\cite{wang13rembo}.
Expensive high dimensional functions occur in several problems 
in fields such as computer vision \cite{bergstra13modelsearch}, 
antenna design \cite{hornby06antenna}, computational astrophysics
\cite{parkinson06wmap3} and biology \cite{gonzalez14gene}.
Scaling \gpbbos methods to high dimensions for practical problems has been
challenging.  Even
current theoretical results suggest that \gpbbos is exponentially
difficult in high dimensions without further assumptions 
\cite{srinivas10gpbandits, bull11boRates}.
To our knowledge, the only approach to date has been to perform regular 
\gpbbos on a low 
dimensional subspace. This works only  under strong assumptions. 

We identify  two key challenges in scaling \gpbbos
to high dimensions. \textbf{The first is the statistical challenge in estimating
the function}. 
Nonparametric regression is inherently difficult in high dimensions
with known lower bounds depending exponentially in dimension
\cite{gyorfi02distributionfree}.
The often exponential sample complexity for regression is invariably reflected
in the regret bounds for \gpbbo. 
\textbf{The second is the computational challenge in maximising $\utilt$}. 
Commonly used global optimisation heuristics used to maximise $\utilt$ themselves
require computation exponential in dimension.
Any attempt to scale \gpbbos to high dimensions must effectively address these
two concerns. 

In this work, we embark on this challenge by treating $\func$
as an \emph{additive function} of mutually exclusive lower dimensional components.
\textbf{Our contributions} in this work are:
\begin{enumerate}
\item We present the  
\addgpbalgos algorithm for optimisation and bandits of an additive function.
An attractive property is that we use an acquisition function which is easy to
optimise in high dimensions.
\item In our theoretical analysis we bound the 
regret for \addgpbalgo. We show that it has only linear dependence on 
the dimension $D$ when
$\func$ is additive\footnote{Post-publication it was pointed out to us that there
was a bug in our analysis. We are working on resolving it and will post an update
shortly. See Section~\ref{sec:conclusion} for more details.
\label{ftn:bug}
}.
\item 
Empirically we demonstrate that \addgpbalgos outperforms naive BO on 
synthetic experiments, an astrophysical simulator and the
Viola and Jones face detection problem.
Furthermore \addgpbalgos does well on several examples
\emph{when the function is not additive}.
\end{enumerate}

A Matlab implementation of our methods  is available 
online at {\small\url{github.com/kirthevasank/add-gp-bandits}}.

\section{Related Work}
\label{sec:relwork}

\gpbbos methods follow a family of GP based active learning methods which
select the next experiment based on the posterior
\citep{osborne12alBayesianQuadrature, ma15pointillistic,
kandasamy15activePostEst}.
In the \gpbbos setting, common acquisition functions include 
Expected improvement \cite{mockus94bo}, probability of improvement
\cite{jones98expensive}, Thompson sampling \cite{thompson33sampling} and
upper confidence bound \cite{auer03ucb}.
Of particular interest to us, is the Gaussian process upper confidence bound
(\gpucb).
It was first proposed and analysed in the noisy setting 
by \citet{srinivas10gpbandits} and extended to the
noiseless case by  \citet{defreitas12expregret}.
\ifthenelse{\boolean{isabridged}}
{
\nocite{azimi10batchbo}
\nocite{hoffman11portfolio}
}
{
Some literature studies variants, such as combining several acquisition 
functions \cite{hoffman11portfolio} and querying in batches
\cite{azimi10batchbo}.
}

To our knowledge, most literature for \gpbbos in high dimensions
are in the setting where the function varies only along a very low dimensional 
subspace \cite{chen12varselbandits,wang13rembo, djolonga13highdimbandits}. 
In these works, the authors do not encounter
either challenge as they perform GPB/ BO in either a
random or carefully selected lower dimensional subspace. However,
assuming that the problem is an easy (low dimensional) one hiding in a high
dimensional space is often too restrictive.
Indeed, our experimental results confirm that such methods perform poorly on
real applications when the assumptions are not met.
While our additive assumption is strong in its own right, it is considerably more
expressive. 
It is more general than the setting in \citet{chen12varselbandits}. 
Even though it 
does not contain the settings in \citet{djolonga13highdimbandits, wang13rembo},
unlike them,
we still allow the function to vary along the entire domain.

\insertTableSummary

Using an additive structure is standard in high 
dimensional regression literature both in the GP framework and otherwise. 
\citet{hastie90gam, ravikumar09spam} treat the function as a sum of one
dimensional components. Our additive framework is more general.
\citet{duvenaud11additivegps} assume a sum of functions of
 all combinations of lower dimensional coordinates.
These literature argue that using an additive model has several advantages even
if $\func$ is \emph{not} additive.
It is a well understood notion in statistics that
when we only have a few samples, using a simpler model to fit our data may give us a
better trade off for estimation error against approximation error. 
This observation is \emph{crucial}: in many applications for Bayesian
optimisation we are forced to work in the low sample regime since calls to the
blackbox are expensive. 
Though the additive assumption is biased for nonadditive functions, it enables us to
do well with only a few samples. 
While we have developed theoretical results only for additive $f$,
empirically we show that our additive model outperforms naive \gpbbos even when
the underlying function is not additive.

Analyses of \gpbbos methods focus on the query complexity of $\func$
which is the dominating cost in relevant applications.
It is usually assumed that $\utilt$ can be maximised to arbitrary precision at
negligible cost. 
Common techniques to maximise $\utilt$ include grid search, Monte Carlo and
multistart methods \cite{brochu12bo}. In our work we use 
the Dividing Rectangles (\direct) algorithm of \citet{jones93direct}.
While these methods are efficient in low dimensions they 
require exponential computation in high dimensions. It is widely
acknowledged in the community that this is a critical bottleneck 
in scaling \gpbbos to high dimensions \cite{defreitas14challenges}.
While we still work in the paradigm where evaluating $\func$ is expensive
and characterise our theoretical results in terms of query complexity,
we believe that
assuming arbitrary computational power to optimise $\utilt$ is too restrictive.
For instance, in hyperparameter tuning the
budget for determining the next experiment is dictated by the cost of the learning
algorithm.
In online advertising and robotic reinforcement learning we need to
act in under a few seconds or real time.

In this manuscript, Section~\ref{sec:probStatement} 
formally details our problem and assumptions.
We present \addgpbalgos in Section~\ref{sec:algorithm} and
our theoretical results in Section~\ref{sec:analysis}.
All proofs are deferred to Appendix~\ref{sec:appAnalysis}.
We summarize the regrets for \addgpbalgos and \gpbalgos in Table~\ref{tb:theorySummary}.
\ifthenelse{\boolean{isabridged}}
{
The experiments section is in Appendix~\ref{sec:experiments}.
Section~\ref{sec:expSummary} presents a summary.
}
{
In Section~\ref{sec:experiments} we present the experiments.
}

\allowdisplaybreaks

\section{Problem Statement \& Set up}
\label{sec:probStatement}

We wish to maximise a function $\func:\Xcal \rightarrow \RR$ where $\Xcal$ is
a \emph{rectangular} region in $\RR^D$. We will assume w.l.o.g $\Xcal = [0,1]^D$.
$\func$ may be nonconvex and gradient information is not available. 
We can interact with $\func$ only by querying at some $x \in
\Xcal$ and obtain a noisy observation $y = \func(x) + \epsilon$. 
Let an optimum point be $\xopt = \argmax_{x\in\Xcal} \func(x)$.
Suppose at time $t$ we choose to query at $\xptt$. Then we incur
\emph{instantaneous regret} $r_t = \funcopt - \func(\xptt)$. In the bandit setting, we are
interested in the \emph{cumulative regret} $R_T = \sum_{t=1}^T r_t = \sum_{t=1}^T
\funcopt - \func(\xptt)$, and in the optimisation setting we are interested in
the \emph{simple regret} $S_T = \min_{t\leq T} r_t = \funcopt - \max_{\xptt}
\func(\xptt)$. 
For a bandit algorithm, a desirable property is to have \emph{no regret}:
 $\lim_{T\rightarrow \infty}\frac{1}{T}R_T = 0$. 
Since $S_T \leq \frac{1}{T}R_T$, any such procedure  
is also a consistent procedure for optimisation.

\textbf{Key structural assumption: }
In order to make progress in high dimensions,
we will assume that $\func$ decomposes into the
following additive form,
\begin{equation}
\func(x) = \funcii{1}(\xii{1}) + \funcii{2}(\xii{2}) + \dots + 
  \funcii{M}(\xii{M}).
\label{eqn:addmodel}
\end{equation}
Here each $\xii{j} \in \Xcalj = [0,1]^{d_j}$ are
lower dimensional components. 
We will refer to the $\Xcalj$'s as ``groups" and the grouping of different
dimensions into these groups $\{\Xcalj\}_{j=1}^M$ as the ``decomposition".
The groups are \emph{disjoint} -- i.e.
if we treat the elements of the
vector $x$ as a set, $\xii{i}\cap\xii{j} = \emptyset$.
We are primarily interestd in the case when $D$ is very large and
the group dimensionality is bounded: $d_j \leq d \ll D$. 
We have $D \asymp dM \geq \sum_j d_j$.
Paranthesised superscripts index the groups and a union over the
groups denotes the reconstruction of the whole from the groups (e.g.
$x = \bigcup_j \xii{j}$ and $\Xcal = \bigcup_j \Xcalj$).
$\xptt$  denotes the point chosen by the algorithm for querying at time $t$.
We will ignore $\log D$ terms in $\bigO(\cdot)$ notation.
Our theoretical analysis assumes that the decomposition is known but
we also present a modified algorithm to handle unknown decompositions and non-additive
functions.

Some smoothness assumptions on $f$ are warranted to make the problem tractable.
A standard in the Bayesian paradigm is to assume $f$ is sampled from a Gaussian
Process \cite{rasmussen06gps} with a covarince kernel
$\kernel:\Xcal\times\Xcal \rightarrow \RR$ and that $\epsilon\sim\Ncal(0,
\eta^2)$.
Two commonly used kernels are the squared exponential (SE) $\kernel_{\sigma, h}$ and
the \matern $\kernel_{\nu, h}$ kernels with parameters $(\sigma, h)$ and $(\nu,
h)$ respectively. 
Writing $r = \|x-x'\|_2$, they are defined as
\begingroup
\allowdisplaybreaks
\begin{align*}
  \kernel_{\sigma, h}(x, x') &= \sigma \exp\left( \frac{-r^2}{2h^2} \right),
  \numberthis \label{eqn:sekernel} \\
  \kernel_{\nu, h}(x, x') &= \frac{2^{1-\nu}}{\Gamma(\nu)}
    \left( \frac{\sqrt{2\nu}r}{h} \right)^\nu
    B_\nu\left( \frac{\sqrt{2\nu}r}{h} \right).
  \numberthis \label{eqn:maternkernel} 
\end{align*}
\endgroup
\ifthenelse{\boolean{isabridged}}{}
{Here $\Gamma, B_\nu$ are the Gamma and modified Bessel functions.
}
A principal convenience in modelling our problem via a GP is that posterior
distributions are analytically tractable.

In keeping with this,
we will assume that each $\funcj$ is sampled from a GP, $\GP(\gpmeanj,
\kernelj)$ where the $\funcj$'s are independent. Here,
$\gpmeanj:\Xcalj \rightarrow \RR$ is the mean and 
$\kernelii{j}:\Xcalj \times \Xcalj \rightarrow \RR$ is the covariance for
$\funcj$. W.l.o.g  let $\gpmeanj = \zero$ for all $j$.
This implies that $\func$ itself is sampled from a GP with
an additive kernel $\kernel(x,x') = \sum_j \kernelj(\xii{j},{\xii{j}}')$.
We state this formally for nonzero
mean as we will need it for the ensuing discussion. \\[\thmparaspacing]

\begin{observation}
Let $\func$ be defined as in Equation~\eqref{eqn:addmodel}, where $\funcj\sim
\GP(\gpmeanii{j}(x), \kernelj(\xii{i}, {\xii{j}}'))$.
Let $y = f(x) + \epsilon$ where $\epsilon~\sim\Ncal(0,\eta^2)$.
Denote $\delta(x,x') = 1 \text{ if } x = x', \text{ and } 0 \text{ otherwise}$.
Then
$y \sim \GP(\gpmean(x), \kernel(x,x') + \eta^2\delta(x,x'))$ where
\begingroup
\allowdisplaybreaks
\begin{align*}
\gpmean(x) &= \gpmeanii{1}(\xii{1}) + \dots + \gpmeanii{M}(\xii{M}) 
  \numberthis \label{eqn:sumKernel} \\
\kernel(x,x') &= \kernelii{1}(\xii{1}, {\xii{1}}') +
  \dots + \kernelii{M}(\xii{M}, {\xii{M}}').
\end{align*}
\endgroup
\label{thm:sumIsGP}
\end{observation}
\vspace{\thmparaspacing}
\vspace{-0.2in}
We will call a kernel such as $\kernelj$ which acts only on $d$ variables a
$d^{th}$ order kernel. A kernel which acts on all the variables is a $D^{th}$
order kernel.
Our kernel for $\func$ is a sum of $M$ at most $d^{th}$ order kernels
which, we will show, is statistically simpler than a $D^{th}$ order kernel.

We conclude this section by looking at some seemingly straightforward approaches to 
tackle the problem. The first natural question is of course why not
directly run \gpbalgos using the additive kernel? Since it is simpler than a
$D$\superscript{th} order kernel we
can expect statistical gains. While this is true, it still requires optimising 
$\utilt$ in $D$ dimensions to determine the next point which is
expensive. 

Alternatively, for an additive function, we could adopt a sequential approach where we
use $1/M$ fraction of our query budget to maximise the first group
by keeping the rest of the coordinates constant.
Then we proceed to the second group and so on. While optimising a $d$
dimensional acquisition function is easy, 
this approach is not desirable for 
several reasons. First, it will not
be an anytime algorithm as we will have to pre-allocate our query budget to
maximise each group. Once we proceed to a new group we cannot come back and
optimise an older one.  Second, such an approach places too much faith in the
additive assumption. We will only have explored $M$
$d$-dimensional hyperplanes in the entire space. Third, it is not suitable as a
bandit algorithm as we suffer high regret until we get to the last group.
We further elaborate on the deficiencies of this and other sequential
approaches in Appendix~\ref{sec:appSeqGPB}.

\section{Algorithm}
\label{sec:algorithm}

Under an additive assumption, our algorithm has two components.
First, we obtain the posterior GP for each $\funcj$
using the query-value pairs until time $t$. Then we maximise a $d$
dimensional \gpucb-like acquisition function on \emph{each} GP to construct the
next query point. 
\ifthenelse{\boolean{isabridged}}{}{
Since optimising $\utilt$ depends exponentially in dimension,
this is cheaper than optimising one acquisition on the combined
GP. 
}

\insertFigureAddGPGraph

\subsection{Inference on Additive GPs}
\label{sec:infAddGPs} 
Typically in GPs, given noisy labels, $Y=\{y_1, \dots, y_n\}$ at points
$X = \{x_1, \dots, x_n\}$, we are interested in inferring the posterior
distribution for $f_* = f(x_*)$ at a new point $x_*$. 
In our case though, we will be primarily interested in  the distribution of
$\funcj_* = \funcj(\xii{j}_*)$ conditioned on $X, Y$. 
We have illustrated this graphically
in Figure~\ref{fig:addGPgraph}.
The joint distribution of $\funcj_*$ and $Y$ can be written as
\[
\left(\!\!\! \begin{array}{c} \funcj_* \\ Y \end{array}
 \!\!\!\right)\;
\sim \Ncal \left( \zero , 
\left[ \!\!\!\begin{array}{cccc}
  \kernelj(\xii{j}_*, \xii{j}_*)  & \kernelj(\xii{j}_*, \Xii{j}) \\
  \kernelj(\Xii{j}, \xii{j}_*) & \kernel(X,X) + \eta^2 I_n
  \end{array} \!\!\!\right] \right).
\]
The $p$\superscript{th} element of $\kernelj(\Xii{j}, \xii{j}_*) \in \RR^n$
is $\kernel(\xii{j}_p, \xii{j}_*)$ and the $(p,q)$\superscript{th} element of
$\kernel(X, X) \in \RR^{n\times n}$ is
$\kernel(x_p, x_q)$.
We have used the fact $\Cov(\funcii{i}_*, y_p) = 
\Cov(\funcii{i}_*, \sum_j \funcj(\xii{j}_p) + \eta^2 \epsilon)
= \Cov(\funcii{i}_*, \funcii{i}(\xii{i}_p)) = \kernelii{i}(\xii{i}_*,
\xii{i}_p)$ as $\funcj \perp \funcii{i},\forall i\neq j$.
By writing $\Kappa = \kappa(X,X) + \eta^2 I_n \in \RR^{n\times n}$,
the posterior for $\funcj_*$ is,
\begingroup
\allowdisplaybreaks
\begin{align*}
&\funcj_* | x_*, X, Y \,\sim\, \Ncal \big( \kernelj(\xii{j}_*, \Xii{j}) 
  \Kappa^{-1} Y , 
\numberthis \label{eqn:addPosterior}\\
&\hspace{0.2in}\kernelj(\xii{j}_*, \xii{j}_*) - \kernelj(\xii{j}_*, \Xii{j}) 
    \Kappa^{-1} \kernelj(X, \xii{j}) \big)
\end{align*}
\endgroup

\subsection{The \addgpbalgos Algorithm}
\label{sec:addgpbalgo}

In \gpbbos algorithms, at each time step $t$ we maximise an acquisition function
$\utilt$ to determine the next point: $\xptt =
\argmax_{x\in\Xcal} \utilt(x)$. 
The acquisition function is itself constructed using the posterior GP. 
The \gpbalgos acquisition function, which we focus on here is,
\[
\utilt(x) = \gpmeantmo(x) + \betath \gpstdtmo(x).
\]
Intuitively, the $\gpmeantmo$ term in the \gpbalgos objective prefers points where
$\func$ is known to be high, the $\gpstdtmo$ term prefers points where we
are uncertain about $\func$ and $\betath$ negotiates the tradeoff. The former
contributes to the ``exploitation" facet of our problem, in that we wish to
have low instantaneous regret. The latter contributes to the ``exploration"
facet since we also wish to query at regions we do not know much about $\func$
lest we miss out on regions where $\func$ is high.
We provide a brief summary of \gpbalgos and its theoretical
properties in Appendix~\ref{sec:reviewGPB}.

As we have noted before,
maximising $\utilt$ which is typically multimodal to obtain $\xptt$ is itself
a difficult problem. 
In any grid search or branch and bound methods such as \direct,
maximising a function to within $\zeta$ accuracy, 
requires $\bigO(\zeta^{-D})$ calls to $\utilt$.
Therefore, for large $D$ maximising $\utilt$ is extremely difficult. 
In practical settings, especially in situations where we are computationally
constrained, this poses serious limitations for \gpbbos as
we may not be able to optimise $\utilt$ to within a desired accuracy.

Fortunately, in our setting we can 
be more efficient. We propose an alternative acquisition function
which applies to an additive kernel. We define the \emph{Additive Gaussian Process
Upper Confidence Bound} (\addgpucb) to be 
\vspace{-0.08in}
\begin{equation}
\addutilt(x) = \gpmeantmo(x) + \betath \sum_{j=1}^M \gpstdjtmo(\xii{j}).
\label{eqn:addgpbutility}
\end{equation}
We immediately see that we can write $\addutilt$ as a sum of functions
on orthogonal domains:
$\addutilt(x) = \sum_j \addutiltj(\xii{j})$ where
  $\addutiltj(\xii{j}) = \gpmeanjtmo(\xii{j}) + \betath
\gpstdjtmo(\xii{j})$. This means that $\addutilt$ can be maximised by maximising
each $\addutiltj$ separately on $\Xcalj$.
As we need to solve
$M$ at most $d$ dimensional optimisation problems, it requires only 
$\bigO(M^{d+1}\zeta^{-d})$ calls to the utility function in total -- 
far more favourable than maximising $\utilt$.

Since the cost for maximising the acquisition function is a key theme in this
paper let us delve into this a bit more. One call to $\utilt$ requires $\bigO(Dt^2)$
effort. For $\addutilt$ we need $M$ calls each requiring $\bigO(d_jt^2)$ effort. So
both $\utilt$ and $\addutilt$ require the same effort in this front.
For $\utilt$, we need to know the posterior for only  $\func$ whereas
for $\addutilt$ we
need to know the posterior for each $\funcj$.
However, the brunt of the work in obtaining the posterior is  
the $\bigO(t^3)$ effort in inverting the
$t\times t$ matrix $\Kappa$ in~\eqref{eqn:addPosterior}
which needs to be done for both $\utilt$ and
 $\addutilt$. For $\addutilt$, we can obtain the inverse once and reuse it $M$
times, so the cost of obtaining the posterior is $\bigO(t^3 + Mt^2)$.
Since the number of queries needed will be super linear in $D$ and hence
$M$, the $t^3$ term dominates. Therefore obtaining each posterior $\funcj$ is only
marginally more work than obtaining the posterior for $\func$.
Any difference here is easily offset by the cost for maximising the 
acquisition function.

The question remains then if maximising $\addutilt$ would result in low regret. 
Since $\utilt$ and $\addutilt$ are neither equivalent nor have the same maximiser
 it is not immediately apparent that this should work.
Nonetheless, intuitively this seems like a reasonable scheme
since  the $\sum_j \gpstdjtmo$ term captures some notion of the uncertainty
and contributes to exploration.
In Theorem~\ref{thm:addGPBthm} we show that this intuition is reasonable --
 maximising $\addutilt$ achieves the
\emph{same} rates as $\utilt$ for cumulative and simple regrets if the kernel is
additive. 

We summarise the resulting algorithm in Algorithm~\ref{alg:addgpbalgo}.
In brief, at time step $t$, we obtain the posterior distribution for $\funcj$ and
maximise $\addutiltj$ to determine the coordinates $\xpttii{j}$. We do this for
each $j$ and then combine them to obtain $\xptt$.
\vspace{-0.2in}
\insertAlgorithmADDGPBALGO
\vspace{-0.2in}


\subsection{Main Theoretical Results}
\label{sec:analysis}

Now, we present our main theoretical contributions.
We bound the regret  for \addgpbalgos  under different kernels. 
Following \citet{srinivas10gpbandits}, we first bound the statistical difficulty
of the problem as determined by the kernel. We show that under
additive kernels the problem is much easier than when using a full
$D$\superscript{th}
order kernel. Next, we show that the \addgpbalgos algorithm is able to exploit
the additive structure and obtain the same rates as \gpbalgo. 
The advantage to
using \addgpbalgos is that it is much easier to optimise the acquisition
function.
For our analysis, we will need Assumption~\ref{asm:partialDerivLipschitz} and
Definition~\ref{def:infgainGP}. 
\\[-0.1in]
\begin{assumption} Let $\func$ be sampled from a GP with kernel $\kernel$.
 $\kernel(\cdot, x)$ is $L$-Lipschitz for all $x$. Further,
the partial derivatives of $\func$  satisfies the following high probability bound. 
There exists constants $a, b >0$ such that,
\[
\PP\left( \sup_{x} 
\Big|\partialfrac{x_i}{\func(x)}\Big| > J \right)
\leq a e^{-(J/b)^2}.
\]
\label{asm:partialDerivLipschitz}
\end{assumption}
\vspace{-0.2in} 
The Lipschitzian condition is fairly mild and
the latter condition holds for four times differentiable 
stationary kernels such as the SE and \matern kernels
for $\nu > 2$ \cite{ghosal06gpconsistency}.
\citet{srinivas10gpbandits} showed that the statistical difficulty of \gpbbos
is determined by
the \emph{Maximum Information Gain} as defined below.
We bound this quantity for additive SE and \matern kernels in
Theorem~\ref{thm:infGainBounds}. This is our first main theorem.
\\[-0.1in]

\begin{definition}(Maximum Information Gain)
Let $f\sim \GP(\gpmean, \kernel)$, $y_i = f(x_i) + \epsilon$ where
$\epsilon\sim\Ncal(0, \eta^2)$. 
Let $A = \{x_1, \dots, x_T\}
\subset \Xcal$ be a finite subset, $\func_A$
denote the function values at these points and $y_A$
denote the noisy observations. 
Let $I$ be the Shannon Mutual Information.
The Maximum Information Gain between $y_A$ and $f_A$ is
\[
\gamma_T = \max_{A\subset \Xcal, |A| = T} I(y_A; f_A).
\]
\label{def:infgainGP}
\end{definition}
\vspace{-0.1in}
\begin{theorem}
Assume that the kernel $\kernel$ has the additive form of~\eqref{eqn:sumKernel},
and that each $\kernelj$ satisfies Assumption~\ref{asm:partialDerivLipschitz}.
W.l.o.g assume $\kernel(x, x') = 1$. Then,
\begin{enumerate}
\item If each $\kernelj$ is a $d_j^{th}$ order squared exponential
kernel~\eqref{eqn:sekernel} where $d_j \leq d$, then
$\gamma_T \in \bigO(D d^{d} (\log T)^{d+1})$.
\item If each $\kernelj$ is a $d_j^{th}$ order \matern 
kernel~\eqref{eqn:maternkernel} where $d_j \leq d$ and $\nu > 2$, then
$\gamma_T \in \bigO(D 2^d T^{\frac{d(d+1)}{2\nu + d(d+1)}} \log(T))$.
\end{enumerate}
\label{thm:infGainBounds}
\end{theorem}
\vspace{\thmparaspacing}
\ifthenelse{\boolean{isabridged}}{
The proof is given in Appendix~\ref{sec:appAnalysisInfGain}.
}
{
We use bounds on the eigenvalues of the SE and \matern kernels
from \citet{seeger08information} 
and a result from \citet{srinivas10gpbandits} which
 bounds the information gain via the eigendecay of the kernel.
We bound the eigendecay of the sum $\kernel$ via $M$ and the 
eigendecay of a single $\kernelj$.
The complete proof is given in Appendix~\ref{sec:appAnalysisInfGain}.
}
The important observation is that the dependence on $D$ is linear for an additive
kernel. In contrast, for a $D$\superscript{th} order kernel this is
exponential \cite{srinivas10gpbandits}. 
\ifthenelse{\boolean{isabridged}}{}{

}
Next, we present our second main theorem 
which bounds the regret for \addgpbalgos for an additive kernel as
given in Equation~\ref{eqn:sumKernel}.
\\[-0.1in]

\begin{theorem}
Suppose $\func$ is constructed by sampling $\funcj\sim \GP(\zero, \kernelii{j})$
for $j = 1, \dots, M$ and then adding them. Let all kernels
$\kernelii{j}$ satisfy assumption~\ref{asm:partialDerivLipschitz} for some
$L,a,b$. Further, we maximise the acquisition function $\addutilt$ to within
$\zetabase t^{-1/2}$ accuracy at time step $t$.
Pick $\delta\in(0,1)$ and choose
\begingroup
\begin{align*}
\beta_t &= 2\log\left(\frac{M\pi^2 t^2}{2\delta}\right) + 
  2d\log\left(Dt^3\right) 
  \in \; \bigO\left( d\log t \right).
\end{align*}
\endgroup
Then, {\rm\addgpbalgo} attains cumulative regret $R_T \in 
\bigO\left(\sqrt{\vphantom{T^1}D\gamma_TT\log T}\right)$ and hence
simple regret $S_T \in \bigO\left(\sqrt{\vphantom{T^1}D\gamma_T\log T/T}\right)$.
Precisely, with probability $> 1- \delta$,
\[
\forall T \geq 1, \;\;\;
R_T \leq \sqrt{8C_1\beta_TMT\gamma_t} + 2\zetabase \sqrt{T} + C_2.
\]
where $C_1 = 1/\log(1+\eta^{-2})$ and $C_2$ is a constant depending on 
$a$, $b$, $D$, $\delta$, $L$ and $\eta$.
\label{thm:addGPBthm}
\end{theorem}
\vspace{\thmparaspacing}

\ifthenelse{\boolean{isabridged}}{
Our proof uses ideas from \citet{srinivas10gpbandits}. We also show that
complete maximisation of $\addutilt$ is not required provided that
the accuracy improves at rate $O(t^{1/2})$.
The proof is given in Appendix~\ref{sec:appAnalysisRates}.
}
{
Part of our proof uses ideas from \citet{srinivas10gpbandits}.
We show that 
$\sum_j \beta_t \gpstdjtmo(\cdot)$ forms a credible
interval for $\func(\cdot)$ about the posterior mean
$\gpmean_t(\cdot)$ for an additive kernel in \addgpbalgo.
We relate the regret to this confidence set 
using a covering argument. We also show that our regret doesn't suffer severely
if we only approximately optimise the acquisition provided that the accuracy
improves at rate $\bigO(t^{-1/2})$. For this we establish
smoothness of the posterior mean.  The
correctness of the algorithm follows from the fact that \addgpbalgos
can be maximised by individually maximising 
$\addutiltj$ on each $\Xcalj$.
The complete proof is given in Appendix~\ref{sec:appAnalysisRates}. 
}
When we combine the results in Theorems~\ref{thm:infGainBounds}
and~\ref{thm:addGPBthm} we obtain the rates given in
Table~\ref{tb:theorySummary}\footnote{See Footnote~\ref{ftn:bug}.}.

One could consider alternative lower order kernels --  
one candidate is the sum of all possible $d^{th}$ order kernels 
\cite{duvenaud11additivegps}. Such a kernel would arguably allow us to represent
a larger
class of functions than our kernel in~\eqref{eqn:sumKernel}.
If, for instance, we choose each of them to be
a SE kernel, then it can be shown that  $\gamma_T \in \bigO(D^d
d^{d+1}(\log T)^{d+1})$. Even though this is worse than our kernel in
$\textrm{poly}(D)$
factors, it is still substantially better than using a
$D$\superscript{th} order kernel.
However, maximising the corresponding utility function, 
either of the form $\utilt$ or $\addutilt$, is still a $D$ dimensional problem. 
We reiterate that what renders our algorithm attractive
in large $D$ is not just the statistical gains due to
the simpler kernel. It is also the fact that our acquisition function can
be efficiently maximised.


\subsection{Practical Considerations}
\label{sec:practicalImplementation}

Our practical implementation differs from our theoretical analysis in the following 
aspects.

\textbf{Choice of $\beta_t$:} $\beta_t$ as specified by
Theorems~\ref{thm:addGPBthm},
usually tends to be conservative in practice 
\cite{srinivas10gpbandits}. For good empirical performance a
more aggressive strategy is required.
In our experiments, we set $\beta_t = 0.2d\log(2t)$ which offered a 
good tradeoff between exploration and exploitation.
Note that this captures the correct dependence on $D, d$ and $t$ in
Theorems~\ref{thm:addGPBthm} and~\ref{thm:GPBthm}. 

\textbf{Data dependent prior: } Our 
 analysis assumes that we know the GP kernel of the prior. 
In reality 
this is rarely the case. In our experiments, we choose the hyperparameters of
the kernel by maximising the GP marginal likelihood \cite{rasmussen06gps} every
$N_{cyc}$ iterations.

\textbf{Initialisation: } Marginal likelihood based kernel tuning can be unreliable
with few data points. This is a problem in the first few iterations. 
Following the recommendations in \citet{bull11boRates} we initialise
\addgpbalgos (and \gpbalgo) using $N_{init}$ points selected uniformly at random.

\textbf{Decomposition \& Non-additive functions: }
If $\func$ is additive and the decomposition is known, we use it directly. But it
may not always be known or $\func$ may not be additive. 
Then, we could treat the decomposition as a hyperparameter of the
additive kernel and maximise the marginal likelihood w.r.t the decomposition.
However, given that there are $D!/{d!}^M M!$ possible
decompositions, computing the marginal likelihood for all of them is 
infeasible. 
We circumvent this issue by randomly selecting a few ($\bigO(D)$)
decompositions and choosing the one with the largest marginal likelihood.
Intuitively,
if the function is not additive, with such a ``partial maximisation" we can 
hope to capture some existing marginal structure in $\func$.
At the same time, even an exhaustive
maximisation will not do much better than a partial maximisation if there is no
additive structure.
Empirically, we found that partially optimising for the
decomposition performed slightly  
better than using a fixed decomposition
or a random decomposition at each step.
We incorporate this procedure for finding an appropriate decomposition as part
of the kernel hyper parameter learning procedure every $N_{cyc}$ iterations.

How do we choose $(d,M)$ when $\func$ is not additive?
If $d$ is large we allow for richer class of functions, but risk high variance.
For small $d$, the kernel is too simple and we have high bias but low variance -- 
further optimising $\addutilt$ is easier.
In practice we found that our procedure was fairly robust for 
reasonable choices of $d$. 
Yet this is an interesting theoretical question. We also believe it is a
difficult one. Using the marginal likelihood alone will not work
as the optimal choice of $d$ also depends on the computational budget for 
optimising $\addutilt$.
We hope to study this question in future work. 
For now, we give some recommendations at the end.
Our modified algorithm with these practical considerations is
given below. 
Observe that in this specification if 
we use $d=D$ we have the original \gpbalgos algorithm.

\insertAlgorithmPRACTICALADDGPBALGO

\ifthenelse{\boolean{isabridged}}
{
\insertFigureSummaryToy

\section{Summary of Experiments}
\label{sec:expSummary}

In this summary we present results in the optimisation setting.
Refer Appendix~\ref{sec:experiments} for results on bandits.
Following, \citet{brochu12bo} we use \directs to maximise $\utilt,
\addutilt$. 
To demonstrate the efficacy of \addgpbalgos we optimise the
acquisition function under a constrained budget.
We compare \addgpbalgos against \gpbalgo, random querying (RAND) and
\direct.
On the real datasets we also compare it to the Expected Improvement (\gpei) acquisition
function which is popular in BO applications and the method of 
\citet{wang13rembo} which uses a random projection before applying BO (REMBO).
We have multiple instantiations of \addgpbalgos for different values for
$(d,M)$. 

In contrast to existing literature in the BO community, we found that the UCB
acquisitions outperformed \gpei. One possible reason may be that under
a constrained budget, UCB
is robust to imperfect maximisation (Theorem~\ref{thm:addGPBthm})
whereas \gpeis may not be. Another reason may be our choice of constants in
UCB (Section~\ref{sec:practicalImplementation}).

\subsection{Simulations on Synthetic Functions}
\label{sec:sumExpSynthetic}

We create a series of additive functions by replicating a $\dtilde$ dimensional
function $f_\dtilde$ in $\Mtilde$ groups. 
(We use the prime to avoid confusion with our \addgpbalgos
instantiations with different $(d,M)$ values.)
So the function doesn't depend on $D-\dtilde\Mtilde$ coordinates.
We have illustrated $f_\dtilde$ for $\dtilde=2$ in the first figure in
Fig~\ref{fig:summaryToyResults} (See Eq~\eqref{eqn:fdtilde}
in~\ref{sec:syntheticExp}).
Since each $f_\dtilde$ has $3$ modes, the function has $3^\Mtilde$ modes.
In the synthetic experiments we use an instantiation of \addgpbalgos that knows 
the decomposition--i.e. $(d,M) =
(\dtilde, \Mtilde)$ and the grouping of coordinates. We refer to this as
\addgpbexps. For the rest we use a $(d,M)$ decomposition by creating $M$ groups
of size at most $d$ and
find a good grouping by partially maximising the
marginal likelihood (Section~\ref{sec:practicalImplementation}).
We refer to them as \addgpbexpd.

For \gpbalgos and \gpeis we allocate a budget of
$\min(5000, 100D)$ \directs function evaluations to optimise the
acquisition function. 
For all $\addgpbexpd$ methods we set it to $90\%$ of this
amount to account for the additional overhead in 
posterior inference for each $\funcj$.
While the $90\%$ seems arbitrary, in our experiments this was
hardly a factor as the cost was dominated by the inversion of $\Kappa$.
Therefore, for our $10D$ problem 
we maximise $\utilt$
with $\beta_t = 2\log(2t)$ with $1000$ evaluations whereas for
$\addgpbexpii{5/2}$ we maximise each $\addutiltj$ with $\beta_t = \log(2t)$ with
$450$ evaluations.

We refer to each example by the configuration of the additive function--its
$(D,\dtilde,\Mtilde)$ values. 
In the $(10,3,3)$ example \addgpbexpss does best
since it knows the correct model and the acquisition function can be maximised
within the budget. However \addgpbexpii{3/4} and \addgpbexpii{5/2} models do well
too and outperform \gpbexp. \addgpbexpii{1/10} 
performs poorly since it is statistically not expressive enough to capture the
true function (high bias). 
In the $(24,11,2)$, $(40,18,2)$ and $(96,29,3)$ examples 
\addgpbexpss outperforms \gpbexp. However, it is not competitive with the
\addgpbexpds for small $d$.
Even though \addgpbexpss knows the correct decomposition,
there are two possible failure modes since $\dtilde$ is large. The 
variance is very high in the absence of sufficient data points. In
addition, optimising the acquisition function is also difficult. This illustrates
our previous argument that using an additive kernel can be advantageous even
on non-additive functions.
In the $(40,5,8)$, $(96,5,19)$ examples \addgpbexpss performs best as
$\dtilde$ is small enough. But again, almost all \addgpbexpds instantiations 
outperform \gpbexp. 
In contrast to the small $D$ examples, for large $D$, \gpbalgos and \addgpbexpds
with large $d$ perform worse than \direct. 
This is probably because the acquisition cannot be maximised to sufficient
accuracy within the budget.
We have only presented a subset of our simulations here. Please see
Appendix~\ref{sec:syntheticExp} for more experiments and other details.

\insertFigureReal

\subsection{Real Experiments}
\label{sec:sumExpReal}

\textbf{SDSS Galaxy Data: }
Here, we use galaxy data from the Sloan Digital Sky Survey to find the maximum
likelihood values for $20$ cosmological parameters. The likelihood is computed via an
astrophysical simulation. Software is obtained
from~\citet{tegmark06lrgs}. 
Each query to the likelihood takes 2-5 seconds. 
The likelihood only depends on $9$ of the parameters but we augment it to $20$
dimensions to emulate the fact that in real astrophysical applications we may not
know the relevant parameters.
In order to be
wall clock time competitive with RAND and \directs we use $500$ evaluations
for \gpbalgo, \gpeis and REMBO and $450$ for \addgpbexpds to maximise the acquisition function.
We have elaborated more details in Appendix~\ref{sec:lrgsExp}.
The results are given in~\ref{fig:sumlrg}.
Despite the fact that the function may not be
additive, all \addgpbexpds methods outperform \gpbalgos and \gpei. 
Since the function only depends on $9$ parameters 
we used REMBO with a $9$ dimensional projection. Despite this advantage to REMBO 
it is not as competitive with the \addgpbexpds methods. 
Here
\addgpbexpii{5/4} performs slightly better than the rest since it seems to have
the best tradeoff between being statistically expressive enough to capture the
function while at the same time being easy enough to optimise the acquisition 
function within the allocated budget.

\textbf{Viola \& Jones Face Detection: }
The Viola \& Jones Cascade Classifier (VJ) \cite{viola01cascade}
is a popular method for face detection in computer vision based on the Adaboost
algorithm. In this experiment we use the VJ face dataset and the OpenCV implementation
\cite{bradski08opencv}
 which implements the classifier as a 22-stage cascade. 
The task is to find the $22$ threshold values for each stage to maximise
classification accuracy.
Each function call takes 30-40 seconds and is the the dominant cost
in this experiment. We use $1000$ \directs evaluations to optimise the
acquisition function for \gpbalgo, \gpeis and REMBO and $900$ for \addgpbexpd.
We use REMBO with a $5$ dimensional projection.
The results are given in Figure~\ref{fig:sumvj}. 
Not surprisingly, REMBO performs  worst since it is searching only on a $5$ dimensional
space.
Barring \addgpbexpii{1/22} all other \addgpbexpds instantiations outperform \gpbalgos and
\gpeis with \addgpbexpii{6/4} performing best.
Interestingly, we also find a configuration for the thresholds that outperforms the one
used in OpenCV.

}
{
\section{Experiments}
\label{sec:experiments}

\insertFigureFuncTwoD

To demonstrate the efficacy of \addgpbalgos over \gpbalgos we optimise the
acquisition function under a constrained budget. 
Following, \citet{brochu12bo} we use \directs to maximise $\utilt,
\addutilt$. 
We compare \addgpbalgos against \gpbalgo, random querying (RAND) and
\direct
\footnote{There are several optimisation methods based on
simulated annealing, cross entropy and genetic algorithms. We
use \directs since its easy to configure and known to work well in practice.}.
On the real datasets we also compare it to the Expected Improvement (\gpei) acquisition
function which is popular in BO applications and the method of 
\citet{wang13rembo} which uses a random projection before applying BO (REMBO).
We have multiple instantiations of \addgpbalgos for different values for
$(d,M)$. 
For optimisation, we perform comparisons based on the simple regret $S_T$ and
for bandits we use the time averaged cumulative regret $R_T/T$. 

For all \gpbbos methods we set $N_{init} = 10$, $N_{cyc} = 25$ 
in all experiments. Further, for the first $25$ iterations we set the bandwidth
to a small value $(10^{-5})$ to encourage an explorative strategy.
We use SE kernels for each additive kernels and use the same scale $\sigma$ and
bandwidth $h$ hyperparameters for all the kernels. Every $25$ iterations 
we maximise the marginal
likelihood with respect to these $2$ hyperparameters in addition to the
decomposition.

In contrast to existing literature in the BO community, we found that the UCB
acquisitions outperformed \gpei. One possible reason may be that under
a constrained budget, UCB
is robust to imperfect maximisation (Theorem~\ref{thm:addGPBthm})
whereas \gpeis may not be. Another reason may be our choice of constants in
UCB (Section~\ref{sec:practicalImplementation}).

\insertFigureToyResults
\insertFigureToyResultsThree

\subsection{Simulations on Synthetic Data}
\label{sec:syntheticExp}

First we demonstrate our technique on a series of synthetic examples.
For this we construct additive functions for different values for the
maximum group size $\dtilde$ and the number of groups $\Mtilde$.
We use the prime to distinguish it from \addgpbalgos instantiations with
different combinations of $(d,M)$ values.
The $\dtilde$ dimensional function $f_\dtilde$ is, 
\ifthenelse{\boolean{istwocolumn}}
{
  \begin{align*}
  &f_\dtilde(x) = \log\Bigg(
    0.1\frac{1}{h_\dtilde^\dtilde} \exp\left( \frac{\|x-v_1\|^2}{2h_\dtilde^2} \right) + 
  \numberthis \label{eqn:fdtilde} \\
  &\hspace{0.1in}  0.1\frac{1}{h_\dtilde^\dtilde} 
      \exp\left( \frac{\|x-v_2\|^2}{2h_\dtilde^2} \right) + 
    0.8\frac{1}{h_\dtilde^\dtilde} \exp\left( \frac{\|x-v_3\|^2}{2h_\dtilde^2} \right)
    \Bigg)
  \end{align*}
}
{
  \begin{align*}
  f_d(x) = \log\Bigg(
    0.1\frac{1}{h_\dtilde^\dtilde} \exp\left( \frac{\|x-v_1\|^2}{2h_\dtilde^2} \right) + 
  \hspace{0.1in}  0.1\frac{1}{h_\dtilde^\dtilde} 
      \exp\left( \frac{\|x-v_2\|^2}{2h_\dtilde^2} \right) + 
    0.8\frac{1}{h_\dtilde^\dtilde} \exp\left( \frac{\|x-v_3\|^2}{2h_\dtilde^2} \right)
    \Bigg) \numberthis
  \label{eqn:fdtilde}
  \end{align*}
}
where $v_1,v_2,v_3$ are fixed $\dtilde$ dimensional vectors and $h_\dtilde = 
0.01\dtilde^{0.1}$.
Then we create $\Mtilde$ groups of coordinates
by randomly adding $\dtilde$ coordinates into each group. On each such group we 
use $f_\dtilde$ and then add them up to obtain the composite function $f$. Precisely,
\[
f(x) = f_\dtilde(\xii{1}) + \dots + f_\dtilde(\xii{M})
\]
The remaining $D - \dtilde\Mtilde$ coordinates do not contribute to the
function. 
Since $f_\dtilde$ has $3$ modes, $f$ will have $3^\Mtilde$ modes.
We have illustrated $f_\dtilde$ for $\dtilde = 2$ in Figure~\ref{fig:fdillus}.

In the synthetic experiments we use an instantiation of \addgpbalgos that knows 
the decomposition--i.e. $(d,M) =
(\dtilde, \Mtilde)$ and the grouping of coordinates. We refer to this as
\addgpbexps. For the rest we use a $(d,M)$ decomposition by creating $M$ groups
of size at most $d$ and
find a good grouping by partially maximising the
marginal likelihood (Section~\ref{sec:practicalImplementation}).
We refer to them as \addgpbexpd.

For \gpbalgos we allocate a budget of
$\min(5000, 100D)$ \directs function evaluations to optimise the
acquisition function. For all \addgpbexpds methods we set it to $90\%$ of this
amount\footnote{While the $90\%$ seems arbitrary, in our experiments this was
hardly a factor as the cost was dominated by the inversion of $\Kappa$.} 
to account for the additional overhead in posterior inference for each $\funcj$.
Therefore, in our $10D$ problem 
we maximise $\utilt$
with $\beta_t = 2\log(2t)$ with $1000$ \directs evaluations whereas for
$\addgpbexpii{2/5}$ we maximise each $\addutiltj$ with $\beta_t = 0.4\log(2t)$ with $180$
evaluations. 
%

The results are given in Figures~\ref{fig:toy} and~\ref{fig:toythree}.
We refer to each example by the configuration of the additive function--its
$(D,\dtilde,\Mtilde)$ values. 
In the $(10,3,3)$ example \addgpbexpss does best
since it knows the correct model and the acquisition function can be maximised
within the budget. However \addgpbexpii{3/4} and \addgpbexpii{5/2} models do well
too and outperform \gpbexp. \addgpbexpii{1/10}
performs poorly since it is statistically not expressive enough to capture the
true function. In the $(24,11,2)$, $(40,18,2)$, $(40,35,1)$, $(96,29,3)$ and
$(120, 55,2)$ examples 
\addgpbexpss outperforms \gpbexp. However, it is not competitive with the
\addgpbexpds for small $d$.
Even though \addgpbexpss knew the correct decomposition,
there are two possible failure modes since $\dtilde$ is large. The kernel is complex
and the estimation error is very high in the absence of sufficient data points. In
addition, optimising the acquisition is also difficult. This illustrates
our previous argument that using an additive kernel can be advantageous even if
the function is not additive or the decomposition is not known.
In the $(24,6,4)$, $(40,5,8)$ and $(96,5,19)$ examples \addgpbexpss performs best as
$\dtilde$ is small enough. But again, almost all \addgpbexpds instantiations outperform
\gpbexp. 
In contrast to the small $D$ examples, for large $D$, \gpbalgos and \addgpbexpds
with large $d$ perform worse than \direct. 
This is probably  
because our budget for maximising $\utilt$ is inadequate to optimise the
acquisition function to sufficient accuracy.
For some of the large $D$ examples the cumulative regret is low for \addgpbalgos
and \addgpbexpds with large $d$. This is probably since they have already started 
exploiting where
as the \addgpbexpds with small $d$ methods are still exploring. We posit
that if we run for more iterations we will be able to
see the improvements. 

\subsection{SDSS Astrophysical Dataset}
\label{sec:lrgsExp}

Here we used Galaxy data from the Sloan Digital Sky Survey (SDSS). The task is
 to find the
maximum likelihood estimators for a simulation based astrophysical likelihood model.
Data and software for computing the likelihood are taken from~\citet{tegmark06lrgs}.
The software itself takes in only $9$ parameters but we augment this to $20$
dimensions to emulate the fact that in practical astrophysical problems we may
not know the true parameters on which the problem is dependent. 
This also allows us to effectively demonstrate the
superiority of our methods over alternatives.
Each query to this likelihood function takes about 2-5 seconds. In order to be
wall clock time competitive with RAND and \direct we use only $500$ evaluations
for \gpbalgo, \gpeis and REMBO and $450$ 
for \addgpbexpds to maximise the acquisition function.

We have shown the Maximum value obtained over $400$ iterations of each algorithm
in Figure~\ref{fig:sumlrg}. 
Note that RAND outperforms \directs here since a random query strategy is
effectively searching in $9$ dimensions. Despite this advantage to RAND all BO
methods do better. Moreover, despite the fact that the function may not be
additive, all \addgpbexpds methods outperform \gpbexp. 
Since the function only depends on $9$ parameters we use
REMBO with a $9$ dimensional projection. Yet, 
it is not competitive with the \addgpbexpds methods.
Possible reasons for this may include the scaling of the parameter space by
$\sqrt{d}$ in REMBO and the imperfect optimisation of the acquisition function.
Here
\addgpbexpii{5/4} performs slightly better than the rest since it seems to have
the best tradeoff between being statistically expressive enough to capture the
function while at the same time be easy enough to optimise the acquisition 
function within the allocated budget.

\subsection{Viola \& Jones Face Detection}
\label{sec:vjExp}

The Viola \& Jones (VJ) Cascade Classifier \cite{viola01cascade}
is a popular method for face detection in computer vision based on the Adaboost
algorithm.
The $K$-cascade has $K$ weak classifiers which outputs a score for any given
image. When we wish to classify an image we
pass that image through each classifier. If at any point the score falls below a
certain threshold the image is classified as negative. If the image passes
through all classifiers then it is classified as positive. The threshold values
at each stage are usually pre-set based on prior knowledge. There is no reason to
believe that these threshold values are optimal. In this experiment we wish to
find an optimal set of values for these thresholds by optimising the
classification accuracy over a training set.

For this task, we use $1000$ images from the Viola \& Jones face dataset
containing both face and non-face images. 
We use the implementation of the VJ classifier  that comes with OpenCV 
\cite{bradski08opencv} which uses a 22-stage cascade
and modify it to take in the threshold values as a parameter.
As our domain $\Xcal$ we choose a neighbourhood around the configuration given in
OpenCV.
Each function call takes about 30-40 seconds and is the the dominant cost
in this experiment. We use $1000$ \directs evaluations to optimise the
acquisition function for \gpbalgo, \gpeis and REMBO and $900$ for the \addgpbexpds instantiations.
Since we do not know the structure of the function we use 
REMBO with a $5$ dimensional projection. 
The results are given in Figure~\ref{fig:sumvj}. 
Not surprisingly, REMBO performs worst as
it is only searching on a $5$ dimensional space.
Barring
\addgpbexpii{1/22} all other instantiations perform better than \gpbalgos and
\gpeis with
\addgpbexpii{6/4} performing the best.
Interestingly, we also find a value for the thresholds that outperform the 
configuration used in OpenCV.

\insertFigureReal

}

\section{Conclusion}
\label{sec:conclusion}

\ifthenelse{\boolean{isabridged}}{
\vspace{-0.04in}
}
{}
\textbf{Recommendations: } Based on our experiences, we recommend the following.
If $\func$ is \emph{known} to be additive, the decomposition is known and
$d$ is small enough so that $\addutilt$ can be efficiently optimised,
then running \addgpbalgos with the known decomposition is likely to
produce the best results. If not, then use a small value for
$d$ and run \addgpbalgos while partially  optimising for
the decomposition periodically (Section~\ref{sec:practicalImplementation}). In
our experiments we found that using $d$ between $3$ an $12$ seemed reasonable choices. 
However, note that this depends on the computational budget
for optimising the acquisition, the query budget for $\func$
 and to a certain extent the the function $f$ itself.

\textbf{Summary: }
Our algorithm takes into account several practical considerations in real world
\gpbbos applications such as computational constraints in optimising the
acquisition and the fact that we have to work with a relatively few
data points since function evaluations are expensive.
Our framework effectively addresses these concerns without
considerably compromising on the statistical integrity of the model.
We believe that this provides a promising direction to scale \gpbbos methods to high
dimensions.

\textbf{Future Work: }
Our experiments indicate that our methods perform well beyond the scope
suggested by our theory. Developing an analysis that takes into
account the bias-variance and computational tradeoffs in approximating 
and optimising a 
non-additive function via an additive model is an interesting challenge.
\ifthenelse{\boolean{isabridged}}
{
We also intend to extend this framework to other acquisition functions.
}
{
We also intend to extend this framework to discrete settings, 
other acquisition functions and handle more general decompositions.
}

\section*{Acknowledgements}
We wish to thank Akshay Krishnamurthy and Andrew Gordon Wilson 
for the insightful discussions
and Andreas Krause, Sham Kakade and Matthias Seeger for
the helpful email conversations.
This research is partly funded by DOE grant DESC0011114.

Our current analysis, specifically equation~\ref{eqn:varTermBound}, has an error.
We are working on resolving this and will post an update shortly.
We would like to thank Felix Berkenkamp and Andreas Krause from ETH Zurich for
pointing this out.

{
\bibliography{./kky}
\bibliographystyle{icml2015}
}
\vfill

\pagebreak
\appendix
\onecolumn
\setboolean{istwocolumn}{false}

\section{Some Auxiliary Material}
\label{sec:appAncillary}

\subsection{Review of the \gpbalgos Algorithm}
\label{sec:reviewGPB}

In this subsection we present a brief summary of the \gpbalgos algorithm in
\cite{srinivas10gpbandits}. The algorithm is given in
Algorithm~\ref{alg:gpbalgo}.

The following theorem gives the rate of
convergence for \gpbalgo. Note that under an additive kernel, this 
is the same rate as Theorem~\ref{thm:addGPBthm} which uses a different
acquisition function.
Note the differences in the choice of $\beta_t$. 
\\[\thmparaspacing]

\begin{theorem}(Modification of Theorem 2 in \cite{srinivas10gpbandits})
Suppose $\func$ is constructed by sampling $\funcj\sim \GP(\zero, \kernelii{j})$
for $j = 1, \dots, M$ and then adding them. Let all kernels
$\kernelii{j}$ satisfy assumption~\ref{asm:partialDerivLipschitz} for some
$L,a,b$. Further, we maximise the acquisition function $\addutilt$ to within
$\zetabase t^{-1/2}$ accuracy at time step $t$.
Pick $\delta\in(0,1)$ and choose
\begingroup
\begin{align*}
\beta_t &= 2\log\left(\frac{2t^2\pi^2}{\delta}\right) + 
  2D\log\left(Dt^3\right)
  \;\;\in \; \bigO\left( D\log t \right).
\end{align*}
\endgroup
Then, {\rm\gpbalgo} attains cumulative regret $R_T \in 
\bigO\left(\sqrt{\vphantom{T^1}D\gamma_TT\log T}\right)$ and hence
simple regret $S_T \in \bigO\left(\sqrt{\vphantom{T^1}D\gamma_T\log T/T}\right)$.
Precisely, with probability $> 1- \delta$,
\[
\forall T \geq 1, \;\;\;
R_T \leq \sqrt{8C_1\beta_TMT\gamma_t} + 2\zetabase \sqrt{T} + C_2.
\]
where $C_1 = 1/\log(1+\eta^{-2})$ and $C_2$ is a constant depending on 
$a$, $b$, $D$, $\delta$, $L$ and $\eta$.
\begin{proof}
\citet{srinivas10gpbandits} bound the regret for exact maximisation of the
\gpbalgos acquisition $\utilt$. By following an analysis similar to our proof of
Theorem~\ref{thm:addGPBthm} the regret can be shown to be the same for an
$\zetabase t^{-1/2}$- optimal maximisation.
\end{proof}
\label{thm:GPBthm}
\end{theorem}
\vspace{\thmparaspacing}

\insertAlgorithmGPBALGO

\subsection{Sequential Optimisation Approaches}
\label{sec:appSeqGPB}

If the function is known to be additive, we could consider several other
approaches for maximisation. We list two of them here and explain their
deficiencies.
We recommend that the reader read the main text before reading this section.

\subsubsection{Optimise one group and proceed to the next}
First, fix the
coordinates of $\xii{j}, j \neq 1$ and optimise w.r.t $\xii{1}$ by querying the
function for a pre-specified number of times. Then we
proceed sequentially optimising with respect to $\xii{2}, \xii{3} \dots $.
We have outlined this algorithm in Algorithm~\ref{alg:seqgpbalgo1}.
There are several reasons this approach is not desirable.
\begin{itemize}
\item
First, it places too much faith on the additive assumption and requires
that we know the decomposition at the start of the algorithm. Note that this
strategy will only have searched the space in $M$ $d$-dimensional subspaces.  In
our approach even if the function is not additive we can still hope to do well
since we learn the best additive approximation to the true function.
Further, if the decomposition is not known we could learn the decomposition
``on the go" or at least find a reasonably good decomposition as we have
explained in Section~\ref{sec:practicalImplementation}. 

\item
Such a sequential approach is \emph{not} an anytime algorithm. This in
particular means that we need to predetermine the number of queries to be
allocated to each group.
After we proceed to a new group it is not straightforward to come back and 
improve on the solution obtained for an older group.

\item This approach is not suitable for the bandits setting.
We suffer large instantaneous regret up until we get to the last group.
Further, after we proceed beyond a group since we cannot come back, we cannot
improve on the best regret obtained in that group.

\end{itemize}
Our approach does not have any of these deficiencies.

\insertAlgorithmSEQADDGPBALGOONE

\subsubsection{Only change one Group per Query}

In this strategy, the approach would be very similar to \addgpbalgos except that
at each query we will only update one group at time. If it is the $k$\superscript{th}
group the query point is determined by maximising $\addutiltk$ for $\xpttii{k}$
and for all other groups we use values from the previous rotation. After $M$
iterations we cycle through the groups. We have outlined this in
Algorithm~\ref{alg:seqgpbalgo2}.

This is a reasonable approach and does not suffer from the same deficiencies as
Algorithm~\ref{alg:seqgpbalgo1}. Maximising the acquisition function will also
be slightly easier $\bigO(\zeta^{-d})$ since we need to optimise only one group 
at a time.
However, the regret for this approach would be
$\bigO(M\sqrt{\vphantom{T^a}D \gamma_T T\log T})$ which is a factor of $M$ worse
than the regret in our method (This can be show by following an
analysis similar to the one in section~\ref{sec:appAnalysisRates}. 
This is not surprising, since at each iteration
you are moving in $d$-coordinates of the  space and you have to wait $M$
iterations before the entire point is updated.

\insertAlgorithmSEQADDGPBALGOTWO

\newpage

\section{Proofs of Results in Section~\ref{sec:analysis}}
\label{sec:appAnalysis}

\subsection{Bounding the Information Gain $\gamma_T$}
\label{sec:appAnalysisInfGain}

For this we will use the following two results 
from \citet{srinivas10gpbandits}. \\

\begin{lemma}(Information Gain in GP, \cite{srinivas10gpbandits} Lemma 5.3)
Using the basic properties of a GP, they show that 
\[
I(y_A; \func_A) = \frac{1}{2} \sum_{t=1}^{n} \log(1 + \eta^{-2}
  \sigma^2_{t-1}(x_t) ).
\]
where $\sigma^2_{t-1}$ is the posterior variance after observing the first $t-1$
points. \\[\thmparaspacing]
\end{lemma}

\begin{theorem}(Bound on Information Gain, \cite{srinivas10gpbandits} Theorem 8)
Suppose that $\Xcal$ is compact and $\kernel$ is a kernel on $d$ dimensions
 satisfying Assumption~\ref{asm:partialDerivLipschitz}.
Let $n_T = C_9 T^\tau \log T$ where $C_9 = 4d+2$.
For any $T_* \in \{1, \dots, \min(T, n_T)\}$,
 let $B_\kernel(T_*) = \sum_{s>T_*}
\lambda_s$. Here $(\lambda_n)_{n\in \NN}$ are the eigenvalues of $\kernel$ w.r.t
the uniform distribution over $\Xcal$. 
Then,
\begin{equation*}
\gamma_T \leq  \inf_\tau  \left(
  \frac{1/2}{1-e^{-1}} \max_{r\in\cbr{1,\dots,T}} \left( T_* \log(r n_T/
\eta^2) + C_9 \eta^2 (1 - r/T) (T^{\tau+1} B_\kernel(T_*) + 1) \log T \right)
  + \bigO(T^{1-\tau /d}) 
\right).
\end{equation*}
\label{thm:srinivas}
\end{theorem}

\subsubsection{ \textbf{Proof of Theorem~\ref{thm:infGainBounds}-1}}
\begin{proof}
We will use some bounds on the eigenvalues for the simple squared exponential
kernel given in \cite{seeger08information}. It was shown that the eigenvalues
$\{\lambdaii{i}_s\}$ of $\kernelii{i}$ satisfied $\lambdaii{i}_s \leq
c^dB^{s^{1/d_i}}$ where $B < 1$ (See Remark~\ref{rmk:eigvaldistros}).
Since the kernel is additive, and $\xii{i} \cap
\xii{j} = \emptyset$ the eigenfunctions corresponding to $\kernelii{i}$ and
$\kernelii{j}$ will be orthogonal. Hence the eigenvalues of $\kernel$ will
just be the union of the eigenvalues of the individual kernels -- i.e. 
$\{\lambda_s\} = \bigcup_{j=1}^M \{\lambdaii{j}_s \}$.  
As $B<1$, $\lambdaii{i}_s \leq c^dB^{s^{1/d}}$. Let $\Thash = \floor{T_*/M}$ and
$\alpha = -\log B$.
Then,
\begin{align*}
B_\kernel(T_*) &= \sum_{s>T_*} \lambda_s \leq Mc \sum_{s>\Thash} B^{s^{1/d}} \\
  &\leq c^dM \left( B^{\Thash^{1/d}} + \int_{\Thash}^\infty \exp(-\alpha x ^{1/d})
            \right) \ud x \\
  &\leq c^dM \left( B^{\Thash^{1/d}} + d \alpha^{-d} \Gamma(d, \alpha \Thash^{1/d})
            \right) \\ 
  &\leq c^dMe^{-\alpha \Thash^{1/d}} 
  \left( 1 + d! d \alpha^{-d} (\alpha \Thash^{1/d})^{d-1} \right).
\end{align*}
The last step holds true whenever $\alpha \Thash^{1/d} \geq 1$.
Here in the second step we bound the series by an integral and in the third step
we used the substitution $y =
\alpha x^{1/d}$ to simplify the integral. Here $\Gamma(s,x) = \int_x^\infty
t^{s-1} e^{-t} \ud t$ is the (upper) incomplete Gamma function. In the last step
we have used the following identity and the bound for integral $s$ and $x\geq 1$
\[
\Gamma(s,x) = (s-1)! e^{-x} \sum_{k=0}^{s-1} \frac{x^k}{k!} \leq s! e^{-x} x^{d-1}.
\]
By using $\tau = d$ and by using
$T_* \leq (M+1) \Thash$, we
use Theorem~\ref{thm:srinivas} to obtain the following bound on $\gamma_T$,
\begin{align*}
\gamma_T &\leq 
  \frac{1/2}{1-e^{-1}} \max_{r\in\cbr{1,\dots,T}} \bigg( (M+1) \Thash \log(r n_T/
\eta^2) + \\
 &\hspace{0.2in} 
  C_9 \eta^2 (1 - r/T)  \log T \left( 1 + c^dMe^{-\alpha \Thash^{1/d}} T^{d+1} 
  \left( 1 + d! d \alpha^{-d} (\alpha \Thash^{1/d})^{d-1} \right)
  \right) \bigg).
\label{eqn:gammarbound} \numberthis
\end{align*}
Now we need to pick $\Thash$ so as to balance these two terms. We will choose
$\Thash = \left(\frac{\log(Tn_T)}{\alpha}\right)^d $ which is less than $\min(T,
n_T)/M$ for sufficiently large $T$. 
Then $e^{-\alpha \Thash^{1/d}} = 1/Tn_T$. Then the first term $S_1$ inside the
paranthesis is,
\begin{align*}
S_1 = (M+1) \log^d\left(\frac{Tn_T}{\alpha}\right)
        \log\left(\frac{rn_T}{\eta^2}\right) 
  &\in \bigO\left(M \left(\log(Tn_T)\right)^{d}\log(rn_T) \right) \\
  &\in \bigO\left(M \left(\log(T^{d+1}\log T)\right)^{d} \log(rT^d \log T)\right) \\
  &\in \bigO\left(M d^{d+1} (\log T)^{d+1} + M d^d (\log T)^d\log(r) \right).
\end{align*}
Note that the constant in front has exponential dependence on $d$ but we ignore
it since we already have $d^d$, $(\log T)^d$ terms.
The second term $S_2$ becomes,
\begin{align*}
S_2 &= 
  C_9 \eta^2 (1 - r/T)  \log T \left( 1 + \frac{c^dM}{Tn_T} T^{d+1} 
  \left( 1 + d! d \alpha^{-d} (\log(Tn_T)^{d-1} \right) 
  \right) \bigg) \\
  &\leq C_9 \eta^2 (1 - r/T)  \left( \log T  + \frac{c^dM}{C_9}  
  \left( 1 + d! d \alpha^{-d} (\log(Tn_T)^{d-1} \right) 
  \right) \bigg) \\
  &\leq C_9 \eta^2 (1 - r/T)  \left( \bigO(\log T) + \bigO(1) +
   \bigO(d! d^{d} (\log T)^{d-1}) \right) \bigg) \\
  &\in \bigO\left( (1-r/T)d! d^{d} (\log T)^{d-1} \right).
\end{align*}
Since $S_1$ dominates $S_2$, we should choose $r=T$ to maximise the RHS
in~\eqref{eqn:gammarbound}. This gives us,
\[
\gamma_T 
\;\in\; \bigO\left(M d^{d+1} (\log T)^{d+1}\right) 
\;\in\; \bigO\left(D d^{d} (\log T)^{d+1} \right).
\]
\end{proof}

\subsubsection{ \textbf{Proof of Theorem~\ref{thm:infGainBounds}-2}}
\begin{proof}
Once again, we use bounds given in \cite{seeger08information}. 
It was shown that the eigenvalues $\{\lambdaii{i}_s\}$ for $\kernelii{i}$
satisfied $\lambdaii{i}_s \leq c^ds^{-\frac{2\nu+d_j}{d_j}}$
(See Remark~\ref{rmk:eigvaldistros}). By following a
similar argument to above we have 
$\{\lambda_s\} = \bigcup_{j=1}^M \{\lambdaii{j}_s \}$ and
$\lambdaii{i}_s \leq c^ds^{-\frac{2\nu+d}{d}}$. Let $\Thash = \floor{T_*/M}$.
Then,
\begin{align*}
B_\kernel(T_*) = \sum_{s>T_*} \lambda_s 
  \leq Mc^d\sum_{s>\Thash} s^{-\frac{2\nu+d}{d}} 
  \leq Mc^d\left( \Thash^{-\frac{2\nu+d}{d}} + 
      \int_{\Thash}^\infty s^{-\frac{2\nu+d}{d}}\right) 
  \leq C_8 2^d M\Thash^{1-\frac{2\nu+d}{d}}.
\end{align*}
where $C_8$ is an appropriate constant.
We set $\Thash = (Tn_T)^{\frac{d}{2\nu+d}} (\log(Tn_T))^{-\frac{d}{2\nu+d}}$ and
accordingly we have the following bound on $\gamma_T$ as a function of $\Thash \in
\{1,\dots, \min(T, n_T)/M\}$,
\begin{align*}
\gamma_T \leq 
  \inf_\tau \Bigg(
  \frac{1/2}{1-e^{-1}} \max_{r\in\cbr{1,\dots,T}} \bigg( (M+1) \Thash \log(r n_T/
\eta^2) +  
  C_9 \eta^2 (1 - r/T)   \left( \log T + C_8 2^d M \Thash \log(Tn_T) \right)
  \bigg) + \bigO(T^{1-\tau/d}) \Bigg).
\label{eqn:gammarboundmatern} \numberthis
\end{align*}
Since this is a concave function on $r$ we can find the optimum by setting the
derivative w.r.t $r$ to be zero. We get $r \in \bigO(T/2^d\log(Tn_T))$ and hence,
\begin{align*}
\gamma_T &\in \inf_\tau 
\left( \bigO\left( M \Thash \log\left(\frac{Tn_T}{2^d\log(Tn_T)}\right) \right)
  + \bigO\left(M 2^d \Thash \log(Tn_T) \right)
      + \bigO(T^{1-\tau/d}) \right) \\
&\in \inf_\tau \left(\bigO \left(M 2^d \log(Tn_T)
   \left(\frac{T^{\tau+1}\log(T)}{(\tau+1)\log(T) + \log\log T}\right)^\frac{d}{2\nu+d}
   \right) 
  + \bigO(T^{1-\tau/d}) \right) \\
&\in \inf_\tau \left(\bigO \left(M2^d\log(Tn_T)
   T^{\frac{(\tau+1)d}{2\nu+d}} \right) 
  + \bigO(T^{1-\tau/d}) \right) \\
&\in \bigO\left( M 2^d T^{\frac{d(d+1)}{2\nu + d(d+1)}} \log(T) \right).
\end{align*}
Here in the second step we have substituted the values for $\Thash$ first and
then $n_T$. In the last step we have balanced the polynomial dependence on $T$
in both terms by setting $\tau = \frac{2\nu d}{2\nu + d(d+1)}$. \\
\end{proof}

\begin{remark}
The eigenvalues and eigenfunctions for the kernel are defined with respect to a
base distribution on $\Xcal$. In the development of Theorem~\ref{thm:srinivas},
\citet{srinivas10gpbandits} draw $n_T$ samples from the uniform distribution on
$\Xcal$. Hence, the eigenvalues/eigenfunctions should be w.r.t the uniform
distribution. The bounds given in \citet{seeger08information} are for the uniform
distribution for the \matern kernel and a Gaussian Distribution for the Squared
Exponential Kernel. For the latter case, \citet{srinivas10gpbandits} argue that
the uniform distribution still satisfies the required tail constraints and
therefore the bounds would only differ up to constants.
\label{rmk:eigvaldistros}
\end{remark}

\vspace{0.1in}

\subsection{Rates on \addgpbalgo}
\label{sec:appAnalysisRates}

Our analysis in this section draws ideas from \citet{srinivas10gpbandits}. We will
try our best to stick to their same notation. However, unlike them we also
handle the case where the acquisition function is optimised within some error.
In the ensuing discussion, we will use $\xpttm = \bigcup_j \xpttmii{j}$ to
denote the true maximiser of $\addutilt$ -- i.e. $\xpttmii{j} = \argmax_{z \in
\Xcalj} \addutiltj(z)$.  $\xptt = \bigcup_j\xpttii{j}$  denotes the point
chosen by \addgpbalgos at the $t^{th}$ iteration. Recall that $\xptt$ is
$\zetabase t^{-1/2}$--optimal; I.e. $\addutilt(\xpttm) - \addutilt(\xptt) \leq
\zetabase t^{-1/2}$.

Denote $p = \sum_j d_j$.
$\pi_t$
denotes a sequence such that $\sum_t \pi_t^{-1} = 1$. For e.g. when we use
$\pi_t = \pi^2 t^2/6$ below, we obtain the rates in Theorem~\ref{thm:addGPBthm}.

In what follows, we will construct discretisations
$\Omegaj$ on each group $\Xcalj$ for the sake of analysis.
Let $\omega_j = |\Omegaj|$
and $\omega_m = \max_j \omega_j$.
The discretisation of the individual groups induces a discretisation $\Omega$
on $\Xcal$ itself, $\Omega = \{ \xpt = \bigcup_j \xptii{j} : \xptii{j} \in \Omegaj,
j = 1,\dots, M \}$. Let $\omega = |\Omega| = \prod_j \omega_j$.
We first establish the following two lemmas before we prove
Theorem~\ref{thm:addGPBthm}.
\\[\thmparaspacing]

\begin{lemma}
Pick $\delta \in (0,1)$ and set $\beta_t = 2\log(\omega_m M \pi_t/\delta)$. Then
with probability $> 1-\delta$,
\[ \forall t \geq 1, \forall \xpt \in \Omega,
 \hspace{0.2in}
|f(\xpt) - \mu_{t-1}(\xpt)| \leq \betath \sumsigma{\xptii{j}}.\]
\begin{proof}
Conditioned on $\Dcal_{t-1}$, at any given $\xpt$ and $t$ we have
$f(\xptii{j}) \sim \Ncal(\gpmeanjtmo(\xptii{j}), \gpstdjtmo{j}), \; \forall j = 1,
\dots M$. Using the tail bound, $\PP(z>M) \leq \frac{1}{2}e^{-M^2/2}$ for $z\sim
\Ncal(0,1)$ we have with probability $>1-\delta/\omega M \pi_t$,
\[
 \frac{|\funcj(\xptii{j}) -
\gpmeanjtmo(\xptii{j})|}{\gpstdjtmo(\xptii{j})} > \betath  \leq
e^{-\beta_t/2} = \frac{\delta}{\omega_m M\pi_t}.
\]
By using a union bound $\omega_j\leq \omega_m$ times over all 
$\xptii{j} \in \Omegaj$ and then $M$ times over all discretisations 
the above holds with
probability $>1-\delta/\pi_t$ for all $j=1,\dots,M$ and $\xptii{j} \in \Omegaj$.
Therefore, we have $|f(\xpt) - \gpmeantmo(\xpt)|
\leq |f(\xptii{j}) - \gpmeanjtmo(\xptii{j})| \leq 
\betath\sum_j \gpstdjtmo(\xptii{j})$ for all $\xpt \in \Omega$.
Now using the union bound on all $t$ yields
the result. \\[\thmparaspacing]
\end{proof}
\label{lem:discBound}
\end{lemma}

\begin{lemma}
The posterior mean $\gpmeantmo$ for a GP whose
kernel $\kernel(\cdot, x)$ is $L$-Lipschitz satisfies,
\[
\PP\left( \forall t \geq 1 \;\;\;
|\gpmeantmo(x) - \gpmeantmo(x')| \leq
\left(\func(\xopt) + \eta\sqrt{2\log(\pi_t/2\delta)}\right) L\eta^{-2} t
 \|x-x'\|_2 \right)
\geq 1-\delta.
\]
\begin{proof}
Note that for given $t$, 
\begin{align*}
&\PP\left(y_t < \func(\xopt) + \eta\sqrt{2\log(\pi_t/2\delta)}\right) \leq 
  \PP\left(\epsilon_t/\eta < \sqrt{2\log(\pi_t/2\delta)}\right) \leq 
  \delta/\pi_t. 
\end{align*}
Therefore the statement is true with probability $>1-\delta$ for all $t$.
Further, $\Delta \succ \eta^2 I$ implies $\|\Delta^{-1}\|_{op}\leq \eta^{-2}$ and 
$|k(x, z) - k(x',z)| \leq L \|x-x'\|$.
Therefore
\begin{align*}
&|\gpmeantmo(x) - \gpmeantmo(x')| = |Y_{t-1}^\top \Delta^{-1} (k(x,X_T) -
k(x',X_T)| \leq
  \|Y_{t-1}\|_2 \|\Delta^{-1}\|_{op} \|k(x,X_{t-1}) - k(x',X_{t-1})\|_2 \\
&\hspace{0.2in}\leq 
\left(\func(\xopt) + \eta\sqrt{2\log(\pi_t/2\delta)}\right) L\eta^{-2} (t-1)
\|x-x'\|_2.
\end{align*}
\end{proof}
\label{lem:smoothPostMean}
\end{lemma}


\subsubsection{\textbf{Proof of Theorem~\ref{thm:addGPBthm}}}
\begin{proof}
First note that by Assumption~\ref{asm:partialDerivLipschitz} and the union
bound we have,
$
\PP( \forall i\; \sup_{\xii{j} \in \Xcalj} |
{\partial \funcj(\xii{j})}/{\partial \xii{j}_i}| > J )
\leq d_i a e^{-(J/b)^2}
$.
Since, ${\partial \func(x)}/{\partial \xii{j}_i} =
{\partial \funcj(\xii{j})}/{\partial \xii{j}_i}$, we have,
\[
\PP\left( \forall i =1,\dots,D\; \sup_{x \in \Xcal} \Big|
\partialfrac{x_i}{\func(x)}\Big| > J \right)
\leq p a e^{-(J/b)^2}.
\]
By setting $\delta/3=pae^{-J^2/b^2}$ we have with probability $>1-\delta/3$,
\begin{equation*}
\forall x, x' \in \Xcal,\; |f(x) - f(x')| \leq b \sqrt{\log(3ap/\delta)} \|x -
x'\|_1.
\numberthis \label{eqn:fLipBound}
\end{equation*}

Now, we construct a sequence of discretisations $\Omegajt$ 
satisfying $\|\xii{j} - [\xii{j}]_t]\|_1
\leq d_j/\tau_t\;\;
\forall \xii{j} \in \Omegajt$. Here, $[\xii{j}]_t$
is the closest point to $\xii{j}$ in $\Omegajt$ in an $L_2$ sense.
A sufficient discretisation is a grid with $\tau_t$
uniformly spaced points. 
Then it follows that for all $x\in \Omega_t$, $\|x-[x]_t\|_1 \leq p/\tau_t$.
Here $\Omega_t$ is the discretisation induced on $\Xcal$ by the $\Omegajt$'s
and $[x]_t$ is the closest point to $x$ in $\Omega_t$.
Note that $\|\xii{j} - [\xii{j}]_t\|_2 \leq \sqrt{d_j}/\tau_t\; \forall
\xii{j} \in \Omegaj$ and $\|x - [x]_t\|_2 \leq \sqrt{p}/\tau_t$.
We will set $\tau_t = pt^3$--therefore,
$\omega_{tj} \leq (pt^3)^d \defeq \omega_{mt}$.
When combining this with~\eqref{eqn:fLipBound}, we get that with probability
$>1-\delta/3$, $|\func(x) - \func([x])| \leq b \sqrt{\log(3ap/\delta)}/t^3$.
By our choice of $\beta_t$ and using Lemma~\ref{lem:discBound} 
the following  is true for all $t \geq 1$ and for all $x \in \Xcal$ 
with probability $>1-2\delta/3$,
\begin{equation}
|f(x) - \gpmeantmo([x]_t)| \leq 
|f(x) - f([x]_t)| + |\func([x]_t) - \gpmeantmo([x]_t)| \leq 
\frac{b \sqrt{\log(3ap/\delta)}}{t^2} + \betath \sumsigma{[\xii{j}]_t}.
\label{eqn:fminusgpmean}
\end{equation}
By Lemma~\ref{lem:smoothPostMean} with probability $>1-\delta/3$ we have,
\begin{equation}
\forall x \in \Xcal, \;\;\;
|\gpmeantmo(x) - \gpmeantmo([x]_t)| \leq
\frac{L\left(\func(\xopt) + \eta\sqrt{2\log(3\pi_t/2\delta)}\right)}
{ \sqrt{p}\eta^2 t^2}.
\label{eqn:smoothPostMean}
\end{equation}
We use the above results to obtain the following bound on the 
instantaneous regret $r_t$ which holds with probability $> 1-\delta$ for all
$t\geq 1$,
\begin{align*}
r_t &= \func(\xopt) - f(\xptt) \\
  &\leq 
  \gpmeantmo([\xopt]_t) + \betath \sumsigma{[\xoptii{j}]_t} 
  -\gpmeantmo([\xptt]_t) + \betath \sumsigma{[\xpttii{j}]_t} +
  \frac{2\baplogdelta}{t^3} \\
  &\leq \frac{2\baplogdelta}{t^3} + \frac{\zetabase}{\sqrt{t}} + 
   \betath \left(\sumsigma{\xpttii{j}} +
      \sumsigma{[\xpttii{j}]_t}\right)
  + \gpmeantmo(\xptt) - \gpmeantmo([\xptt]_t)  \\
  &\leq \frac{2\baplogdelta}{t^3} + 
  \frac{L\left(\func(\xopt) + \eta\sqrt{2\log(\pi_t/2\delta)}\right)}
    { \sqrt{p}\eta^{2} t^2}
    + \frac{\zetabase}{\sqrt{t}} + \betath \left(\sumsigma{\xpttii{j}} +
      \sumsigma{[\xpttii{j}]_t}\right). \numberthis \label{eqn:instantBound}
\end{align*}
In the first step we have applied Equation~\eqref{eqn:fminusgpmean} at $\xopt$ and
$\xptt$. In the second step we have used the fact that 
$\addutilt([\xopt]_t) \leq \addutilt(\xpttm) \leq \addutilt(\xptt) + \zetabase
t^{-1/2}$. In the third step we have used Equation~\eqref{eqn:smoothPostMean}.

For any $x \in \Xcal$ we can bound ${\gpstd_t(x)}^2$ as follows,
\[
{\gpstd_t(x)}^2= \eta^2 \eta^{-2} {\gpstd_t(x)}^2
  \leq \frac{1}{\log(1+\eta^{-2})} \log\left( 1+ \eta^{-2}
{\gpstd_t(x)}^2\right).
\]
Here we have used the fact that $u^2 \leq v^2 \log(1+u^2)
/\log(1+v^2)$ for $u \leq v$ and $ {\gpstd_t(\xpt)}^2 \leq \kernel(x,x) = 1$. 
Write $C_1 = \log^{-1}(1+\eta^{-2})$.
By using Jensen's inequality and Definition~\ref{def:infgainGP} for any set
of $T$ points $\{x_1, x_2, \dots x_T\} \subset \Xcal$,
\begin{equation}
\left(\sum_{t=1}^T \sum_{j=1}^M \gpstdii{j}_{t}(\xii{j}) \right)^2
\leq MT\sum_{t=1}^T \sum_{j=1}^M {\gpstdii{j}_{t}(\xii{j})}^2
\leq C_1 MT\sum_{t=1}^T \log\left( 1+ \eta^{-2}
{\gpstd_t(x)}^2\right)
\leq  2 C_1 MT\gamma_T.
\label{eqn:varTermBound}
\end{equation}

Finally we can bound the cumulative regret with probability $>1-\delta$ for all
$T\geq 1$ by,
\begin{align*}
R_T &= \sum_{t=1}^T r_t \leq C_2(a,b,D,L,\delta) +
  \zeta_0 \sum_{t=1}^T t^{-1/2} + \beta_T^{1/2} \left(\sum_{t=1}^T\sumsigma{\xpttii{j}} +
      \sum_{t=1}^T\sumsigma{[\xpttii{j}]_t}\right) \\
  &\leq C_2(a,b,D,L,\delta) + 2\zetabase \sqrt{T} +
  \sqrt{8C_1\beta_TMT\gamma_T}.
\end{align*}
where we have used the summability of the first two terms in
Equation~\eqref{eqn:instantBound}. Here, for $\delta < 0.8$, 
the constant $C_2$ is given by,
\[
C_2 \geq 
\baplogdelta + \frac{\pi^2 L f(\xopt)}{6 \sqrt{p}\eta^2} + 
\frac{L\pi^{3/2}}{\sqrt{12 p \delta}\eta }.
\]
\end{proof}

\ifthenelse{\boolean{isabridged}}
{
}
{
}

\end{document}